# Solar Power Time Series Forecasting Utilising Wavelet Coefficients

Sarah Almaghrabi[a,b], Mashud Rana[c], Margaret Hamilton[a] and Mohammad Saiedur Rahaman[a]

[a]School of Computing Technologies, RMIT University, Melbourne, Australia
[b]School of Computing Technologies, University of Jeddah, Saudi Arabia
[c]Data61, CSIRO, Sydney, Australia

ARTICLE INFO

ABSTRACT

Accurate and reliable prediction of Photovoltaic (PV) power output is critical to electricity grid stability and power dispatching capabilities. However, Photovoltaic (PV) power generation is highly volatile and unstable due to different reasons. The Wavelet Transform (WT) has been utilised in time series applications, such as Photovoltaic (PV) power prediction, to model the stochastic volatility and reduce prediction errors. Yet the existing Wavelet Transform (WT) approach has a limitation in terms of time complexity. It requires reconstructing the decomposed components and modelling them separately and thus needs more time for reconstruction, model configuration and training. The aim of this study is to improve the efficiency of applying Wavelet Transform (WT) by proposing a new method that uses a single simplified model. Given a time series and its Wavelet Transform (WT) coefficients, it trains one model with the coefficients as features and the original time series as labels. This eliminates the need for component reconstruction and training numerous models. This work contributes to the day-ahead aggregated solar Photovoltaic (PV) power time series prediction problem by proposing and comprehensively evaluating a new approach of employing WT. The proposed approach is evaluated using 17 months of aggregated solar Photovoltaic (PV) power data from two real-world datasets. The evaluation includes the use of a variety of prediction models, including Linear Regression, Random Forest, Support Vector Regression, and Convolutional Neural Networks. The results indicate that using a coefficients-based strategy can give predictions that are comparable to those obtained using the components-based approach while requiring fewer models and less computational time.

## Acronyms

| | |
|---|---|
| $\psi$ | Mother Wavelet |
| $A_1$ | Approximation Component at $DL = 1$ |
| $A_2$ | Approximation Component at $DL = 2$ |
| $cA_1$ | Approximation Coefficients at $DL = 1$ |
| $cA_2$ | Approximation Coefficients at $DL = 2$ |
| $cD_1$ | Detail Coefficients at $DL = 1$ |
| $cD_2$ | Detail Coefficients at $DL = 2$ |
| $D_1$ | Detail Component at $DL = 1$ |
| $D_2$ | Detail Component at $DL = 2$ |
| $DL$ | Decomposition Level |
| $pad_{LR}$ | Padding using Linear Regression |
| $pad_{REP}$ | Padding using Repetition |
| $R^2$ | Coefficient of Determination |
| CNN | Convolutional Neural Networks |
| CWT | Continuous Wavelet Transform |
| DWT | Discrete Wavelet Transform |
| GA | Genetic Algorithm |
| ISWT | Inverse Stationary Wavelet Transform |
| IWT | Inverse Wavelet Transform |
| LR | Linear Regression |
| LSTM | Long Short-Term Memory |
| MAE | Mean Absolute Error |
| MC | Multiple Channels |
| MI | Multiple Inputs |
| MM | Multiple Models |
| MRE | Mean Relative Error |
| NNs | Neural Networks |
| PCA | Principal Component Analysis |
| PV | Photovoltaic |
| RAE | Relative Absolute Error |
| RF | Random Forest |
| RMSE | Root Mean Squared Error |
| RRSE | Root Relative Squared Error |
| SVR | Support Vector Regression |
| SWT | Stationary Wavelet Transform |
| WT | Wavelet Transform |

ORCID(s):

## 1. Introduction

Accurate and reliable forecasting of solar Photovoltaic (PV) power is critical for ensure electricity grid stability and economic operations. It enables grid operators to estimate and balance energy generation and demand. However, forecasting PV energy generation with high accuracy is challenging [1, 2]. That is because of its irregular and non-linear nature, which is inherited from the rapidly expanding rate of penetration in the energy market, and the variability and uncertainty due to different weather parameters [2].

Numerous time series prediction problems employ Wavelet Transform (WT) as a pre-processing step, particularly for non-linear and non-stationary data. This is to generate a more stable time series with less variability and abrupt changes. The most frequently used strategy for applying the WT is summarised in Figure 1. It is based on using the WT to decompose the time series and generate multiple coefficients for the details and approximation components. Some models reconstruct the components using the Inverse Wavelet Transform (IWT) function before the prediction while others reconstruct them after the prediction; the details of the reconstruction process will be discussed in Section 3.

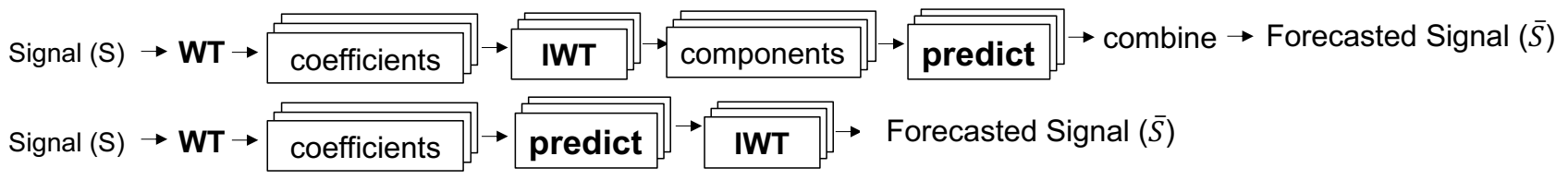

**Figure 1:** Summary of general strategy to apply Wavelet Transform (WT) for time series predictions based on multiple models and referred to as components-based approach

The next step is to configure a prediction model for each of the resulting time series (i.e., coefficients or reconstructed components). The forecasts are then aggregated as the last step, either using an aggregation method or IWT for the predicted coefficients.

While this approach has certain merits, there are a number of difficulties to contend with. Apart from the challenge of determining the optimum settings (including selecting the ideal Mother Wavelet ($\psi$) function, and Decomposition Level ($DL$)), determining the optimal components and developing several models to represent the various components complicates the use of WT.

The choice of the $\psi$ and the DL is a crucial step in demonstrating the benefits of the WT [3–7]. The selection of $\psi$ can either be empirical or depend on the visual inspection of the repeating signal pattern coupled with previous experience. It is challenging to discover the optimal $\psi$ because it varies depending on the complexity of the input data [4]. For the empirical method, there is a need to explore different DLs for the best possible results and then try several $\psi$ functions until the most accurate one is found. Generally, matching the characteristics of the signal to be analysed with the characteristics of the wavelet functions is difficult because there are so many wavelet functions, each with their own distinct attributes. In addition, environmental or energy data exhibits a large amount of variation in the signal, which also needs to be considered when selecting the $\psi$. Therefore, an accurate strategy for finding the most appropriate $\psi$ function or DL remains a challenge. The analysis of the various settings and their impact on the solar PV forecasting problem is still unknown. Most existing studies use the indicated approaches to pre-select the ideal settings (i.e., DL and $\psi$) before building and evaluating the intended models, such as [8] and [9]. The analysis of this decision can aid in understanding the impact of the different parameter selection on prediction error and suggest improved selection procedures.

Moreover, the number of decomposed time series components corresponds directly to the degree of decomposition, i.e., for each $DL \in [1, \infty]$ there are $DL + 1$ components. As the component number rises, the computational complexity increases because every component needs to be modelled and interpolated. To minimise this issue, certain research projects have taken the optimal wavelet component selection into account. For example, Liu et al.[10] use Principal Component Analysis (PCA) to minimise the number of the components. Du et al.[6] discard the detail component and utilise the approximation component only. This step may be beneficial especially when the DL is high. However, it may result in loss of critical information, lowering the quality of forecasts, if not applied properly. Almaghrabi et al.[8] reduce the total complexity by modelling the detail component using a simple LR model. With DL equal to 1, the advantage of this strategy is demonstrated. However, when the DL increases, the complexity of the model still remains a concern.

To address the drawbacks of multiple independent models and long training times, this paper proposes the adoption of a more efficient method for using WT, as a coefficients-based approach. This method first uses the Stationary Wavelet Transform (SWT) to generate the wavelet coefficients from the original signal. Then, a single prediction model is trained using the coefficients and the desired original values as labels. This strategy, in which only one model needs to be processed, requires far less computing as there is no need to apply the wavelet inverse function, no need to apply multiple models (for each component) and no need to recombine the predictions. The approach is to be utilised and analysed for the sequence-to-sequence aggregated day-ahead solar PV power forecasting problem. However, it can be applied to any time series prediction problem.

The motivation for this research is twofold: first, maintaining a stable electricity grid on a large scale requires accurate prediction models with low resource costs; and second, the utility of machine learning and deep learning models to extract information and learn from multiple features/channels. The advantages of the proposed approach can be summarised as: 1) It relies on the use of the coefficients only; 2) There is no need to use the inverse function (e.g., Inverse Stationary Wavelet Transform (ISWT)) to reconstruct the components; 3) It can use one model to find the forecast signal; 4) It allows the time series forecasting model to jointly learn the decomposed data; 5) It is more efficient in terms of time required for model configuration and training; 6) It is a more generic method that can be combined with any prediction algorithm or technique.

The novelty of this research is that it proposes a method for utilising the SWT that uses the decomposed signal's coefficients as features and the original signal as labels. The contributions of this paper to the literature of solar PV power time series forecasting can be summarised as follows:

1. We demonstrate an efficient approach for forecasting solar PV power using WT. The proposed solution is based on the SWT, which is time-invariant. It utilises the coefficients of the transformed data and does not require component reconstruction. We explore two paradigms for implementing the coefficients-based method (Multiple Channels (MC) and MI) and compare them to the standard practice of implementing

WT based on the reconstructed components utilising the Multiple Models (MM) paradigm. With the suggested approach, a single model is used to learn and forecast the data using the coefficients and the original signal, which reduces the number of models needed compared to the standard approach. It significantly reduces the time for training the prediction models and requires fewer model parameters to be configured. Additionally, it increases model generalisation, as different prediction models and techniques can be integrated using the proposed approach while preserving the benefits of WT.

2. We provide the first study to analyse several SWT settings in the context of solar PV power forecasting. These settings include varying the most sensitive SWT parameters: the DL and the wavelet family order. The findings highlight the necessity to find the optimal settings for utilising SWT for solar PV power forecasting. We investigate the model's sensitivity to various configurations using various paradigms, including MM, MC and MI. In comparison to the components-based, the proposed coefficients-based approach is less computationally time-sensitive to the DL.
3. We develop two distinct padding methods for reducing the impact of the border distortion caused by the SWT. The first approach is based on forecasting the future, whereas the second way is based on signal repetition. The second strategy is proven to produce accurate predictions in less time.
4. We conduct a comprehensive evaluation of the performance of the proposed coefficients-based approach using two different solar power generation datasets from Australia. We analyse the performance in terms of accuracy and time efficiency. The analysis is undertaken at various times of the day and year to demonstrate the seasonal effects. The SWT is used in conjunction with several prediction models for solar PV power forecasting. These include machine-learning and deep-learning models such as Linear Regression (LR), Support Vector Regression (SVR), Random Forest (RF), and Convolutional Neural Networks (CNN) with and without the SWT .

The following sections are organized as follows: Section 2 reviews the related work, Section 3 discusses the wavelet process; Section 4 describes the problem statement and the dataset, Section 5 presents the framework of the proposed coefficients-based approach, Section 6 presents the benchmarks and evaluation metrics, Section 7 discusses the results and Section 8 concludes the paper.

## 2. Related Work

This section reviews the literature for solar PV power. It also discusses WT applications to time series data in various disciplines, including solar PV forecasting.

### 2.1. Solar PV power time series forecasting

A great variety of prediction models have been investigated in the literature for solar PV power forecasting including statistical models, machine-learning models, deep-learning models, and even ensemble models [1, 2, 11–18]. The widely used statistical models include Auto-Regressive Integrated Moving Average (ARIMA) [19], LR [20], and Exponential Smoothing (ES) [21]. They demonstrate promising performance, particularly when dealing with univariate data. However, the main limitation of these models is that they require domain knowledge and prior assumptions on the distribution of data, e.g. they assume the signal to be stationary. They also necessitate a constant correlation between inputs and outputs and feature extraction from the input data [22].

The most prominent machine learning models for forecasting solar PV power include Neural Networks (NNs) [23–27], SVR [26, 28–32], and RF [28, 32, 33]. In addition, deep learning models such as Long Short-Term Memory (LSTM) and CNN are also applied in several recent studies [34–38]. Machine learning and deep learning models have distinct advantages compared to statistical models. The key benefits include the potential to learn non-linear relationships between input and target variables without the need for any pre-assumptions of the data [17].

For example, Kraemer et al. [26] apply different machine learning models for solar PV power forecasting that include k-nearest-neighbour (kNN), SVR, NNs, and Decision Tree (DT). As the inputs to the models, they use historical time series data, different weather variables and sun position. They find that all the models can perform similarly except kNN. Zeng et al. [39] develop a model based on a least-square support vector machine (LS-SVM) for short-term solar PV power prediction using meteorological data. Their model provides superior performance over Autoregressive (AR) and Radial Basis Function NNs (RBFNNs) models. Rana et al. [37] consider a variety of models, including NNs, SVR, CNN, and LSTM, for day-ahead aggregated solar PV power prediction. They use historical power data and descriptive statistics as inputs. The SVR and CNN consistently outperform other models, including baselines, with 5– 15% higher accuracy. Hossain and Mahmood [38] predict the solar PV power for the next day using an LSTM model. Their evaluation shows that LSTM is more accurate compared to Recurrent NNs (RNNs), Extreme Learning Machine (ELM), and NNs models. Huang and Kuo [34] introduce a CNN-based forecasting system (PVPNet). According to their study, CNN performs best in terms of accuracy and training time compared to SVR, NNs, RF, and LSTM. Pan et al. [40] propose a deep learning model based on Gated Recurrent Unit (GRU) NNs with attention mechanism and kernel density estimation for interval forecasting of solar PV power prediction. Their model provides the best forecast compared to other models, LSTM, GRU without attention, MLP, and persistence.

## 2.2. WT-based models for forecasting

WT are used effectively to preprocess and denoise the data [41–43] and integrate into a variety of time series prediction models to produce satisfactory results. For instance, WT with ARIMA as a prediction model is applied by Joo and Kim [44] to forecast eight different datasets, including sales, stock prices, house prices, electricity demand and others. Seasonal ARIMA and LSTM models are used with Discrete Wavelet Transform (DWT) for wind power forecasting [45]. Liu et al. [46] apply both ARIMA and NNs to predict the WT components for wind speed time series data. Another model based on NNs and WT is also demonstrated by Liu et al. [10]. Likewise, NNs models for the decomposed data are developed by Khelil et al. [3] for wind speed forecasting and by Saraiva et al. [47] for daily streamflow forecasting and NNs-based methods coupled with SWT for estimating water quality profiles [41]. Similarly, Rana et al. [48, 49] use WT with NNs and other prediction models including LR and SVR for electricity load forecasting. Huang and Wang [50] integrate the wavelet approach with the stochastic recurrent neural network for global energy markets fluctuations forecasting.

Several studies such as [51–54] demonstrate the effectiveness of WT in predicting solar radiation. To decompose the raw solar irradiance sequence data into stable (low-frequency) and fluctuated components, Wang et al. [53] use the DWT with CNN and LSTM models. The decomposed data has higher regularity (e.g., less volatility and outliers) than the original data. For forecasting global-incident solar radiation, Deo et al. [52] combine DWT with SVM model. Similarly, Huang et al. [54] use an Elman NNs (ENNs) with decomposed solar irradiance data. Sharma et al. [51] utilise WT NNs (WNNs) for short-term solar irradiance forecasting. Similarly, Rodríguez et al. [9] use WT components with NNs for solar irradiation prediction.

The effectiveness of WT for modeling solar PV power time series data is also demonstrated in several recent studies (e.g., [8, 55, 56]). An SVR model with the decomposed data using DWT is developed by Eseye et al. [55] for one-hour ahead PV power forecasting. Another similar study [8] also uses the WT with CNN to decompose and forecast the solar PV power production data for day-ahead. Mishra et al. [56] constructs LSTM model that uses the decomposed data coefficients and their descriptive statistics for solar power forecasting. Yet, the application of data transformation to solar PV power forecasting has not been thoroughly examined thus far [22].

With exception of [10, 52, 56], all the other studies configure MM, at least one model for each reconstructed component, to compute the final forecast based on WT. The widely used applications of WT for forecasting applications can be summarised as follows. First apply the WT function on the original signal to produce the coefficients and then apply the inverse WT on the coefficients to produce the reconstructed components. A prediction model is trained separately for each component to produce the forecast for the reconstructed components. Lastly, the predictions of all the components are combined to find the prediction of the original time series. We mention this approach throughout the paper as the components-based approach and it utilises MM architecture.

To eliminate the number of configured models Liu et al. [10] use the PCA for the reconstructed components. Their approach is proven to be efficient and significantly reduce the elapsed time compared to using all the reconstructed components. However, its viability for the solar PV power data has not been analysed to date. Deo et al. [52] select the subgroup of the components based on the correlation coefficients. Mishra et al. [56] use the statistical characteristics of the WT coefficients in conjunction with other meteorological data as inputs to a deep learning prediction model. This model discovers that exploiting the statistical properties of the WT coefficients considerably improves prediction performance compared to the model without WT. However, the performance of this model in terms of time and prediction performance compared to the MM architecture is not discussed. To eliminate the time complexity wile using MM architecture to minimise the complexity of the prediction models, Almaghrabi et al.[8] use a simple LR model for forecasting the noise component.

For the optimal selection of $DL$ and $\psi$, some studies such as [44, 49] evaluate multiple settings to find the optimal setting. Others like Khelil et al. [3] and Liu et al. [57] use Genetic Algorithm (GA) to select the optimal settings. They report only the optimal selection. Yet there is no study for solar power forecasting which analyses the effect of the different selections.

Compared to the existing WT based models in the literature, this paper intends to eliminate the WT requirement for component reconstruction, as well as the necessity to configure and train multiple models. Motivated by prior work for other applications [10], this study will employ one model for solar PV power time series data. To avoid the necessity for component reconstruction, the WT coefficients will be employed in this investigation instead of components as suggested in [56]. The single model approach will ease both implementation and application of WT for time series prediction without sacrificing the prediction accuracy. Moreover, this paper analyses the different wavelet settings to understand the importance of the optimal settings.

# 3. Wavelet Process

WT is a powerful signal processing tool that can represent data in both the time and frequency domains at the same time. There are two types of WT: Continuous Wavelet Transform (CWT) and DWT. The wavelet representation of a signal $f(t)$ in terms of a $\psi$ is $\left\langle f(t), \psi_{s,\tau}(t) \right\rangle$, which can be expressed as Eq.(1) and Eq.(2):

$$CWT_f(s,\tau) = \frac{1}{\sqrt{s}} \int_{-\infty}^{\infty} f(t)\psi^*(\frac{t-\tau}{s})dt \qquad (1)$$

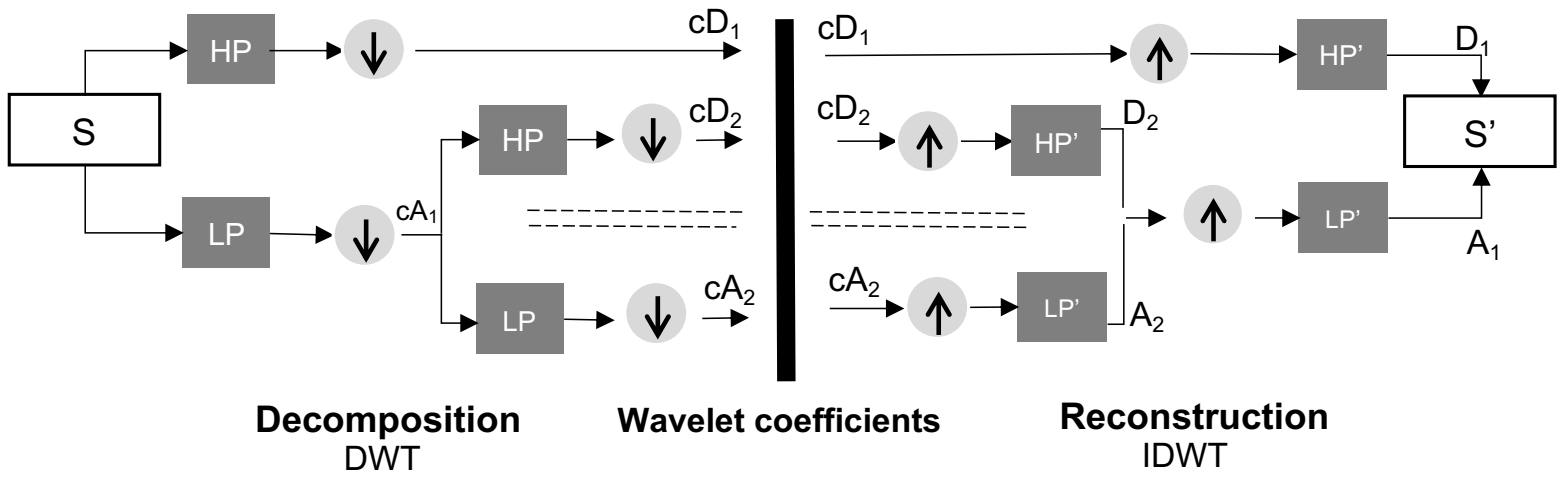

**Figure 2:** Multi-resolution analysis (MRA) using Mallat's algorithm [58]

$$DWT_f(s,\tau) = \frac{1}{\sqrt{|2^{-j}|}} \int_{-\infty}^{+\infty} f(t)\psi^*(\frac{t - k/2^j}{1/2^j})dt \quad (2)$$

where $s$ is the scale parameter, $\tau$ is the translation parameter, $*$ denotes complex conjugation, $j$ and $k \in \mathbb{Z}$, for $DWT_f(s,\tau)$, $s = 2^j$ and $\tau = k2^j$.

CWT calculates wavelet coefficients at every possible scale and location (i.e., enables an infinite set of scales and locations), which generates a redundant and large amount of data. On the other hand, DWT is defined at certain scales and locations (i.e., a finite set of scales and locations) [59]. The scale and translation parameters in DWT, unlike CWT, are based on powers of two (i.e., dyadic). DWT is a faster and simpler method of processing the time series data since it treats the data as discrete data rather than continuous data [49]. It generates a lesser number of coefficients which are required to reconstruct the original signal without information loss. As a result, it is more suitable to use for any time series data. Since DWT is only a discretized version of the CWT, it is accomplished by discretizing the parameters $s$ and $\tau$ using the translation coefficient $j$ and the scaling coefficient $k$.

DWT (also known as Multi Resolution Analysis (MRA)) helps to detect patterns not obvious in the raw signal [58], using a pair of quadrature mirror filters, known as low-pass (LP) and high-pass (HP) filters. These filter values are specified by selecting the $\psi$ and they must have equals length.

Figure 2 presents a summary of MRA based on Mallat's algorithm [58], with $DL = 2$ as an example. For a discrete signal $S$, the first step is to convolve the signal using the LP and HP filters. The resulting signal is then downsampled by 2 via dyadic decimation to yield two sets of coefficients: Approximation Coefficients at $DL = 1$ ($cA_1$) and Detail Coefficients at $DL = 1$ ($cD_1$). The entire process is carried out for $cA_1$ to produce Approximation Coefficients at $DL = 2$ ($cA_2$) and Detail Coefficients at $DL = 2$ ($cD_2$), and iteratively to generate the approximation and detail coefficients for subsequent levels.

The approximation and detail components at each level can be reconstructed from the coefficients using IDWT. IDWT uses two different filters LP' and HP' that are the time-reversed versions of the LP and HP filters. The reconstruction starts with upsampling the signal by inserting zeros at even-indexed elements, then convolving the resulting signal with the reconstruction filters. This means $cA_2$ and $cD_2$ are up-sampled then convolved with LP' and HP' to produce Approximation Component at $DL = 2$ ($A_2$) and Detail Component at $DL = 2$ ($D_2$). The same process is applied to produce the Approximation Component at $DL = 1$ ($A_1$) and Detail Component at $DL = 1$ ($D_1$). The combination of the resulting components, e.g., $A_1$ and $D_1$ can be used to reconstruct the original signal as $S' = A_1 + D_1$. Approximation components depict the broad signal trend, whereas the detail components indicate erratic variations around the approximation.

DWT has a significant drawback - it is not shift invariant. This indicates that, for the signals $S$ and $S''$ where $S''$ is the shifted version of $S$, $DWT(S) \neq DWT(S'')$. The shift invariance property is important for time series applications. Thus, a modified version of DWT, known as $\varepsilon$-decimated DWT, to represent the SWT has been developed to overcome this drawback. The main idea is to compute and retain both the odd and even decimations at each level resulting in no loss of wavelet coefficients. In this paper, the SWT is used for solar PV power forecasting.

## 3.1. Stationary Wavelet Transform (SWT)

The DWT, as illustrated above may be considered as simply a convolution step followed by a decimation step that holds even indexed elements. The choice of the decimation part can be defined by $\varepsilon_{DL}$ where $\varepsilon_{DL} \in \{0 = (even), 1 = (odd)\}$. The SWT retains all the possible $\varepsilon$ and upsamples the filters. At each level, the coefficients are obtained using the signal and the upsampled filters without the downsampling step. Therefore, the resulting coefficient length is equal to the signal length $N$. The ISWT is based on averaging the inverses of each $\varepsilon-$decimated DWT to find the signal $S'$.

## 3.2. Border distortion

Decomposition of a finite length signal using SWT introduces border distortion around the boundaries of the signal. This is because of the nature of convolution process involved [48, 60]. The standard approach to minimizing border distortion is to extend the signal with extra samples at the boundaries. The widely used signal extension method includes symmetric padding (mirroring the points next to the boundary), smooth padding (repeating points next to the boundary) and periodic padding (periodically extending the signal). These methods generally work well for image processing applications but are not suitable for time series as they introduce discontinuity at the boundary [48].

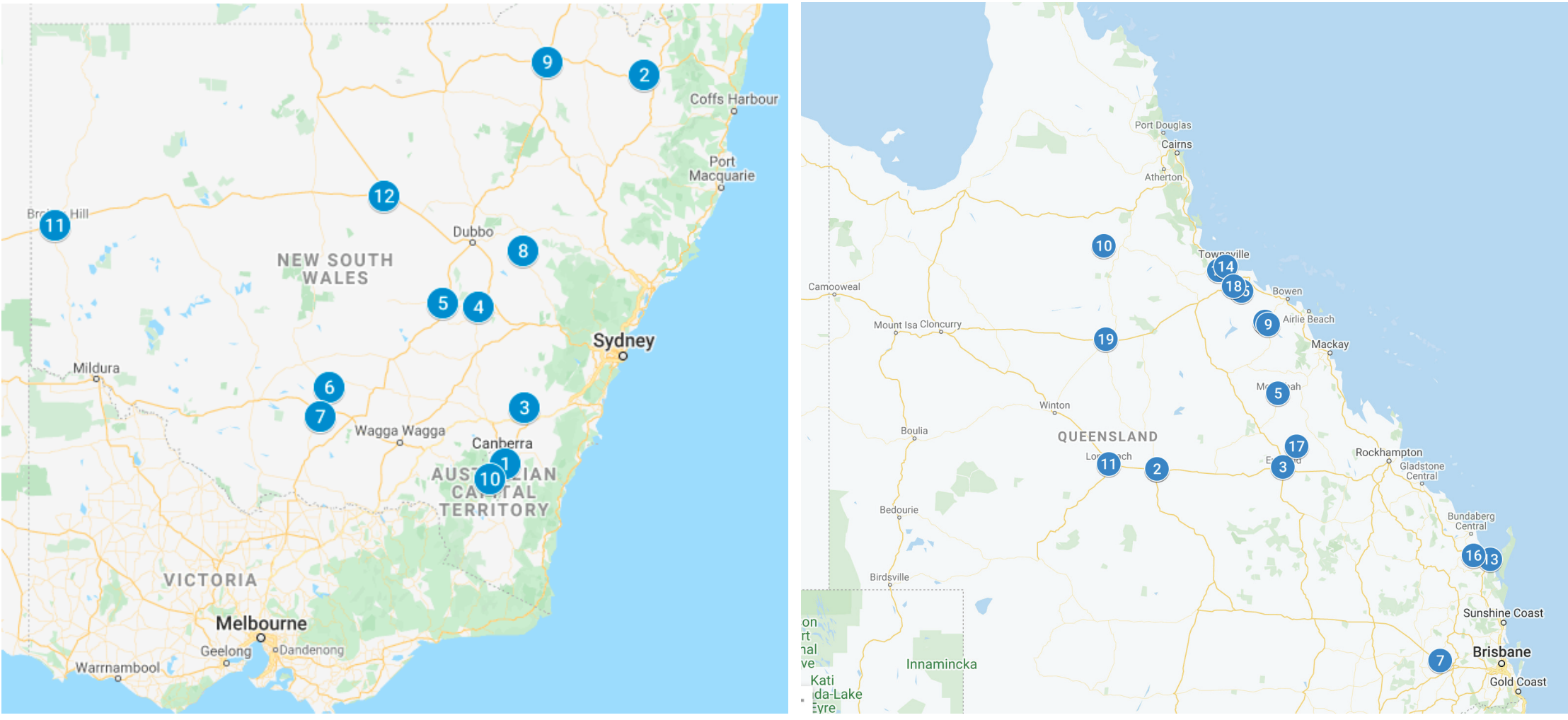

(a) The 12 solar PV farms in NSW (b) The 19 solar PV farms in QLD

**Figure 3:** Spatial distribution of the solar PV farms

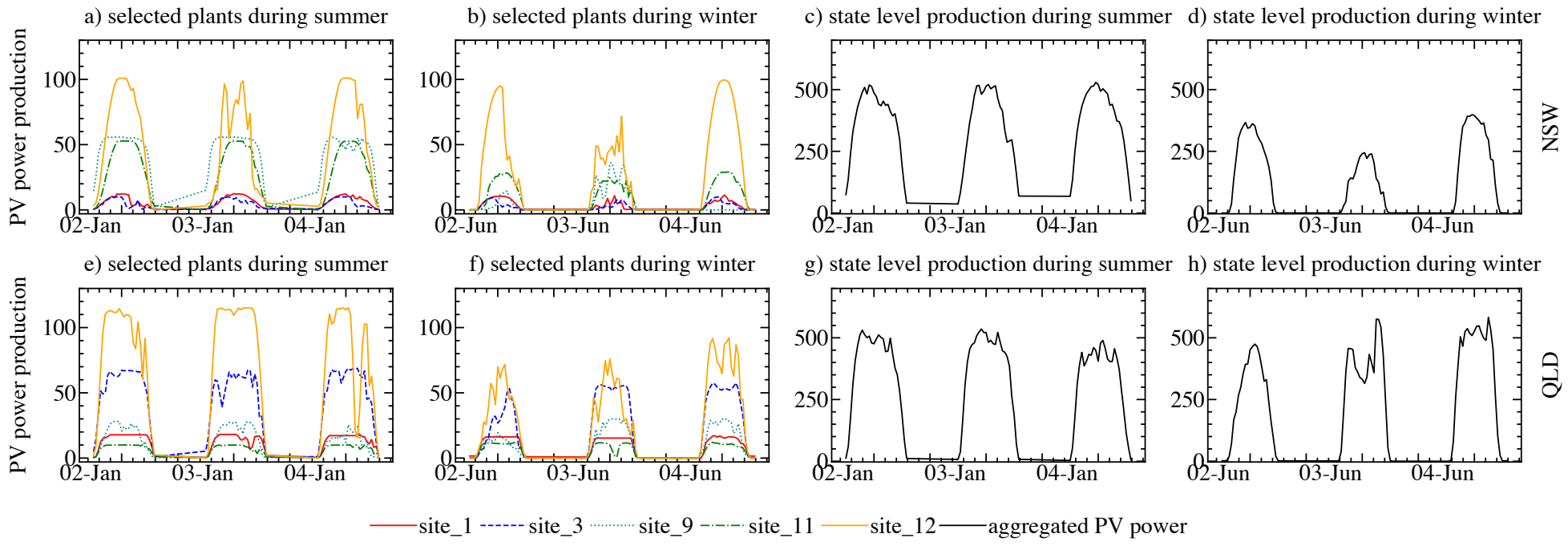

**Figure 4:** PV power production for three consecutive days during summer and winter

# 4. Problem Statement and Dataset

This section provides information about the research problem statement and describes the datasets used in the study.

## 4.1. Problem statement

**Given:** $X = [x^1, x^2, ..., x^D]$, which is an aggregated PV power time series, where $x^{d \in D} = [x^d_{t=1}, x^d_{t=2}, ..., x^d_{t=T}]$, T is the total number of steps per day, D is the total number of days in the dataset.

**Goal:** To investigate to what extent it is efficient (in terms of training time and prediction error) to use the WT as a pre-processing step to simultaneously forecast all the half-hourly power output (aggregated at regional level) for the next day $d + 1$, i.e., to predict all the elements of the vector $x^{d+1}$, which is represented as $[x^{d+1}_{t=1}, x^{d+1}_{t=2}, ..., x^{d+1}_{t=T}]$.

## 4.2. Datasets

The solar power data used in this study are collected from the grid-connected large scaled PV plants installed in the states of New South Wales (NSW) and Queensland (QLD), Australia. QLD was Australia's leading large-scale solar state in 2020 followed by NSW. The data are publicly available online [61]. Figure 3 shows the physical locations of the geographically dispersed PV sites in NSW and QLD. For NSW dataset, there are 12 PV plants with aggregated generation capacity of 600 MW while the capacity for each PV plant varies from 13 to 101 MW. For QLD, the aggregated generation capacity of 19 plants is about 1106 MW and the capacity for each PV plant varies from 11 to 147 MW. The samples in the dataset represent the state level aggregated solar PV power generated during daytime hours from 6:00 to 19:00 (the zero value observations outside this 13-hour window inclusive are not considered). Since the data is sampled at 30 minutes intervals, there are 27 observations for each day, i.e., $T = 27$. The data in both sets spans from 1/1/2019 to 8/5/2020, corresponding to 494 daily vectors, i.e., $D = 494$.

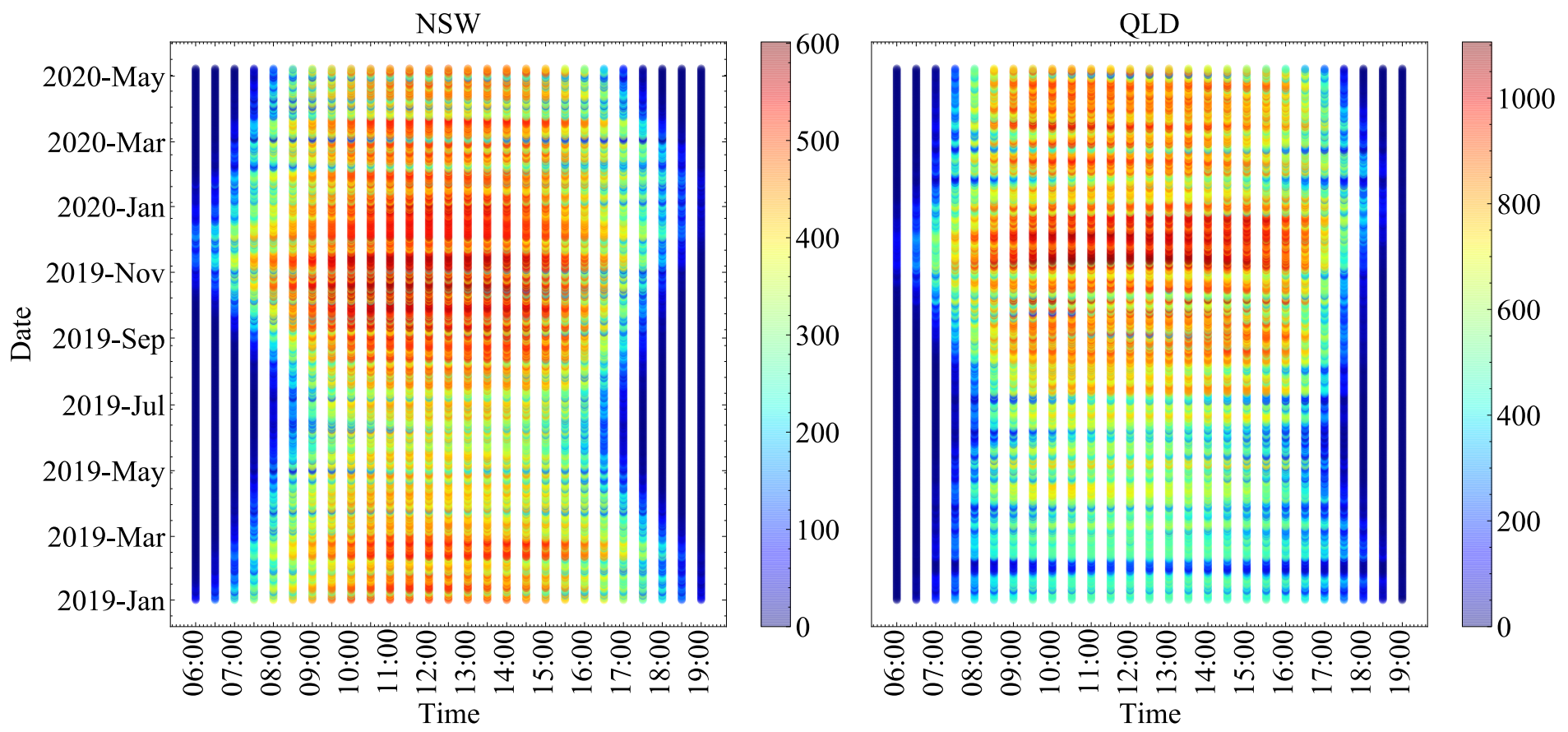

**Figure 5:** Heatmap of the aggregated PV power production temporal distribution for NSW (left) and QLD (right).

Solar PV power generation data is non-stationary, non-linear, and exhibits a high degree of fluctuation [17, 36, 54]. Weather conditions, which are very difficult to predict, contribute to driving these characteristics [62].

Figure 4 depicts the power generation for three consecutive days from two different seasons (mid summer: 2/1/2019 to 4/1/2019, and winter: 2/6/2019 to 4/6/2019) from randomly selected sites as well as the aggregated state level PV power for NSW and QLD. As we can see, data has both spatial and temporal differences. For spatial differences, the power production of the same day from site to site can be very different. For instance, in NSW, sites 1 and 12 exhibit different production levels as shown in Figures 4a and 4b. Similar observations can also be made for the QLD PV data. For example, the difference in production level for sites 3 and 11 is shown in Figures 4e and 4f. The data include multiple cyclic patterns (e.g., daily and yearly). For visualisation, Figure 5 displays the heatmap for the temporal distribution of the datasets. The daily solar radiance cycle causes the daily seasonality while climate variability causes the yearly seasonality.

Moreover, the power output from every site varies depending on the season. The variability is more during winter, as the cloud cover and other weather factors cause unstable power production (see Figure 4 for the selected locations). The PV power time series for the different sites are aggregated in order to minimise the variability. The aggregated PV power production during summer and winter days is shown by Figures 4c, 4d, 4g and 4h. It can be seen that the aggregated data has a better smooth line of production compared to the power output from the individual sites. However, there remain some noise and hidden patterns in the data that can be extracted using the WT.

To justify the use of the aggregated PV data, the historical volatility concept as discussed by [62, 63] is used to assess the aggregated and non-aggregated data volatility for each day (intra-day) and for the same hour in consecutive days (trans-day). Historical volatility ($\sigma_{h,T}$) is the standard deviation of the arithmetic or logarithmic returns over a time window $T$ as defined by [63]. To calculate the PV power data historical volatility, first, find the logarithmic return ($ret_{t,h}$) over the duration $h$ using Eq. 3, where $P_t$ is the PV production at time $t$:

$$ret_{t,h} = \ln\left(\frac{P_t}{P_{t-h}}\right) = \ln\left(P_t\right) - \ln\left(P_{t-h}\right) \tag{3}$$

Second, the historical volatility ($\sigma_{h,T}$) over the time window $T$ can be calculated as Eq. 4, where $\bar{ret}_{h,T}$ is the average values of $ret_{t,h}$ over window $T$ and $N$ is the number of observations. The overall volatility is the average value of all $\sigma_{h,T}$.

$$\sigma_{h,T} = \sqrt{\frac{\sum_1^N (ret_{t,h} - \bar{ret}_{h,T})^2}{N-1}} \tag{4}$$

Two cases intra-day and trans-day are evaluated. Intra-day ($\sigma_{1,27}$) volatility represents the average volatility experienced during the day $d$ from one step to the next, $T = 27$ as in this case there are 27 time steps for each day. Trans-day ($\sigma_{27,27}$) volatility represents the average volatility experienced during the same time step $t$ on consecutive days. The data of the first year is evaluated for volatility and the results are summarised in Tables 1 and 2.

Overall, the aggregated data demonstrates lower volatility (intra-day or trans-day) than the data from almost each individual site in both the NSW and QLD datasets. Because of this, it is more efficient when aggregated data are used to estimate PV power generation across a state.

## 5. Framework of the Coefficients-Based Approach

This section presents the proposed WT based approach for time series forecasting and explains its application for solar PV data. Figure 6 depicts the difference between the standard and proposed approach. Unlike the components-based approach, the coefficients-based approach applies the SWT on the original signal ($S$) to produce the wavelet coefficients and build a single prediction model which can forecast future values of the time series ($\bar{S}$) directly using the resulting coefficients as the inputs.

**Table 1**
Summary of NSW estimated intra-day and trans-day volatility for each site and for the aggregated PV power data

| | $site_1$ | $site_2$ | $site_3$ | $site_4$ | $site_5$ | $site_6$ | $site_7$ | $site_8$ | $site_9$ | $site_{10}$ | $site_{11}$ | $site_{12}$ | state |
|---|---|---|---|---|---|---|---|---|---|---|---|---|---|
| $\sigma_{1,27}$ | 8.42 | 19.92 | 22.03 | 21.8 | 18.09 | 18.27 | 22.93 | 15.4 | 20.82 | 22.21 | 23.77 | 22.43 | 9.35 |
| $\sigma_{27,27}$ | 12.0 | 9.45 | 13.85 | 11.23 | 5.92 | 8.62 | 10.51 | 5.84 | 6.07 | 10.08 | 13.52 | 7.28 | 7.39 |

**Table 2**
Summary of QLD estimated intra-day and trans-day volatility for each site and for the aggregated PV power data

| | $site_1$ | $site_2$ | $site_3$ | $site_4$ | $site_5$ | $site_6$ | $site_7$ | $site_8$ | $site_9$ | $site_{10}$ | $site_{11}$ | $site_{12}$ | $site_{13}$ | $site_{14}$ | $site_{15}$ | $site_{16}$ | $site_{17}$ | $site_{18}$ | $site_{19}$ | state |
|---|---|---|---|---|---|---|---|---|---|---|---|---|---|---|---|---|---|---|---|---|
| $\sigma_{1,27}$ | 11.34 | 29.2 | 24.66 | 27.14 | 14.64 | 28.16 | 15.67 | 30.13 | 16.82 | 24.41 | 13.57 | 27.53 | 15.3 | 28.3 | 26.13 | 16.12 | 21.86 | 16.89 | 20.7 | 10.34 |
| $\sigma_{27,27}$ | 16.28 | 18.66 | 9.50 | 19.32 | 7.01 | 15.26 | 11.99 | 21.04 | 11.27 | 14.62 | 16.68 | 11.26 | 14.67 | 17.60 | 13.48 | 9.50 | 13.23 | 11.16 | 10.29 | 8.58 |

The main advantages of the proposed approach include: i) it directly uses the coefficients as the inputs and hence there is no need to use the inverse WT to reconstruct the components, ii) it trains only one model and allows the model to jointly learn the decomposed data, and iii) it is more efficient in terms of time required for model configuration and training.

Moreover, the proposed approach provides the flexibility of developing the prediction model with either multiple channels (MC) or multiple inputs (MI), see Figure 6. The first implementation, MC, uses the input shape $(n_steps, n_coeff)$ where $n_steps$ is the number of steps in the input sequence, and $n_coeff$ is the number of coefficients obtained by applying the SWT. This implementation can be accomplished using any prediction model either statistical, machine learning (as a vector of multiple features) or deep learning models (as multiple channels). The second implementation (MI) is to leverage the neural network abilities to learn from the different representations of the data. It uses multiple input layers each with the shape $(n_steps, 1)$, where the number of input layers equals $n_coeff$. After using separate input layers to represent the coefficients, hidden layers are used to model the time series information and output layer to produce the predictions. Hence, MI can be utilised with neural networks prediction models only, but MC can be utilised with any regression algorithm. The acronyms MC and MI will be used throughout the paper to refer to the proposed coefficients-based approach.

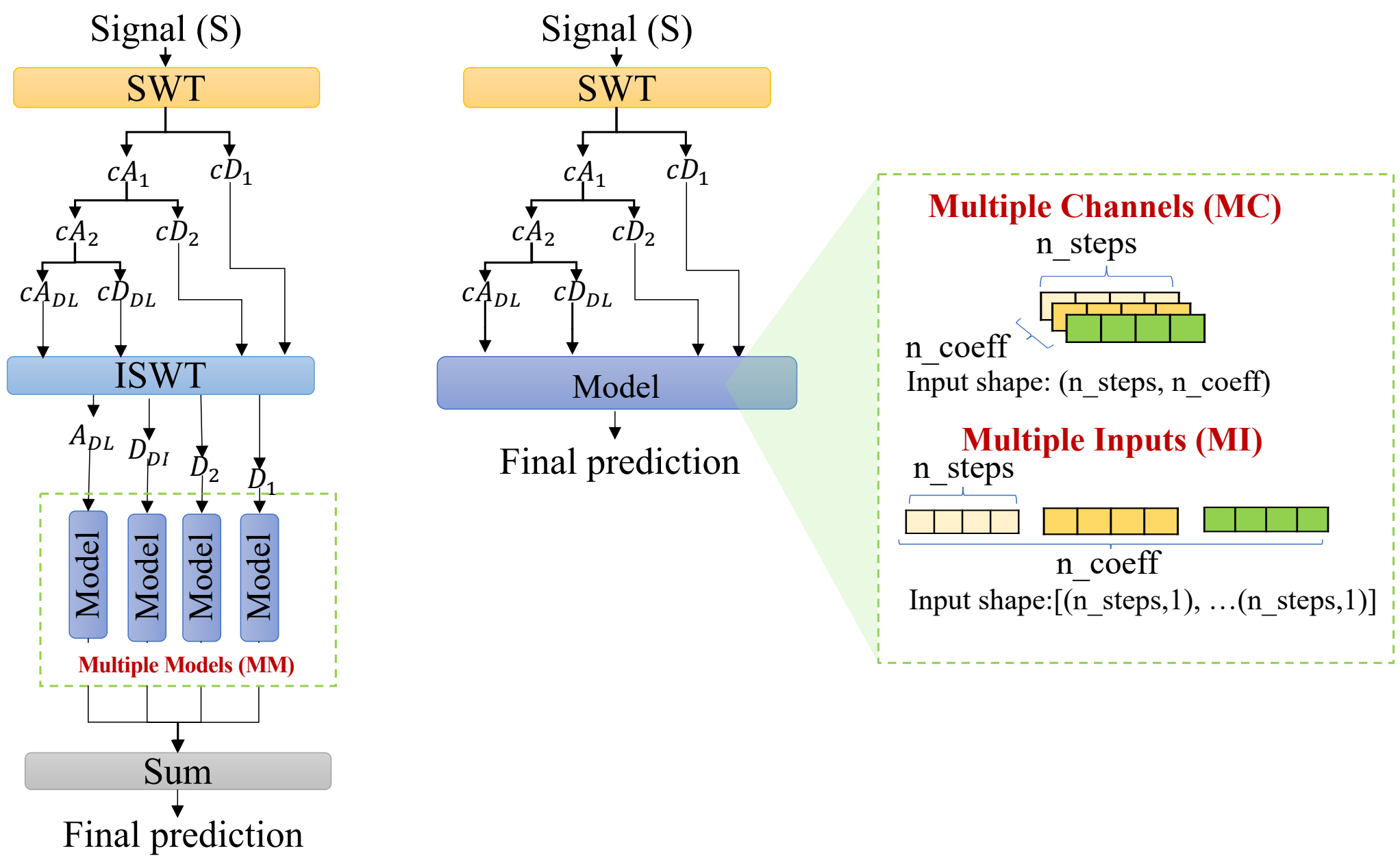

**Figure 6:** Data modeling using (left) multiple models with components-based approach, (right) multiple channels and multiple inputs with coefficients-based approach

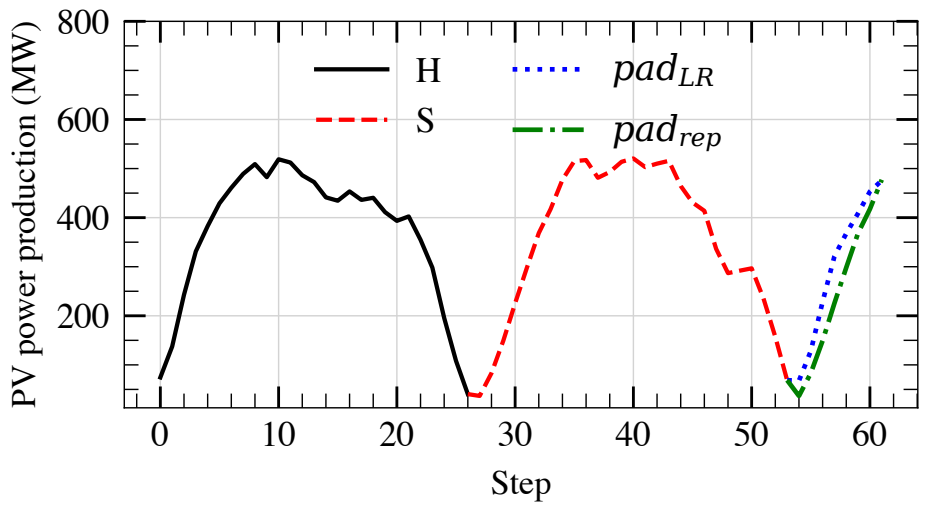

**Figure 7:** Visualisation for an example of a padded signal, using $pad_{LR}$ and $pad_{REP}$, where $len(H) = 27$, $len(S) = 27$, and the length of the predicted values on the right side is 8.

To minimise the distortion, two padding methods are employed in this study. The first method is based on Rana et al. [64] - padding the signal using the predicted values for the next day using simplistic LR as the prediction model. We refer to this method as Padding using Linear Regression ($pad_{LR}$). The second method is based on repeating the signal, we refer to this method as Padding using Repetition ($pad_{REP}$). This is motivated by the idea that the next day forecast could be similar to the day before. This method requires no training phase and hence has less computational cost. The length of the padded data (P) is affected by: the signal length $len(S)$, selected $\psi$ order and DL. P is calculated by Eq. 5 below:

$$P = len(H) + len(S) + F + 2^{DL-1} - 1 \quad (5)$$

where $H$ is the previous day's values used to pad the left side of the signal, $S$ is the historical daily values selected based on previous evaluations, and $F$ is the filter length of the selected $\psi$. For example, for 1 previous day values to pad the left side $len(H) = 27$, for lag of 1 day as historical values $len(S) = 27$, using $DL = 1$ and $\psi = db4$, hence $F = 8$, Thus $P = 62$. Figure 7 shows the signal padding based on both $pad_{LR}$ and $pad_{REP}$ as an example.

Figure 8 presents a schematic diagram of the proposed approach framework. It is divided into three major stages: 1) Data pre-processing, 2) Data modeling and 3) Prediction de-normalization.

## 5.1. Data pre-processing

To implement the proposed approach for forecasting aggregated solar power time series data, the data must be reshaped into a 2D matrix as a first step, where $D$ is the number of days in the data and $T$ is the time steps per day. In this case, $D = 494$ and $T = 27$. The data are then split into training and testing sets with 30% of the training data being used for the model's validation. The training set consists of the data from the first year 2019 (365 days), while the testing set consists of the data from the second year, 2020 (129 days). If $pad_{LR}$ is used for padding, the first 14 days of the training set is used to train the padder model then incrementally the padder model is updated with every new day of observations.

The coefficients-based WT module applies the SWT on the extended signal to obtain the coefficients. It is important to mention that the WT is not applied on the complete dataset at once. This is to guarantee no data leakage from the future. Applying the WT on the complete dataset (before splitting the data) as a pre-processing step provides entirely unrealistic and extremely optimistic predictions. To mimic the real-world situation, the data are transformed on a daily basis.

The PyWavelets package [65] is used to apply the SWT. Besides the signal to be decomposed, the wavelet function and the DL are required to be adjusted. As there is no definitive method to find the optimal wavelet or DL, this study examines a wide range of possibilities. This also helps to evaluate the effect of the different approaches with the different selections. The wavelet function to be applied in this study is the $db$ wavelet family. It is a common WT function, successfully used for time series data such as electricity load [44, 49], wind speed [3], solar radiance and PV power [52, 56]. The $db$ family has multiple orders, based on the number of vanishing moments (M). Often, it is written as $dbM$. For the purpose of this study, $dbM$ where $M \in [1, 7]$ and $DL \in [1, 4]$ are considered. The number of the coefficient sets generated using the coefficients-based wavelet module is $n_coeff$.

The coefficients are normalised to the range between 0.0 and 1.0. This is to enable prediction models to learn the data effectively. The normalization is conducted feature-wise based on Eq.(6), where $X$ is the time series data (coefficients or original signal), and $X_{min}$ and $X_{max}$ are the minimum and maximum values over the training data.

The normalized data then are fed to the prediction model to learn and find the next-day forecast.

$$X_{norm} = \frac{X - X_{min}}{X_{max} - X_{min}} \quad (6)$$

## 5.2. Data modeling

Multiple prediction models from different paradigms are employed to examine the proposed approach. These models include: LR [66], SVR [67] RF [68] and CNN [69]. These models are often used for solar power predictions and found to be powerful. All models are used to implement the MC approach. However, CNN is only used to implement the Multiple Inputs (MI) approach. This is because, as previously established, MI is only applicable to neural network based models.

To solve this Multiple Input Multiple Output (MIMO) prediction problem using LR, SVR, and RF, for each time step $t$ in the forecast horizon, a separate prediction model is used. The Keras [70] and Tensorflow [71] libraries of Python are used to implement the CNN model, while the scikit-learn [72] library is used to develop other models.

All the prediction models require tuning of several hyper-parameters. However, for ease of implementation the parameters for the models have been selected based on previous studies [8] utilising the same datasets. The parameters used for the SVR models are: kernel=either linear or rbf, the tolerance=0.0001, C=10 and epsilon=0.01. The RF models are constructed using either 500 or 1000 estimators with bootstrapping and MSE criterion.

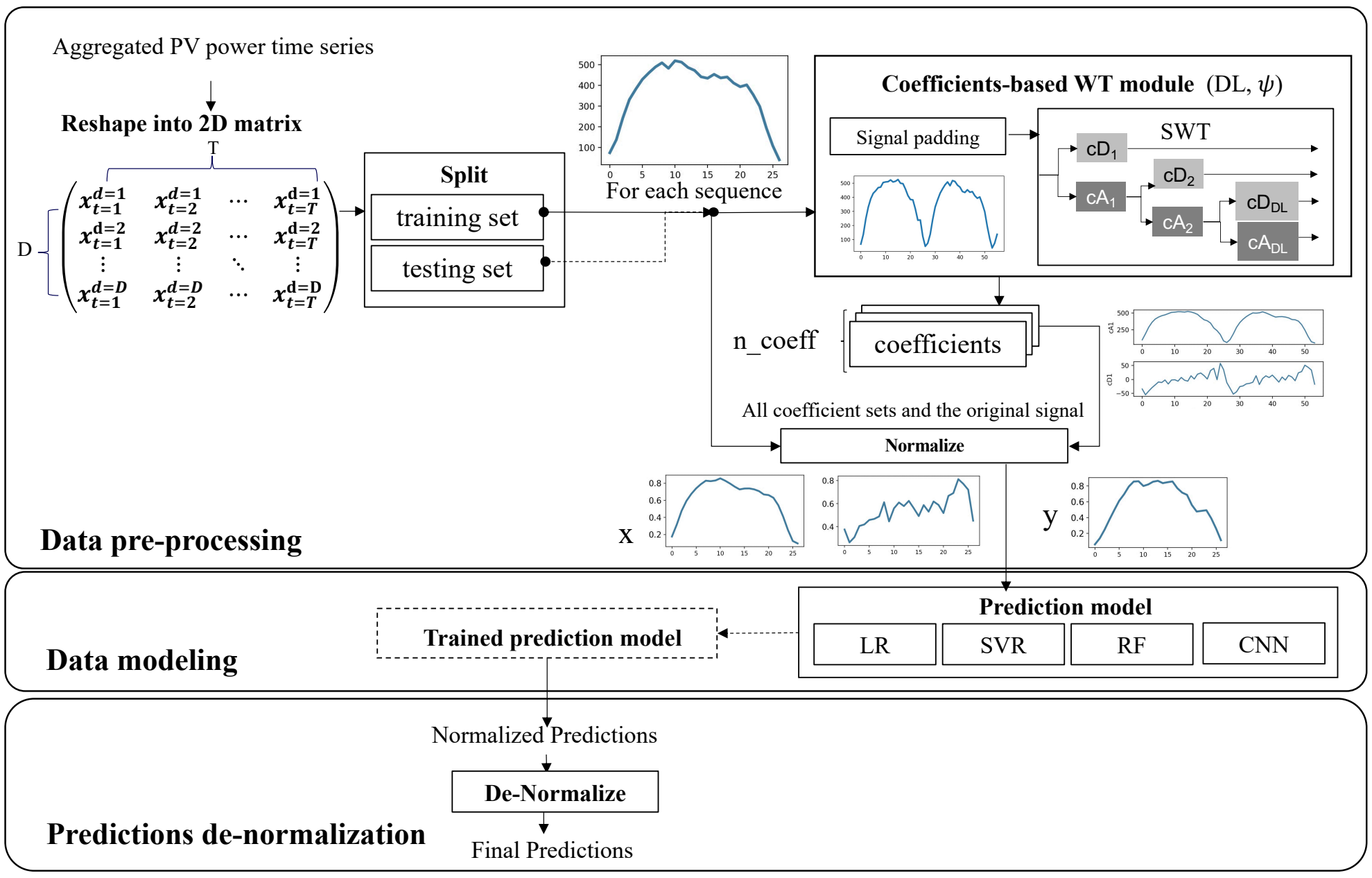

**Figure 8:** The framework of the coefficients-based approach pipeline

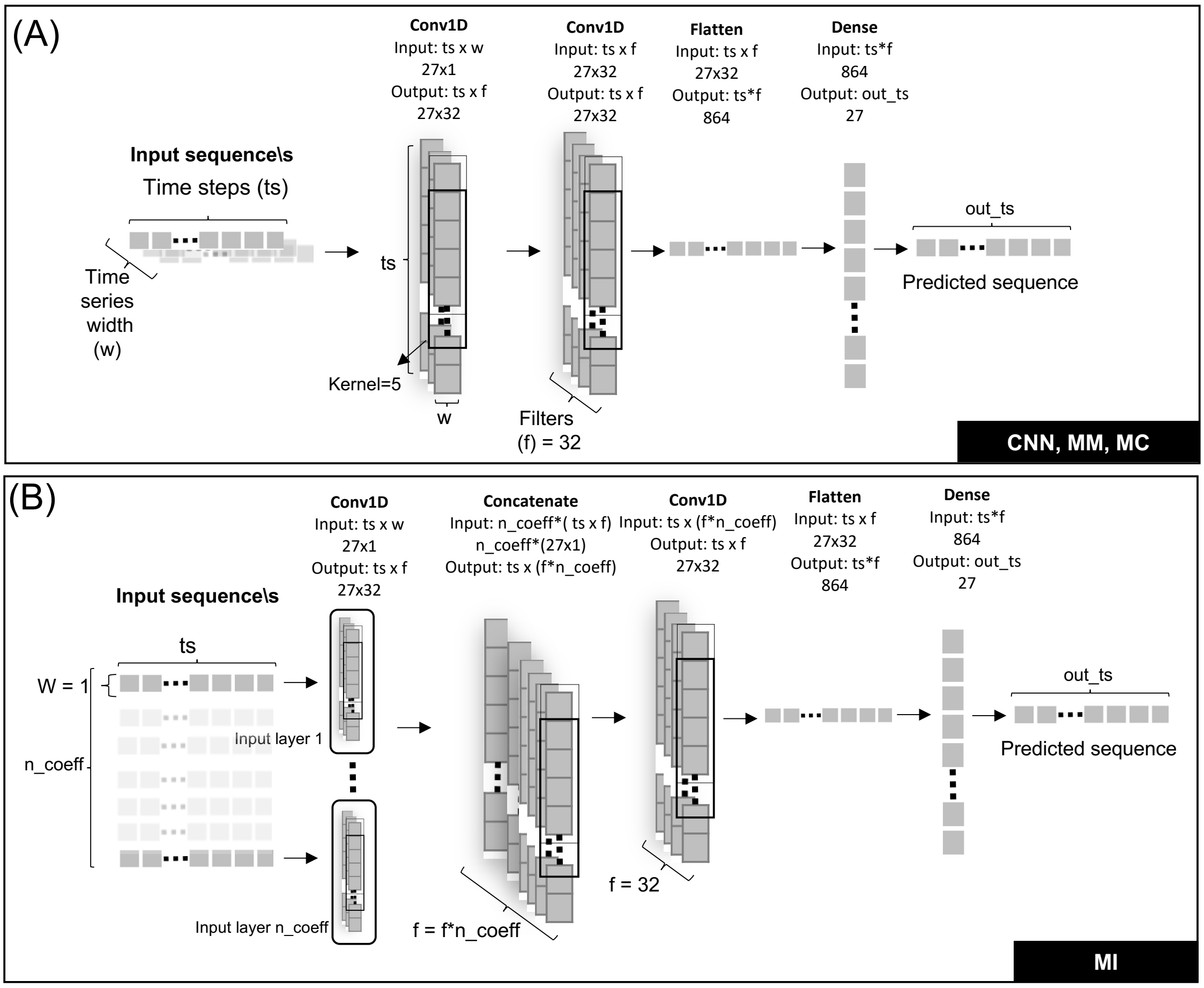

**Figure 9:** CNN model architecture

The CNN model architecture contains two convolutional layers, each with 32 filters ($f$), with kernel size equal to 5, same paddings and 1 stride. All the convolutional layers are trained using the ReLU activation function, followed by a flatten layer and a fully connected layer with 27 neurons, which is equal to the number of time steps ($out_ts$) to be predicted. The output layer uses the linear activation function to produce the predicted values. The different layer weights are initialized using glorot_uniform initialization [73]. The ADAM [74] optimizer is used for updating the network weights. For efficient training, early stopping with 200 epochs and patience equal to 20 with call backs to save the best weights while training are used. All the convolutional layers use the same number of $f$ which is 32. This number is selected after evaluation with $f \in \{32, 64, 256\}$. Figure 9(A) shows the CNN architectures applied to evaluate the application of CNN as a prediction model with different approaches (i.e., with and without WT). It is important to note that for MI implementation, the CNN architecture consists of multiple input layers equal to n_coeff which is the number of the coefficients in the series resulting from the SWT (see Figure 9(B)). A concatenated layer is used to combine the features learned from the different coefficients and the number of created features will be equal to f*n_coeff.

For all the models, the output, i.e. the forecasting horizon is 27 steps ahead, and the input sequence is the data of one historical day (27 steps). A 1-day window is used considering the daily pattern of solar PV power output following availability of solar irradiance.

### 5.3. De-normalization

The last step is to de-normalize the forecasts to be transformed back to the original range based on the equation Eq.7, where $Y_{de_norm}$ are the values of the final predictions, $Y_{norm}$ are the normalized predictions produced by the prediction models and $Y_{max}$ and $Y_{min}$ are the maximum and minimum values of power output in the training data.

$$Y_{de_norm} = Y_{norm} \times (Y_{max} - Y_{min}) + Y_{min} \tag{7}$$

## 6. Benchmarks and Evaluation Metrics

This section discusses the benchmark models that will be used for comparison, as well as the evaluations metrics. The workstation used to conduct all the experiments is equipped with an Intel (R) Core(TM) i7-10700KF CPU @3.80 GHz 3.79GHz, NVIDIA GeForce RTX3070 and 32 GB RAM.

### 6.1. Benchmark models

- **The Persistence model** ($B_{pday}$): this is a standard benchmark model used for solar PV power forecasting [12, 37, 51, 53], it assumes similar weather for consecutive days. It considers the solar power output from the previous day as the next day forecast.
- **Models without SWT**: these imply the use of the prediction models LR, SVR, RF and CNN with the original data i.e., without taking into account the use of WT.

### 6.2. Evaluation metrics

To evaluate the effectiveness of the proposed approach, this study utilizes both relative as well as absolute metrics that are widely used in solar energy forecasting literature. The metrics include Mean Absolute Error (MAE), Mean Relative Error (MRE), Root Mean Squared Error (RMSE), Relative Absolute Error (RAE), Root Relative Squared Error (RRSE) and Coefficient of Determination ($R^2$).

$$MAE = \frac{1}{D * N} \sum_{d=1}^{D} \sum_{t=1}^{N} |\hat{x}_t^d - x_t^d| \tag{8}$$

$$MRE = \frac{1}{D * N} \sum_{d=1}^{D} \sum_{t=1}^{N} |\frac{\hat{x}_t^d - x_t^d}{C}| \times 100\% \tag{9}$$

$$RMSE = \sqrt{\frac{1}{D * N} \Sigma_{d=1}^{D} \Sigma_{t=1}^{N} (\hat{x}_t^d - x_t^d)^2} \tag{10}$$

$$RAE = \frac{\sum_{d=1}^{D} \sum_{t=1}^{N} |\hat{x}_t^d - x_t^d|}{\sum_{d=1}^{D} \sum_{t=1}^{N} |\bar{x}_t^d - x_t^d|} \tag{11}$$

$$RRSE = \sqrt{\frac{\sum_{d=1}^{D} \sum_{t=1}^{N} \left(\hat{x}_t^d - x_t^d\right)^2}{\sum_{d=1}^{D} \sum_{t=1}^{N} \left(\bar{x}_t^d - x_t^d\right)^2}} \tag{12}$$

$$R^2 = \frac{\sum_{d=1}^{D} \sum_{t=1}^{N} \left(\bar{x}_t^d - \hat{x}_t^d\right)^2}{\sum_{d=1}^{D} \sum_{t=1}^{N} \left(\bar{x}_t^d - x_t^d\right)^2} \tag{13}$$

where $\hat{x}_t^d$ and $x_t^d$ are the predicted and actual power production for day $d$ and time $t$, $D$ is the number of days in the test data, $N$ is the number of observations per day, $C$ and $\bar{x}_t^d$ are maximum and average solar power output in the training data.

Additionally, we evaluate the improvement of one model over another as follows:

$$improvement = \frac{MAE_{M1} - MAE_{M2}}{MAE_{M1}} \times 100\% \tag{14}$$

where ($MAE_{M1}$) and ($MAE_{M2}$) are the MAE results of forecasting model without SWT and with SWT, respectively.

**Table 3**
Overall results for the different approaches

| Dataset | Model | MAE [MW] | RMSE [MW] | MRE [%] | RAE | RRSE | $R^2$ | Improvements [%] |
|---|---|---|---|---|---|---|---|---|
| NSW | $LR_{MM}$ | 54.164 | 76.246 | 9.007 | 0.347 | 0.428 | 0.817 | 4.092 |
| | $LR_{MC}$ | 55.772 | 78.267 | 9.274 | 0.357 | 0.439 | 0.807 | 1.404 |
| | $SVR_{MM}$ | 52.936 | 75.206 | 8.803 | 0.339 | 0.422 | 0.822 | 5.141 |
| | $SVR_{MC}$ | 54.000 | 77.331 | 8.980 | 0.345 | 0.434 | 0.812 | 2.251 |
| | $RF_{MM}$ | 53.132 | 76.502 | 8.835 | 0.340 | 0.429 | 0.816 | 0.289 |
| | $RF_{MC}$ | 53.227 | 76.175 | 8.851 | 0.341 | 0.427 | 0.817 | 0.719 |
| | $CNN_{MM}$ | 52.595 | 74.931 | 8.746 | 0.337 | 0.421 | 0.823 | 0.763 |
| | $CNN_{MC}$ | 53.472 | 76.355 | 8.892 | 0.342 | 0.429 | 0.816 | -1.117 |
| | $CNN_{MI}$ | 53.208 | 76.621 | 8.848 | 0.340 | 0.430 | 0.815 | -1.460 |
| QLD | $LR_{MM}$ | 86.903 | 121.875 | 7.857 | 0.308 | 0.376 | 0.858 | 6.293 |
| | $LR_{MC}$ | 92.132 | 127.522 | 8.33 | 0.327 | 0.394 | 0.845 | 1.586 |
| | $SVR_{MM}$ | 84.781 | 118.683 | 7.665 | 0.301 | 0.366 | 0.866 | 6.208 |
| | $SVR_{MC}$ | 86.886 | 120.741 | 7.855 | 0.308 | 0.373 | 0.861 | 4.398 |
| | $RF_{MM}$ | 89.395 | 124.558 | 8.082 | 0.317 | 0.385 | 0.852 | 0.216 |
| | $RF_{MC}$ | 89.288 | 124.338 | 8.072 | 0.317 | 0.384 | 0.853 | 0.394 |
| | $CNN_{MM}$ | 86.795 | 119.803 | 7.847 | 0.308 | 0.37 | 0.863 | 2.19 |
| | $CNN_{MC}$ | 90.124 | 123.944 | 8.148 | 0.32 | 0.383 | 0.854 | -1.224 |
| | $CNN_{MI}$ | 89.691 | 122.702 | 8.109 | 0.318 | 0.379 | 0.856 | -0.225 |

* The improvement shows the percentage improvement of models with SWT compared to models without SWT

## 7. Results and Analysis

This section presents and analyses the performance of the proposed approach for day-ahead forecasting of PV power. The various models will be denoted by the approach name and the prediction model that is employed. For instance, $LR_{MC}$ denotes the use of the multi-channel approach for WT, with the LR serving as the prediction model. Firstly, the overall performance of the proposed approach is discussed. Secondly, the performances of the two variants (MC and MI) of the proposed approach are compared to the utilisation of the components-based approach using multiple models (MM). The $LR_{MM}$, $SVR_{MM}$, $RF_{MM}$ and $CNN_{MM}$ are implemented based on the approach suggested by Almaghrabi et al. [8]. Then, the performance is compared to the previous studies that are based on hybrid DL models. After that, the performance of the proposed approach is compared to that of non-WT models including the baseline $B_{pday}$ and LR, SVR, RF, and CNN models. Next, the performance of the proposed approach is analysed at different granularity levels (hourly and monthly) to check the variations of the accuracy at different time and season. Finally, the sensitivities of the different approaches to the WT settings are discussed.

### 7.1. Overall performance evaluation

Table 3 summarises the average overall results of all the different prediction models with SWT that were created using the various approaches (MM, MC, and MI, as described in Section 5.2). The table shows the results using the optimal settings for each strategy. The improvement column shows the model improvement against the respective prediction models without using SWT.

For NSW, while utilising the MC technique, the $RF_{MC}$ model performs the best in terms of MAE, RMSE, and $R^2$, followed by $CNN_{MC}$, $SVR_{MC}$, and $LR_{MC}$. RF has been shown to be a powerful algorithm for forecasting solar energy in earlier studies, including [32, 75]. RF is a powerful ensemble model that generates prediction values by averaging the predictions of the ensemble decision trees [68]. For the second variation of the proposed approach, $CNN_{MI}$ model achieves predictions with less MAE than other models.

While utilising the MC technique, for QLD, the best models in terms of RMSE and $R^2$, are $SVR_{MC}$ followed by $CNN_{MC}$, $RF_{MC}$, and $LR_{MC}$. SVR is another common machine learning algorithm which is used for solar power forecasting and provides promising results in previous studies, such as [37]. Its main idea is to find an optimal hyperplane in the input space. It has the ability to learn non-linear relationships in the data.

The differences between $CNN_{MI}$ and $CNN_{MC}$, $SVR_{MC}$ and $RF_{MC}$ for both data sets are not statistically significant at $P <= 0.05$ (Wilcoxon rank-sum test). The $CNN_{MI}$ utilise the WT coefficients as multiple inputs to the hidden layers. The kernels of the first layer extract and learn the hidden patterns and information in each coefficient sequence separately and produce multiple filters for each input. Then the filters are concatenated and fed to another hidden convolutional layer. This convolutional layer will jointly learn the information in the multiple coefficients set and produce filters with the extracted information. The dense layer lastly produces the next day predictions. As the results show, the $CNN_{MI}$ provides a model that can learn the information in the coefficient set in two stages: the input layers learn the information of each coefficient sequence separately, and the hidden layers learn the information between the different coefficients jointly.

### 7.2. Performance comparison with the MM approach

This subsection evaluates the proposed approaches (MC and MI) to the MM architecture regarding prediction accuracy and time complexity.

#### 7.2.1. *Prediction accuracy*

As Table 3 shows, almost all prediction models perform better when the transformed data is used instead of the original data. The benefit of adopting the SWT in the non-stationary and non-linear solar PV power series is mostly accomplished with both approaches. The improvements with using the $LR_{MM}$, $SVR_{MM}$, $RF_{MM}$ and $CNN_{MM}$ are 4.092%, 5.141%, 0.289% and 0.763% for NSW and 6.293%, 6.208%, 0.216% and 2.190% for QLD. On the other hand, $LR_{MC}$, $SVR_{MC}$, and $RF_{MC}$ achieve improvements of 1.404%, 2.251% and 0.719% for NSW and 1.586%, 4.398%, and 0.394% for QLD. While utilising prediction models based on components resulted in more accurate predictions, the coefficients-based method also achieves similar improvements in almost all cases except for the CNN model. The reason for relatively lower accuracy of $CNN_{MC}$ and $CNN_{MI}$ compared to the CNN model without SWT could be not tuning the hyper-parameters - parameters from a previous

**Table 4**
Average computation time for training each model using different approaches

| Model | Approach | Time (Sec) | |
|---|---|---|---|
| | | NSW | QLD |
| LR | MC | 36 | 36 |
| | MM | 37 | 39 |
| SVR | MC | 51 | 62 |
| | MM | 65 | 59 |
| RF | MC | 1903 | 1875 |
| | MM | 1245 | 1306 |
| CNN | MM | 65 | 71 |
| | MC | 47 | 44 |
| | MI | 54 | 47 |

**Table 5**
Optimal SWT settings for different models

| Model | Approach | NSW | | | QLD | | |
|---|---|---|---|---|---|---|---|
| | | Padding method | DL | $\psi$ | Padding method | DL | $\psi$ |
| LR | MM | $pad_{REP}$ | 2 | db5 | $pad_{REP}$ | 3 | db7 |
| | MC | $pad_{LR}$ | 1 | db2 | $pad_{REP}$ | 2 | db1 |
| SVR | MM | $pad_{REP}$ | 4 | db3 | $pad_{REP}$ | 3 | db2 |
| | MC | $pad_{REP}$ | 1 | db6 | $pad_{REP}$ | 4 | db6 |
| RF | MM | $pad_{REP}$ | 1 | db2 | $pad_{REP}$ | 1 | db1 |
| | MC | $pad_{REP}$ | 1 | db1 | $pad_{REP}$ | 1 | db1 |
| CNN | MM | $pad_{REP}$ | 2 | db5 | $pad_{LR}$ | 2 | db4 |
| | MC | $pad_{REP}$ | 3 | db1 | $pad_{LR}$ | 1 | db2 |
| | MI | $pad_{REP}$ | 1 | db1 | $pad_{REP}$ | 2 | db3 |

study [8] are used directly without tuning them for ease of implementation.

The differences in accuracy between $LR_{MM}$ and $LR_{MC}$, $SVR_{MM}$ and $SVR_{MC}$, $CNN_{MM}$, $CNN_{MC}$ and $CNN_{MI}$ are statistically significant at $P <= 0.05$ (Wilcoxon rank-sum test). However, error difference between $RF_{MM}$ and $RF_{MC}$ is not statistically significant for any metric. Notably, the RMSE, RRSE, and $R^2$ values for $RF_{MC}$ are slightly better than those for $RF_{MM}$. The ensemble trees are able to provide similar performance with the decomposed data, either components-based or coefficients-based.

### 7.2.2. *Time complexity*

LR, SVR and RF require training a separate model for each time step in the prediction horizon. Therefore, for the MM approach, the number of models is affected by DL and $T$ ( $T = 27$ for our task). Hence, $LR_{MM}$, $SVR_{MM}$ and $RF_{MM}$ require $27x(DL+1)$ models, and $CNN_{MM}$ require DL models to deliver the day ahead predictions. The SWT application based on the coefficients approach contributes to mitigating this issue by reducing the required number of prediction models to a single model only.

Table 4 summarises the time (in seconds) required by each model. This includes the elapsed time required to decompose the data, train the model and provide the prediction. $LR_{MC}$ is more time-efficient with about 3% and 8% less time than $LR_{MM}$ for NSW and QLD, respectively. $SVR_{MC}$ is found to be more efficient than $SVR_{MM}$ with NSW but not QLD. That is because SVR is sensitive to the number of features. While the coefficients-based technique is based on the concept of using coefficients as features, it is found to be less efficient when used with SVR. A similar finding also holds for RF. $RF_{MM}$ is more efficient than $RF_{MC}$ with both datasets. From Table 5, it can be seen that the optimal setting for $RF_{MC}$ and $RF_{MM}$ is with $DL = 1$. This means there are two models configured for $RF_{MM}$ and one model with two features for $RF_{MC}$. Although more models are needed for $RF_{MM}$, the elapsed time is much less. This is because the detail component is implemented using a linear regression model, which reduces the overall time compared with building all the components using more complicated models such as RF, SVR, or CNN. As a result, for models that are sensitive to a large number of features, such as SVR and RF, the components-based approach is more efficient. $CNN_{MC}$ and $CNN_{MI}$ are more efficient than $CNN_{MM}$, using about 28% and 17% less computational time for NSW, and 38% and 34% lower training time for QLD, respectively. The CNN is capable of converging in a short period of time and has good performance. This is a promising result in terms of developing more efficient models utilising CNN with WT.

Figures 10 and 11 show the mean elapsed time over all the configurations for each model, for NSW and QLD datasets. As it is seen, the coefficients-based technique (MC or MI) requires less computational time than the components-based approach (MM). With the exception of SVR and RF, the elapsed time for the MC and MI is nearly constant for all models. When the $DL \in \{1, 2\}$ and SVR or RF are used as the prediction models, the MM technique shows a faster or comparable processing time. The MC technique, on the other hand, is much more efficient when the $DL > 2$. For LR and CNN models, the MC and MI approach have almost constant time requirement regardless of the DL. It is evident that using the coefficients technique speeds up all models when compared to the component approach. As previously discussed in section 1, it can be seen that as the DL grows, the time consumption of components-based models increases. However, the proposed coefficients-based models, particularly $LR_{MC}$, $CNN_{MC}$ and $CNN_{MI}$, are less sensitive to the DL.

Furthermore, Figures 10 and 11 show the effect of the padding method in terms of time. It can be seen that $pad_{REP}$ reduces the computational time for all models. For example, for NSW, when $DL = 1$, $LR_{MM}$ require approximately 50 seconds on average using $pad_{LR}$ as the method for padding. On the other hand, it requires around 38 seconds if $pad_{REP}$ is applied as the padding method. This is because $pad_{LR}$ requires a training phase and continuous updating with new data, but $pad_{REP}$ does not. The effect of the padding method chosen on the prediction error can be seen from Table 5. The majority of the models show the best performance using the $pad_{REP}$ method. Hence, it can be said that the proposed

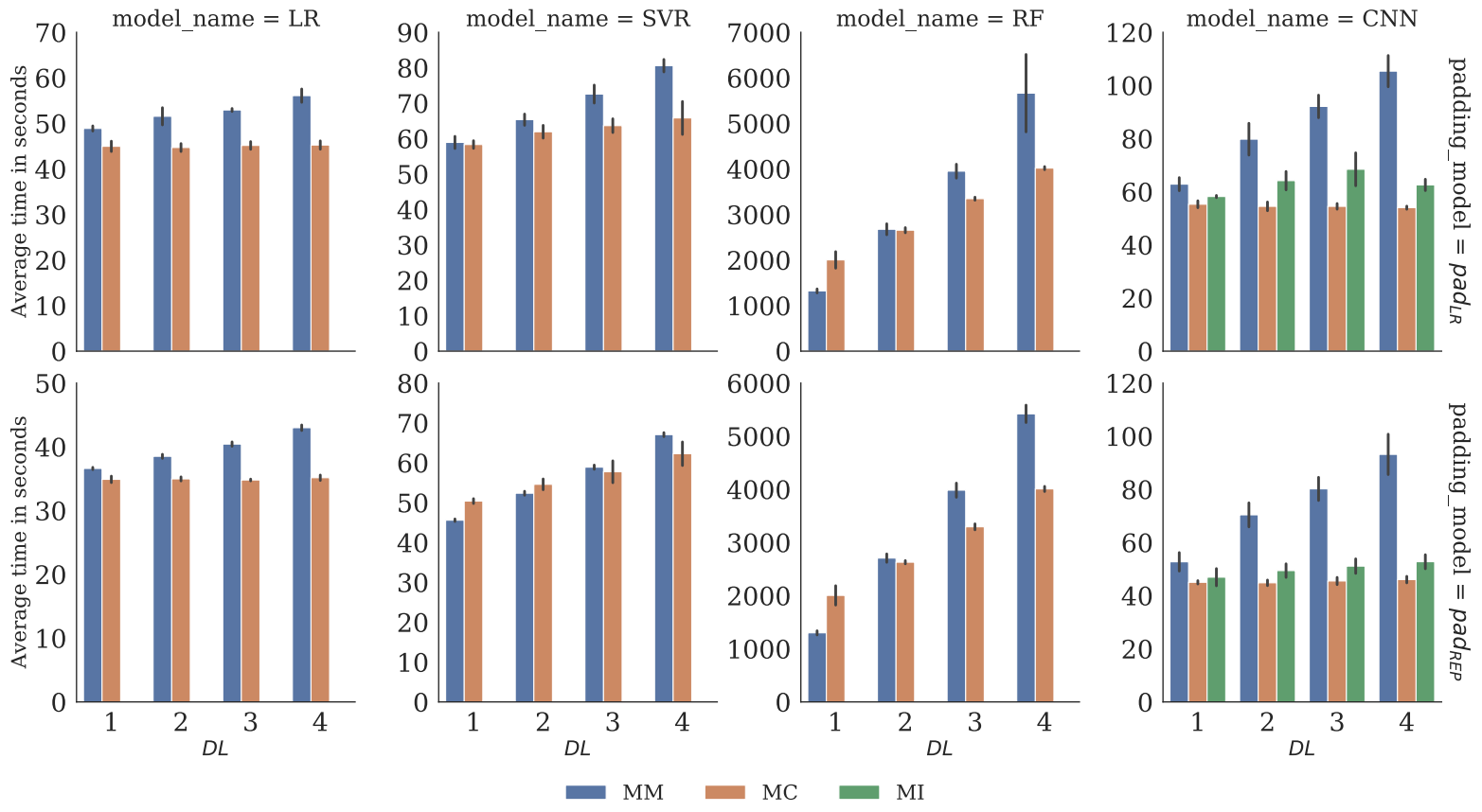

**Figure 10:** Average time spent by each model using different DL, with different padding methods for NSW

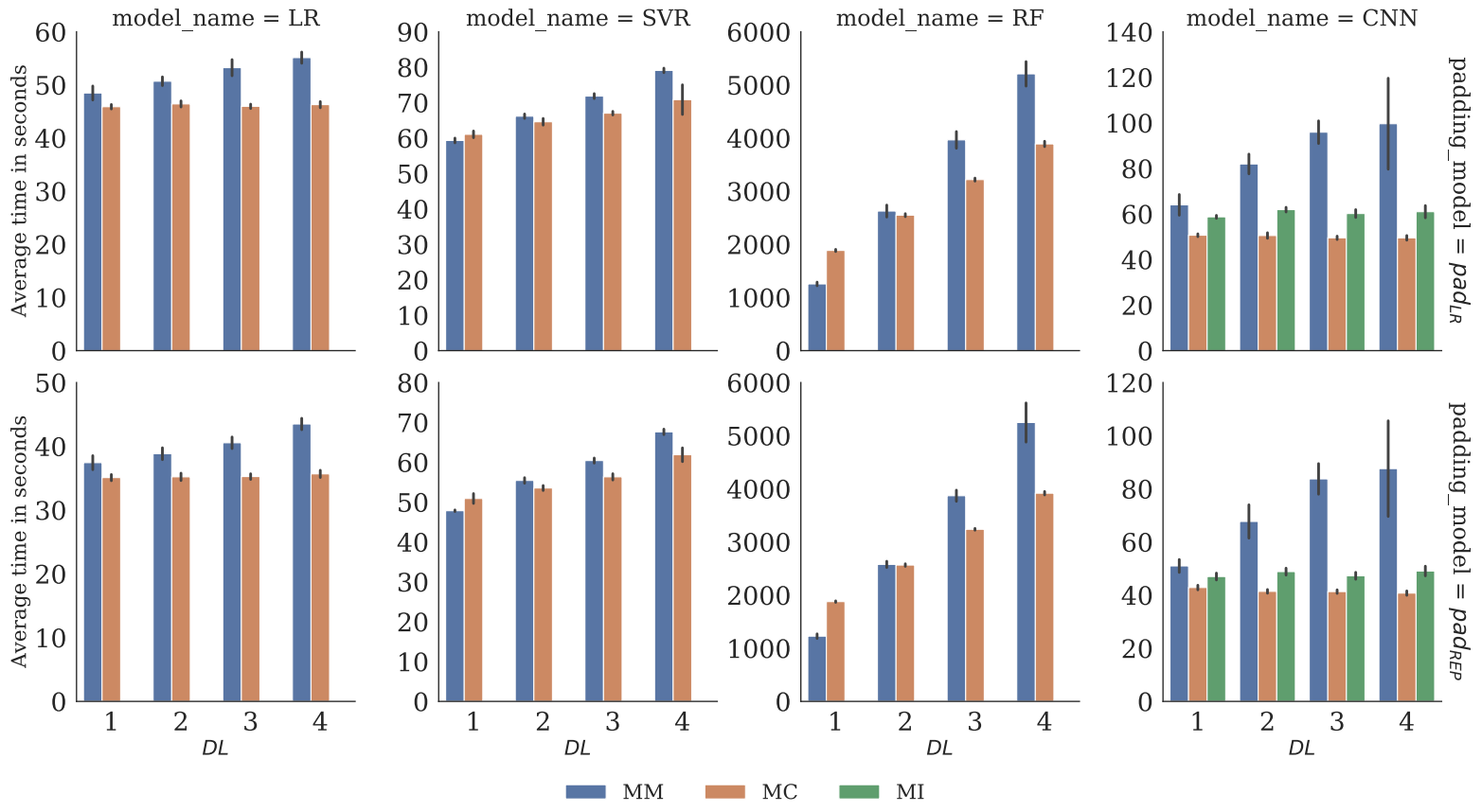

**Figure 11:** Average time spent by each model using different DL, with different padding methods for QLD

padding method $pad_{REP}$ for solar PV power time series with WT is more efficient.

Thus, the results indicate that there is a trade-off between time requirement and prediction accuracy when using the WT. It is feasible to produce forecasts with low error by the reconstructed components at the cost of increased time and model complexity, or to generate slightly less accurate forecasts with significantly lower training time and model complexity. It is critical to note that the time difference is not an issue in this case, given the training data spans only one year and the time difference is measured in seconds. Although it appears to be less critical, it is more important that these conclusions may be applied to other cases in which time is significant, such as for disaster prediction or in health applications.

### 7.3. Comparison with previous studies

The performance of the proposed model is compared with previous studies [8, 76] utilising WT and without WT, in terms of both accuracy and training time requirements. For comparison, [76] and [8] are selected since both studies used the same NSW data set as this study, and they represent hybrid DL model without WT [76] and with WT [76]. Table 6 presents the prediction accuracy and training time of two hybrid models and the proposed approach for NSW data. The first hybrid model CNNs-LSTM Encoder-Decoder (CLED) [76] employs the raw data as inputs. It achieves MAE of 52.257 MW. The time efficiency of the CLED is not reported. The second model, called WCNN [8], is based on using the reconstructed components, i.e. MM approach, and employs SWT with $DL=$ 1. To minimise the overall time complexity of the model, a CNN prediction model is used for the approximation component and an LR model for the details component. The MAE of WCNN is 52.476 MW and it requires about 174 seconds for training. The proposed implementation of $CNN_{MC}$ and $CNN_{MI}$ achieved a more efficient computation time of 47 seconds and 54 seconds for the MC and MI approaches, respectively; however, with slightly higher MAE 53.472 MW and 53.208 MW.

**Table 6**
Comparison with previous work

| Model | MAE [MW] | MRE [%] | Time [Sec] |
|---|---|---|---|
| CLED [76] | 52.257 | 8.690 | - |
| WCNN [8] | 52.476 | 8.726 | 174 |
| $CNN_{MC}$ | 53.472 | 8.892 | 47 |
| $CNN_{MI}$ | 53.208 | 8.848 | 54 |

**Table 7**
Overall results for the benchmark models

| Dataset | Model | MAE [MW] | RMSE [MW] | MRE [%] | RAE | RRSE | $R^2$ |
|---|---|---|---|---|---|---|---|
| NSW | $B_{pday}$ | 61.638 | 90.126 | 10.250 | 0.394 | 0.506 | 0.744 |
| | LR | 56.602 | 79.366 | 9.412 | 0.362 | 0.445 | 0.802 |
| | SVR | 54.932 | 79.072 | 9.134 | 0.351 | 0.444 | 0.803 |
| | RF | 53.853 | 76.723 | 8.955 | 0.345 | 0.431 | 0.815 |
| | CNN | 52.631 | 75.502 | 8.752 | 0.337 | 0.424 | 0.820 |
| QLD | $B_{pday}$ | 94.487 | 134.826 | 8.542 | 0.335 | 0.416 | 0.827 |
| | LR | 93.934 | 129.545 | 8.493 | 0.333 | 0.400 | 0.840 |
| | SVR | 90.694 | 126.051 | 8.200 | 0.322 | 0.389 | 0.849 |
| | RF | 89.756 | 124.827 | 8.115 | 0.318 | 0.385 | 0.851 |
| | CNN | 89.289 | 122.426 | 8.073 | 0.317 | 0.378 | 0.857 |

Despite the fact that the CLED MAE performance is approximately 1.787% better than the proposed approach, The time consumption of the CLED model remains unknown. It is known that LSTM computations are sequential, which means predictions for later timesteps must wait for their predecessors to finish before they can proceed. This could reduce the model's efficiency in terms of time, especially with long sequences. The WCNN model outperforms the proposed approach by around 1.375% in terms of MAE, but it requires approximately 70% more time. Another advantage of the coefficients-based approach, unlike WCNN, is that it does not require configuring more than one model to find the predictions. This comparison emphasises the trade-off between prediction accuracy and efficiency.

## 7.4. Comparison with benchmark models

Table 7 illustrates the overall results of the benchmark models. The $B_{pday}$ model for NSW and QLD achieved 61.638, 94.487 (MAE), 90.126, 134.826 (RMSE) and 0.744, 0.827 ($R^2$), respectively. All models including LR, SVR, RF, and CNN, outperform the baseline model. The results in terms of the RMSE values for the LR, SVR, RF, and CNN are 79.366, 79.072, 76.723, and 75.502 for NSW and 129.545, 126.051, 124.827, and 122.246 for QLD. Similar to the RMSE, all other metrics, including MAE, MRE, RAE, RRSE, and $R^2$, demonstrate improvements over the baseline model. Therefore, utilising the original time series, LR, SVR, RF, and CNN can provide PV power forecast with performance better than the baseline model.

The performance of MC and MI using the different models is significantly superior to the naive $B_{pday}$ model. The improvements are 13.2%, 14.2%, 15.5%, 15.3%, and 15.0% for NSW and 5.4%, 10.5%, 7.8%, 8.1% and 9.0% for QLD using $LR_{MC}$, $SVR_{MC}$, $RF_{MC}$, $CNN_{MC}$ and $CNN_{MI}$, respectively. For both datasets, CNN model provides the most accurate predictions. All models except $CNN_{MC}$ and $CNN_{MI}$ for both datasets perform better than the non-WT models.

## 7.5. Hourly and monthly accuracy

Figures 12 and 13 illustrate the breakdown of prediction performance by time step for the various models in terms of MAE, RMSE, and $R^2$ for two datasets. For both datasets, $LR_{MM}$ and $LR_{MC}$ provide fairly similar predicting performance at each time step between 7:00 and 8:30, and 16:30 to 18:30. However, when production is more volatile, such as between 9:00 and 16:00, $LR_{MM}$ performs better. Similarly, $SVR_{MM}$, also provides better accuracy compared to $SVR_{MC}$ between 9:00 and 16:30. On the other hand, $RF_{MM}$ and $RF_{MC}$ perform similarly for each time step in prediction horizon. Although the overall analysis demonstrates a significant difference in accuracy between $CNN_{MM}$, $CNN_{MC}$ and $CNN_{MI}$, their performances on each time step are almost same.

The hourly and monthly performance evaluations show that almost all models reveal certain times when both approaches perform equally well in terms of prediction error. Additionally, utilising the coefficients-based strategy shows less time cost. This may suggest that both approaches could be combined for time series prediction problems. In other words, it could be worthwhile to explore combining both approaches depending on the time of day. At times of the day when the PV power output is less volatile, the coefficients-based approach is more efficient, whereas, at other times of the day, the components-based approach is more powerful.

Figures 14 and 15 show the prediction performance (in terms of MAE, RMSE, and $R^2$) of different models by month. This analysis is important in order to comprehend the seasonal influence. According to the monthly accuracy results, in general, the models (either MC approach or MM approach) exhibit similar performance, especially for RF and CNN. However, for LR and SVR, they show slight differences. For the purpose of understanding the significance of the differences on a month-by-month basis, the results of the Wilcoxon rank-sum tests for each strategy for each month in the test set are shown in Figure 16. There are two matrices, one for each dataset. The rows represent the months while the columns represent the models. The matrices represent the insignificant assessment of the different prediction models based on the two implementations MC and MI compared to MM. Statistically insignificant results (the light green cells) indicate that utilising the coefficients-based approach can achieve similar results to the traditional components-based approach.

Figures A.1 and A.2 in A show the actual and predicted data for each model for the NSW and QLD datasets respectively. Only four days are presented due to space constraints. The figures demonstrate that models constructed with components or coefficients produce comparable forecasting values. Figure 17 shows the scatter plots of the actual vs predicted data from all models, for all samples in the test set. All models tend to have a linear relationship between the actual and predicted values with very similar scattering for the different approaches (MM, MC and MI).

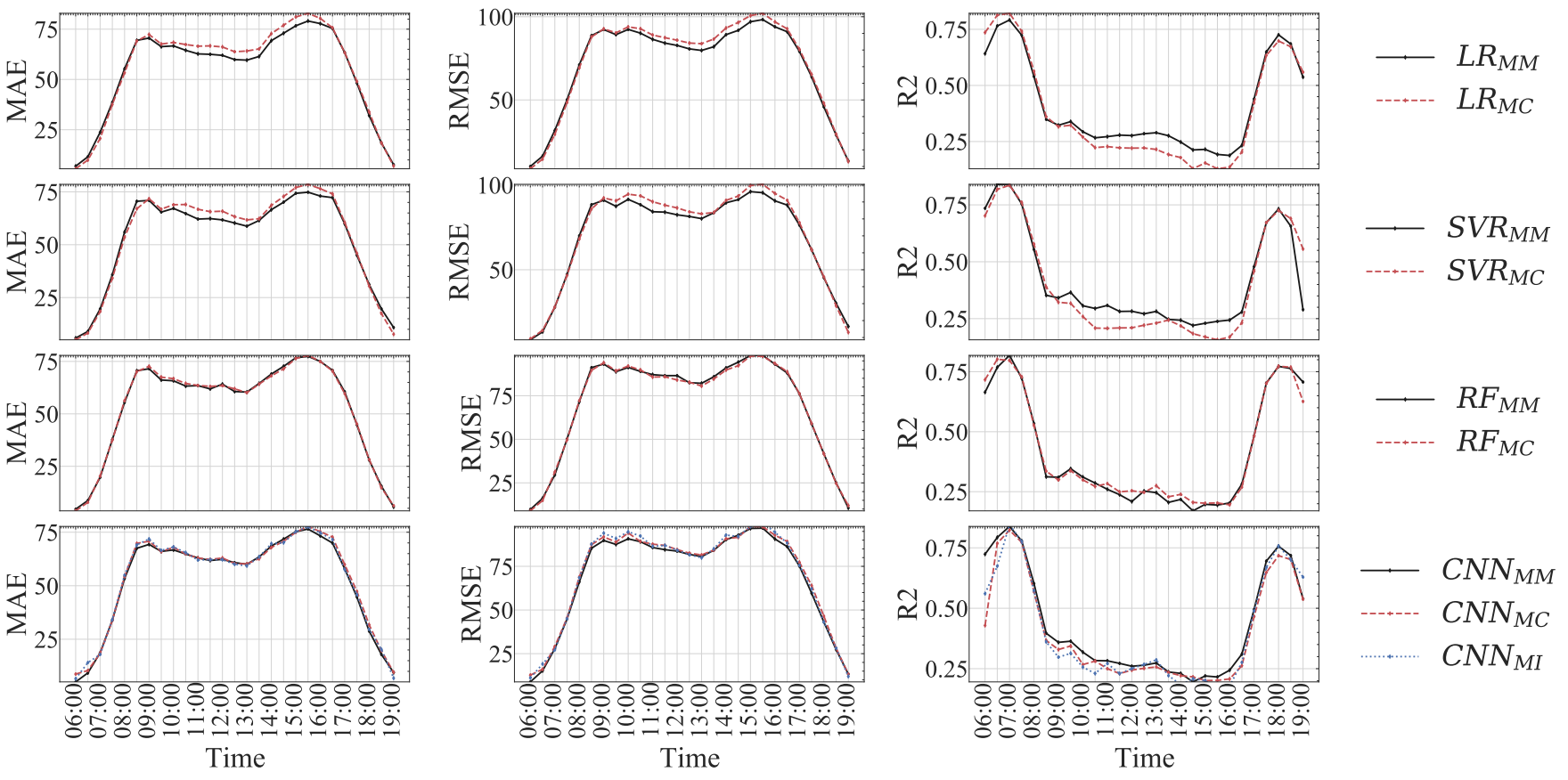

**Figure 12:** Components-based vs coefficients-based approaches performance for each time step in terms of MAE, RMSE and $R^2$ for NSW

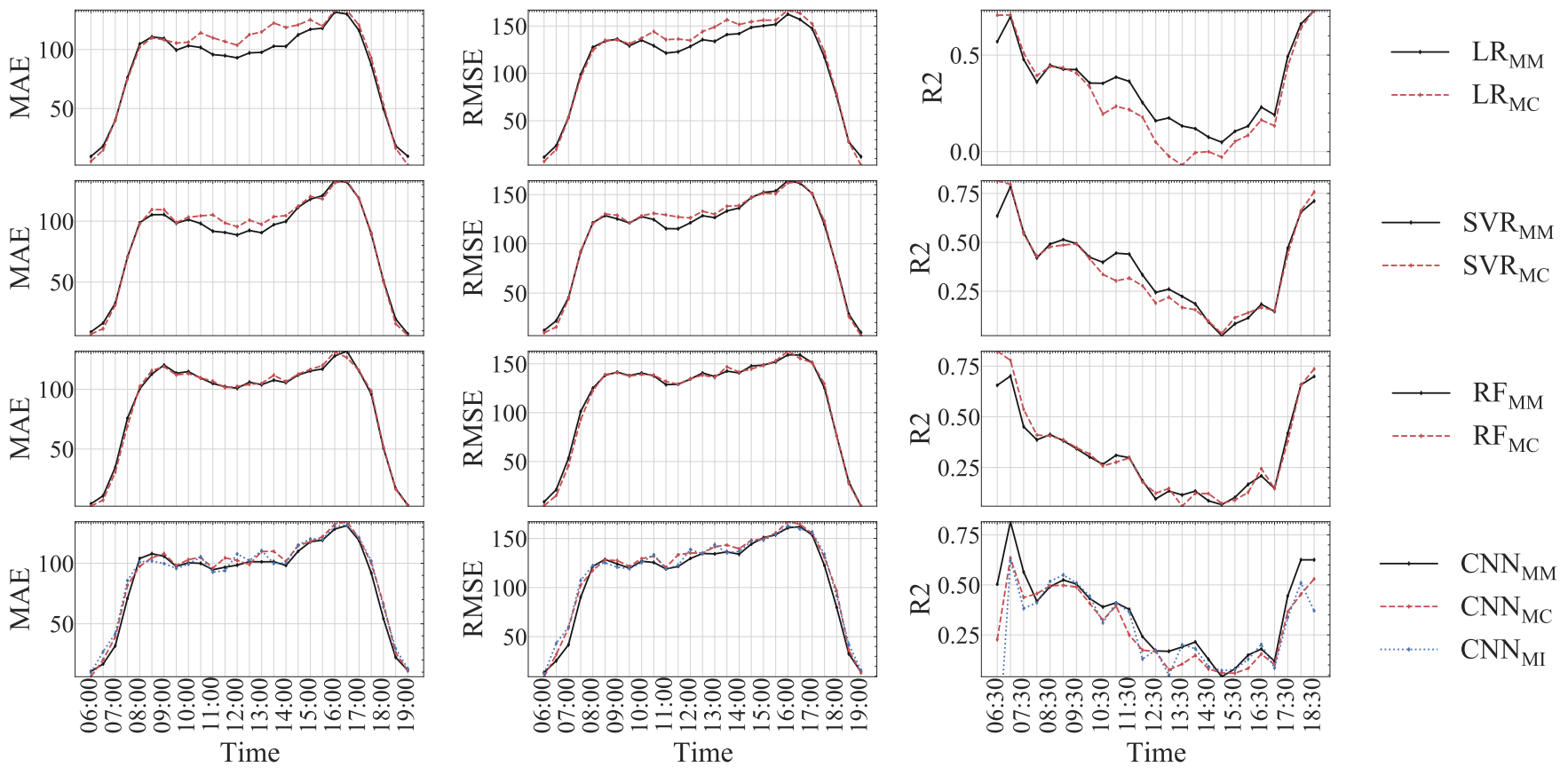

**Figure 13:** Components-based vs coefficients-based approaches performance for each time step in terms of MAE, RMSE and $R^2$ for QLD

## 7.6. Sensitivity to DL and $\psi$

The selection of DL and $\psi$ are important to ensure appropriate usage of the WT. As explained in Section 1, there are currently no standard definitive ways for providing such ideal settings. As a result, this study evaluates various combinations of $DLs$ and $\psi$. Tables B.1-B.8 in Appendix B show the performance of each prediction model (LR, SVR, RF and CNN) for different wavelet setting ($dbM$ where $M \in [1, 7]$ and $DL \in [1, 4]$) in each approach (MM, MC and MI) with two padding methods ($pad_{LR}$ and $pad_{REP}$).

The disparate performance of each configuration demonstrates the importance of determining the ideal parameter choices. Figure 18 provides a visualisation summary for the different models' sensitivity, in terms of MRE, to the different settings utilising the different approaches. Although all the approaches are affected by the selected WT settings, the MM approach is generally less sensitive to the different selections of DL and $\psi$ than the MC and MI approaches. That can be concluded from the span of the boxes of each approach.

For example, for NSW, the MRE span for $SVR_{MM}$ from about 8.898 to 9.065 and from 8.803 to 8.926 with the $pad_{LR}$ and $pad_{REP}$ respectively. Similarly, for QLD it spans from 7.674 to 8.072 and from 7.665 to 8.066 with the $pad_{LR}$ and $pad_{REP}$ respectively. The $MC_{RF}$ is less sensitive to the WT settings with NSW but not QLD. The reason could be because QLD dataset has more volatility than NSW. This means when the data is highly variable the choice of the WT settings is even more critical. The same reason could also be valid for the higher sensitivity of $CNN_{MC}$ and $CNN_{MI}$. Another reason, as mentioned before, the hyperparmeters of a previous study are used for both datasets without tuning process. It is expected that the sensitivity of $CNN_{MC}$ and $CNN_{MI}$ models may decrease if the hyperparameters are tuned for better fitting.

Figures 19 and 20 summarise the results of the Wilcoxon rank-sum test to study the pairwise difference in accuracy for different settings, for two datasets. It is evident that $RF_{MC}$ can provide a similar solar PV power production forecast as $RF_{MM}$ at almost each DL for NSW and QLD. Other models exhibited minor differences at only a few $DLs$.

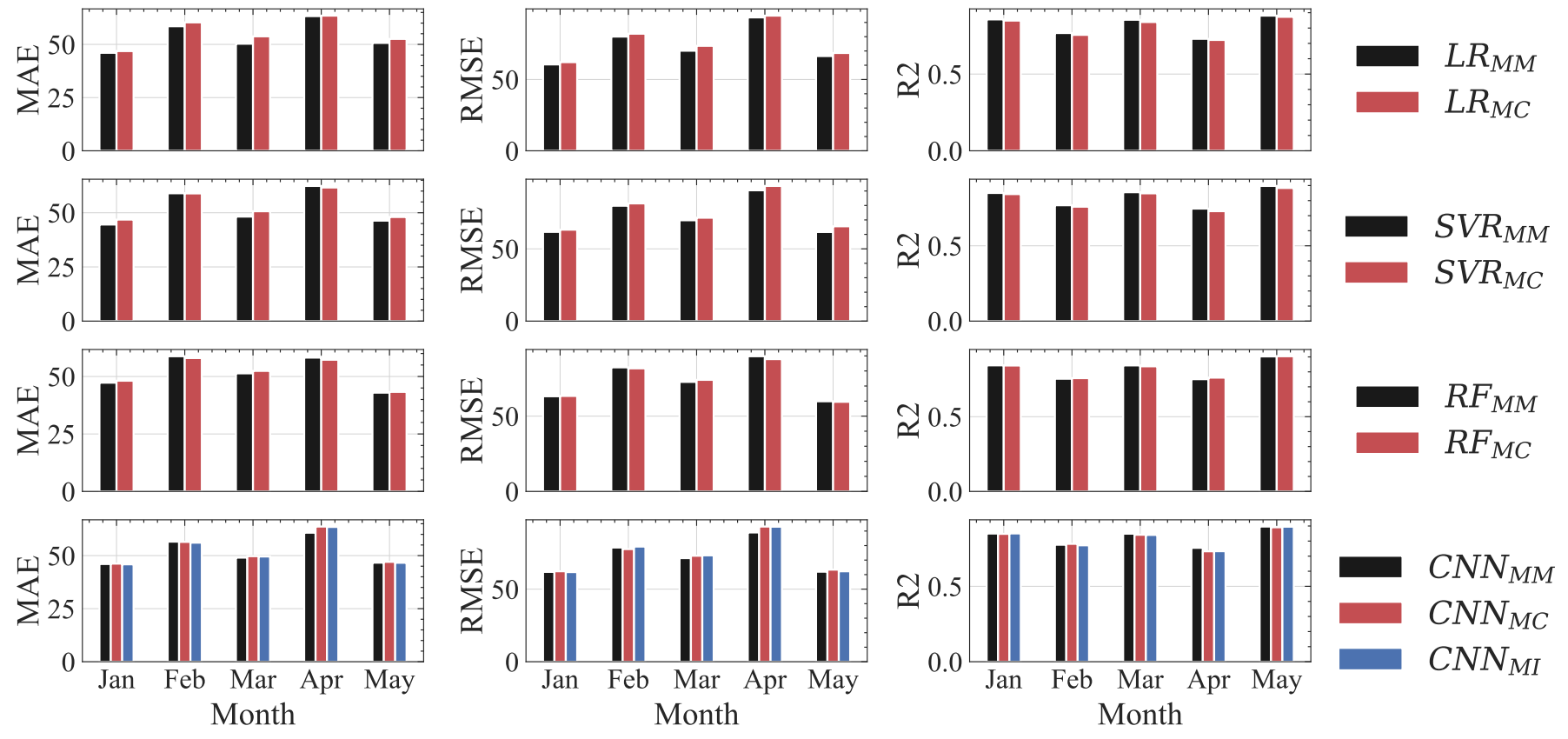

**Figure 14:** Components-based vs coefficients-based approaches performance for each month in terms of MAE, RMSE and $R^2$ for NSW

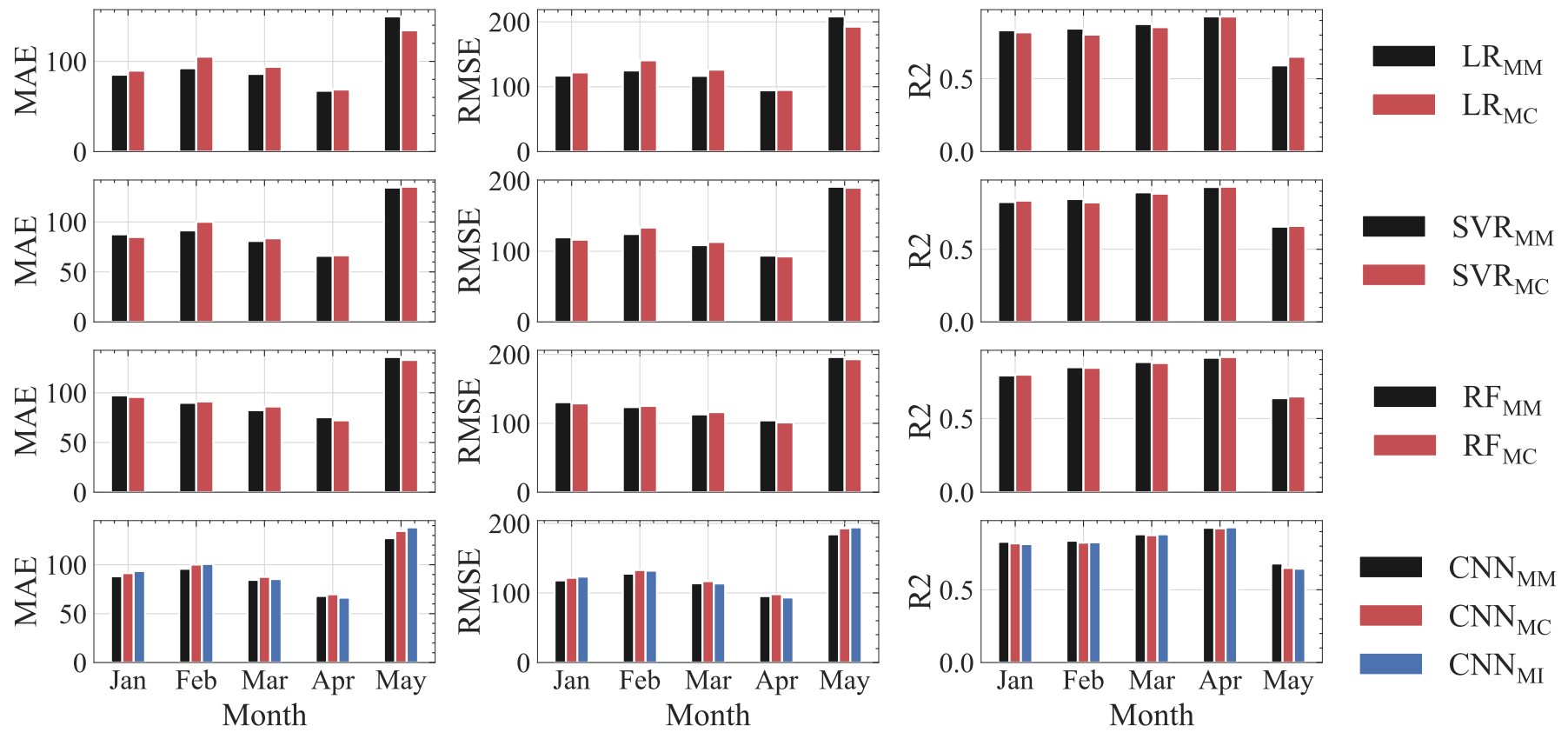

**Figure 15:** Components-based vs coefficients-based approaches performance for each month in terms of MAE, RMSE and $R^2$ for QLD

For all datasets $CNN_{MI}$ provides similar performance as $CNN_{MM}$ in most cases when the $DL = 4$ and $pad_{REP}$ is used. Similarly, $SVR_{MC}$ with $DL = 1$ provides similar performance as $SVR_{MM}$ for all datasets.

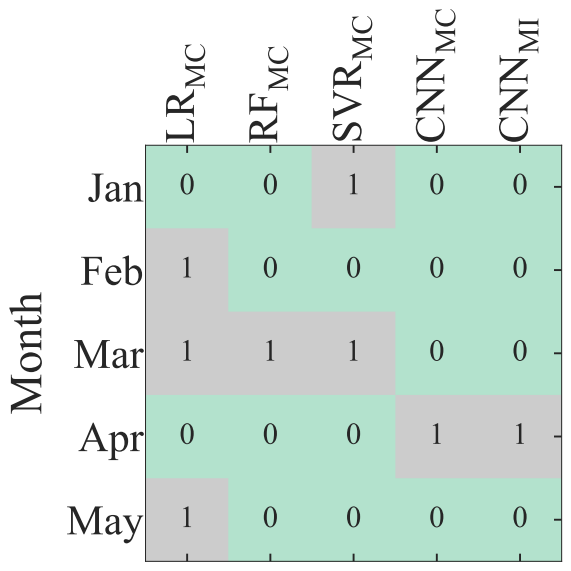

(a) NSW

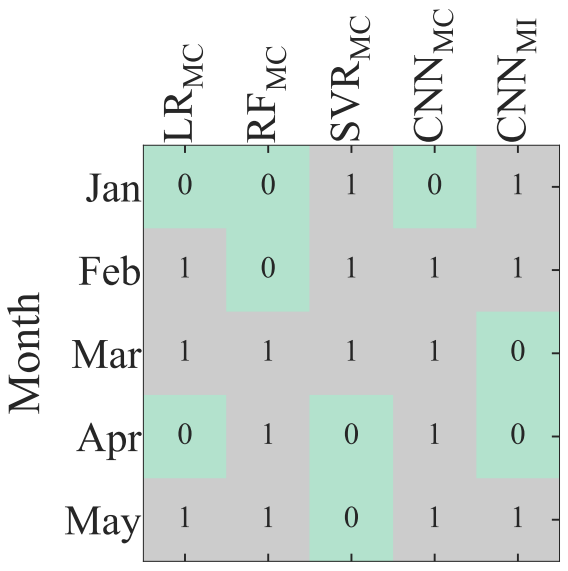

(b) QLD

**Figure 16:** Wilcoxon rank-sum test showing the statistical insignificance of differences in accuracy at $P <= 0.05$ for (MM, MC, and MI) during the different months.

In summary, the optimal choice of DL and $\psi$ is important, especially for data with high variability. The MM approach is less sensitive to the WT settings choices in terms of prediction accuracy. For each model, there are WT settings where both approaches provide predictions with similar error ratios, especially when $DL = 1$.

## 7.7. Key findings

The major findings of this study can be briefly summarised as follows:

1. The coefficients-based approach is found more time-efficient than the components-based approach for utilising SWT for PV power time series with almost all models in the study including LR, SVR, RF and CNN.
2. In comparison to the components-based technique, the coefficients-based approach is less sensitive to the DL in terms of time. This justifies the use of advanced models for PV power prediction with SWT and high DL without sacrificing time complexity.
3. There is a trade-off between prediction error and computing time when employing SWT. While the

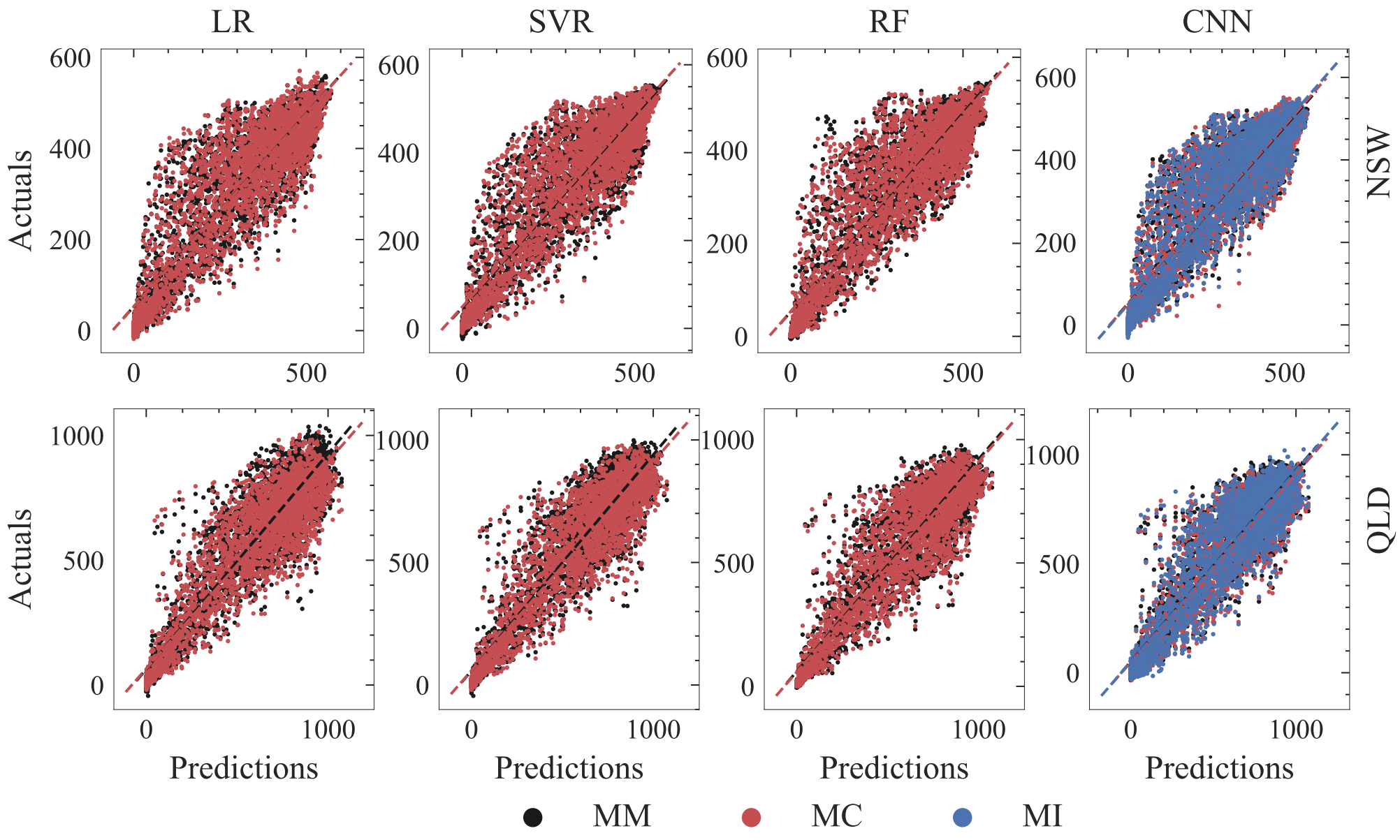

**Figure 17:** Scatter plots of actual vs predicted PV power values

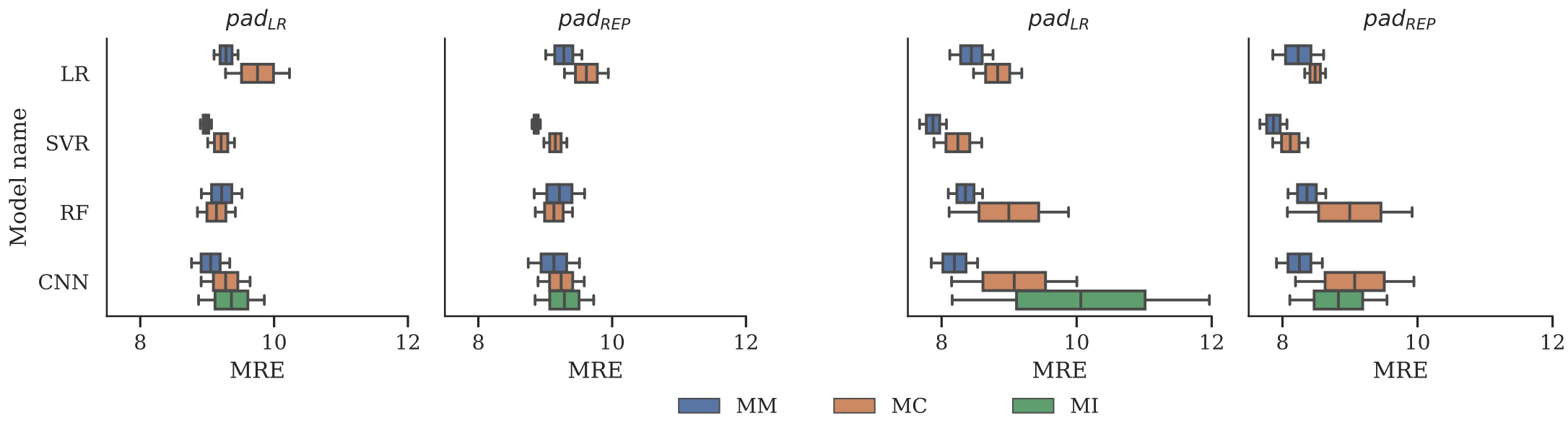

(a) NSW

(b) QLD

**Figure 18:** The MRE distribution of the different approaches to show their sensitivity to different settings

components-based approach produces more accurate forecasts; the coefficients-based approach can produce comparable predictions at certain periods of the day when power generation is less erratic.

4. For PV power time series, the padding method $pad_{REP}$ to extend the signal to avoid the border distortion is found to be more efficient in terms of time and accuracy.

## 8. Conclusions

This study investigates the effectiveness of using WT for time series forecasting and its application for solar PV power data. Two distinct approaches are used to evaluate the WT application efficiency: components-based and coefficients-based. The components-based approach is the most traditional and widely used strategy for applying the WT. It is focused on reconstructing and modelling individual components. The main concern with this approach is the computational complexity for model configuration and training. The coefficients-based approach is the proposed method of using the WT for time series application. It is not dependent on reconstruction of components or independent modelling. Rather it constructs a vector from all the coefficients labelled with the actual data to be learned by the prediction model. The issue of model complexity is evaluated regarding the number of models that must be configured, the elapsed time and predictions accuracy. The evaluation is based on two real-world datasets utilising six error metrics. The results demonstrate the viability of adopting the coefficients-based strategy rather than the components-based approach.

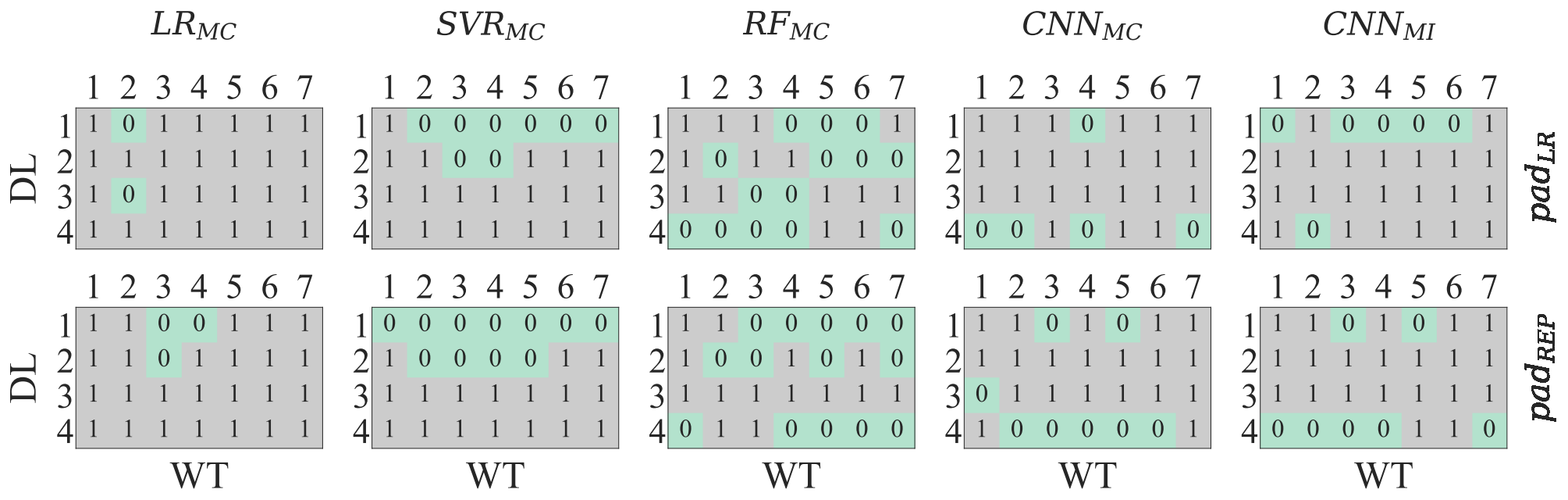

**Figure 19:** Wilcoxon rank-sum test showing the statistical insignificance of differences in accuracy at $P <= 0.05$ for (MM, MC, and MI) with varying DL and $\psi$ for NSW.

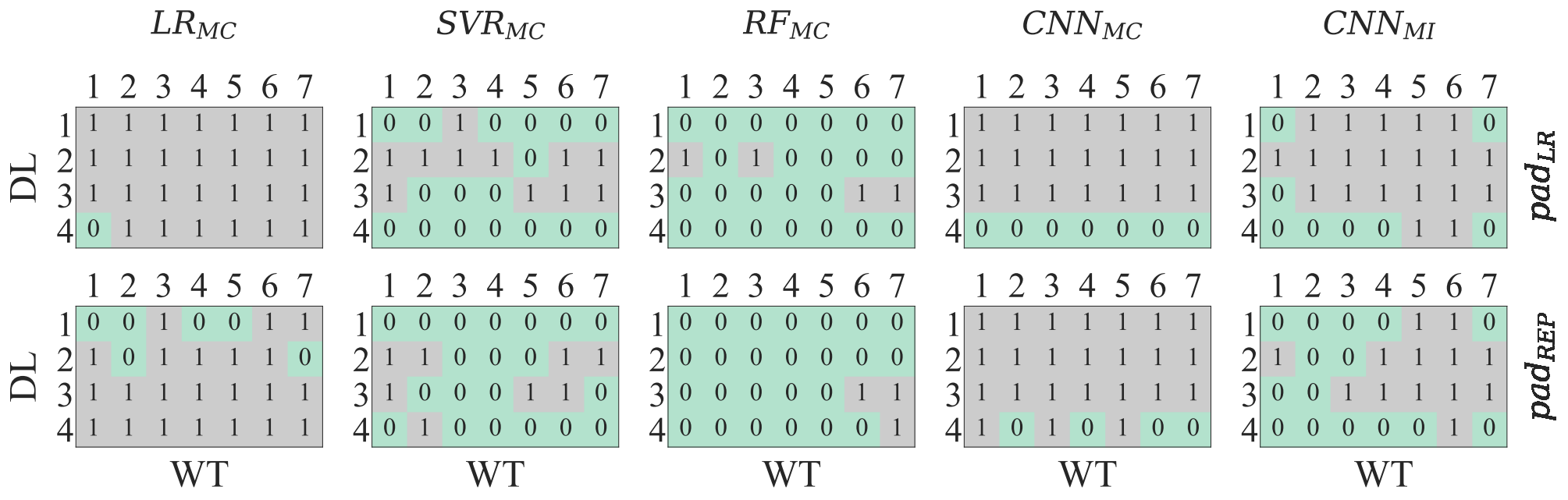

**Figure 20:** Wilcoxon rank-sum test showing the statistical insignificance of differences in accuracy at $P <= 0.05$ for (MM, MC, and MI) with varying DL and $\psi$ for QLD.

The future work will focus on combining both components-based and coefficients-based approaches taking into consideration the time or season. Another possible avenue is to extend the approach for multivariate settings to utilise data from different sources (e.g., weather variables).

## 9. Acknowledgement

Sarah Almaghrabi is supported by a scholarship from Jeddah University, Saudi Arabia.

# A. Actual vs Predicted Values of Models Utilising SWT

Figures A.1 and A.2 show the actual and predicted data for each model for NSW and QLD dataset, respectively. Each row in the figure represents a single prediction model, whereas each column represents a month from the test set.

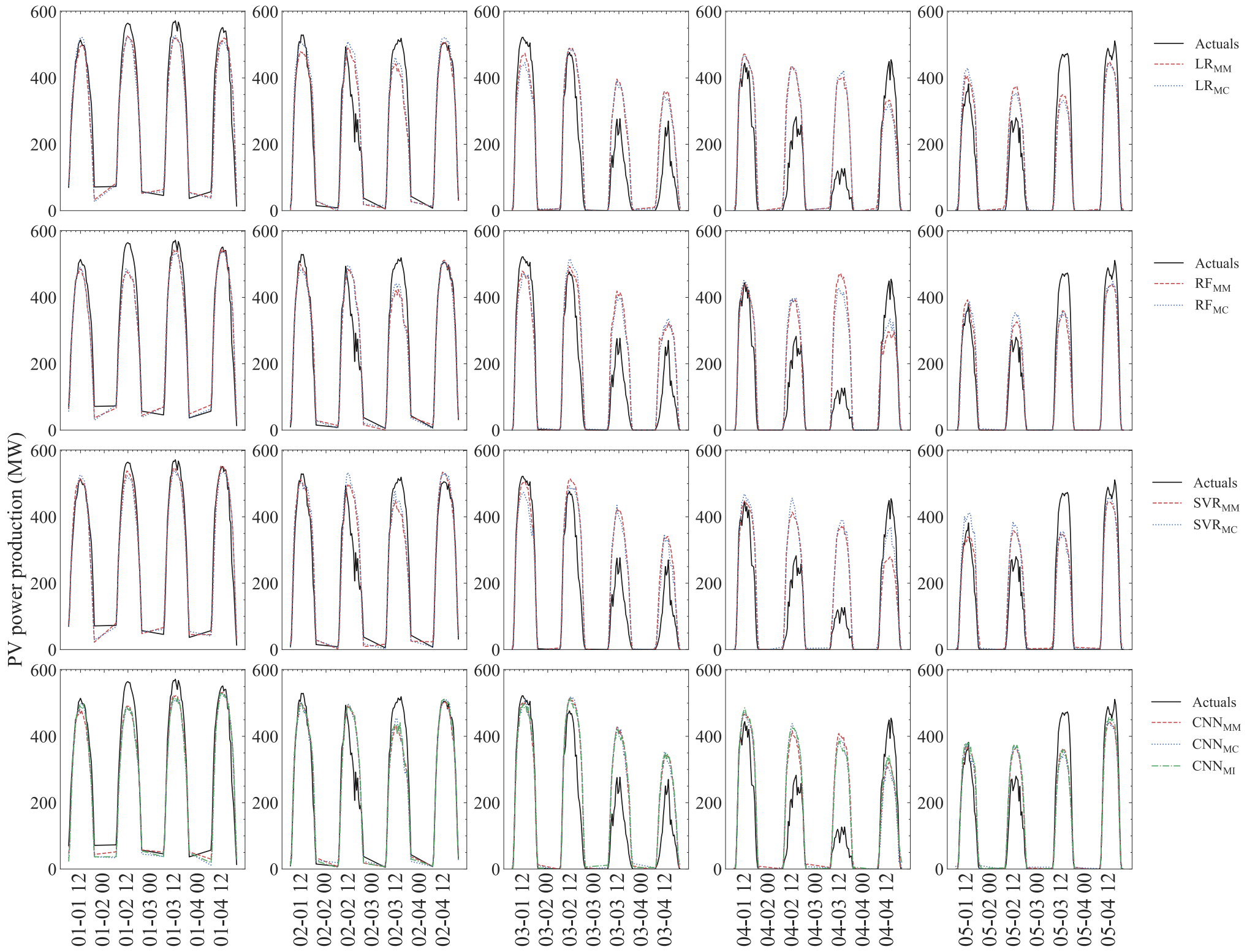

**Figure A.1:** Actual and predicted solar PV power for 4 days from each month in the test set using the different models for NSW

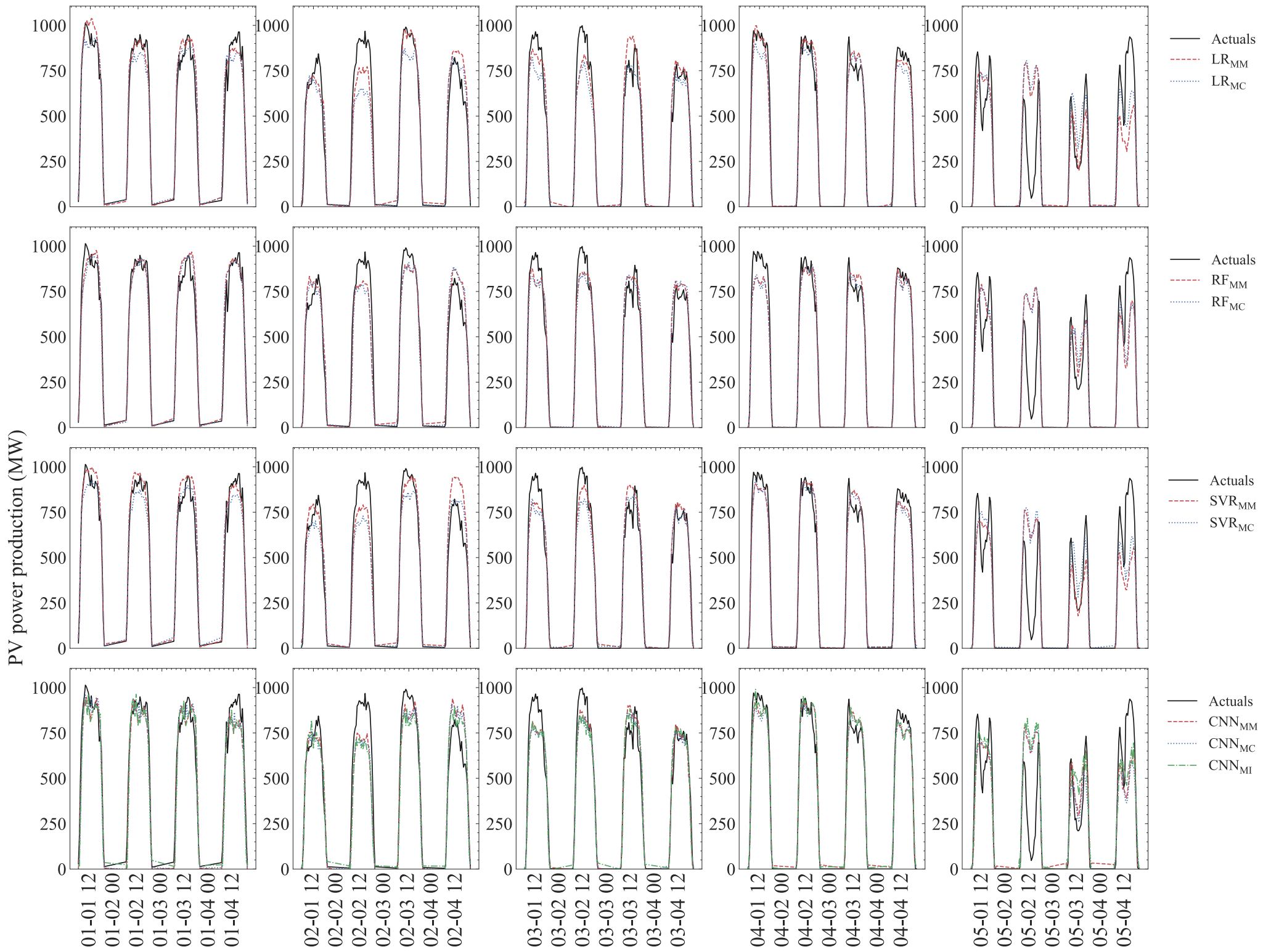

**Figure A.2:** Actual and predicted solar PV power for 4 days from each month in the test set using the different models for QLD

**Table B.1**
Overall performance of all wavelet approaches using paddings generated by $pad_{REP}$ with LR prediction model

| Dataset | Approach | $\psi$ | DL =1 | | | | | | DL =2 | | | | | | DL =3 | | | | | | DL =4 | | | | | |
|---|---|---|---|---|---|---|---|---|---|---|---|---|---|---|---|---|---|---|---|---|---|---|---|---|---|---|
| | | | MAE | RMSE | RAE | MRE | RRSE | $R^2$ | MAE | RMSE | RAE | MRE | RRSE | $R^2$ | MAE | RMSE | RAE | MRE | RRSE | $R^2$ | MAE | RMSE | RAE | MRE | RRSE | $R^2$ |
| NSW | MC | db1 | 56.624 | 79.409 | 0.362 | 9.416 | 0.446 | 0.801 | 56.706 | 79.471 | 0.363 | 9.429 | 0.446 | 0.801 | 56.618 | 79.238 | 0.362 | 9.415 | 0.445 | 0.802 | 56.617 | 79.358 | 0.362 | 9.415 | 0.445 | 0.802 |
| | MM | | 55.800 | 78.279 | 0.357 | 9.279 | 0.439 | 0.807 | 55.624 | 78.094 | 0.356 | 9.249 | 0.438 | 0.808 | 55.534 | 77.744 | 0.355 | 9.235 | 0.436 | 0.81 | 54.414 | 76.083 | 0.348 | 9.048 | 0.427 | 0.818 |
| | MC | db2 | 55.859 | 78.260 | 0.357 | 9.289 | 0.439 | 0.807 | 56.072 | 78.404 | 0.359 | 9.324 | 0.44 | 0.806 | 55.981 | 78.072 | 0.358 | 9.309 | 0.438 | 0.808 | 56.449 | 78.469 | 0.361 | 9.387 | 0.44 | 0.806 |
| | MM | | 56.037 | 78.661 | 0.359 | 9.318 | 0.441 | 0.805 | 55.285 | 77.391 | 0.354 | 9.193 | 0.434 | 0.811 | 55.741 | 77.614 | 0.357 | 9.269 | 0.436 | 0.81 | 55.113 | 76.484 | 0.353 | 9.165 | 0.429 | 0.816 |
| | MC | db3 | 56.060 | 78.443 | 0.359 | 9.322 | 0.440 | 0.806 | 55.931 | 78.102 | 0.358 | 9.301 | 0.438 | 0.808 | 56.68 | 78.782 | 0.363 | 9.425 | 0.442 | 0.805 | 58.781 | 81.948 | 0.376 | 9.775 | 0.46 | 0.789 |
| | MM | | 56.217 | 78.703 | 0.360 | 9.348 | 0.442 | 0.805 | 55.683 | 78.001 | 0.356 | 9.259 | 0.438 | 0.808 | 56.122 | 77.766 | 0.359 | 9.332 | 0.436 | 0.81 | 55.732 | 76.91 | 0.357 | 9.268 | 0.432 | 0.814 |
| | MC | db4 | 56.067 | 78.464 | 0.359 | 9.323 | 0.440 | 0.806 | 56.214 | 78.31 | 0.36 | 9.348 | 0.439 | 0.807 | 57.404 | 80.187 | 0.367 | 9.545 | 0.45 | 0.797 | 58.815 | 81.975 | 0.376 | 9.78 | 0.46 | 0.788 |
| | MM | | 55.828 | 78.013 | 0.357 | 9.284 | 0.438 | 0.808 | 55.083 | 77.304 | 0.352 | 9.16 | 0.434 | 0.812 | 56.232 | 78.007 | 0.36 | 9.351 | 0.438 | 0.808 | 55.577 | 76.209 | 0.356 | 9.242 | 0.428 | 0.817 |
| | MC | db5 | 56.183 | 78.492 | 0.359 | 9.342 | 0.440 | 0.806 | 56.586 | 78.577 | 0.362 | 9.41 | 0.441 | 0.806 | 58.825 | 82.029 | 0.376 | 9.782 | 0.46 | 0.788 | 58.837 | 82.447 | 0.376 | 9.784 | 0.463 | 0.786 |
| | MM | | 55.208 | 77.209 | 0.353 | 9.180 | 0.433 | 0.812 | 54.164 | 76.246 | 0.347 | 9.007 | 0.428 | 0.817 | 57.087 | 79.076 | 0.365 | 9.493 | 0.444 | 0.803 | 55.731 | 76.114 | 0.357 | 9.267 | 0.427 | 0.818 |
| | MC | db6 | 55.852 | 78.205 | 0.357 | 9.287 | 0.439 | 0.807 | 56.489 | 78.862 | 0.361 | 9.393 | 0.443 | 0.804 | 59.141 | 82.628 | 0.378 | 9.834 | 0.464 | 0.785 | 58.795 | 82.32 | 0.376 | 9.777 | 0.462 | 0.787 |
| | MM | | 55.196 | 77.219 | 0.353 | 9.178 | 0.433 | 0.812 | 54.243 | 76.336 | 0.347 | 9.02 | 0.428 | 0.816 | 57.417 | 79.339 | 0.367 | 9.548 | 0.445 | 0.802 | 55.38 | 75.897 | 0.354 | 9.209 | 0.426 | 0.819 |
| | MC | db7 | 55.902 | 78.106 | 0.358 | 9.296 | 0.438 | 0.808 | 57.766 | 80.262 | 0.37 | 9.606 | 0.45 | 0.797 | 59.789 | 83.405 | 0.383 | 9.942 | 0.468 | 0.781 | 59.425 | 83.358 | 0.38 | 9.882 | 0.468 | 0.781 |
| | MM | | 55.202 | 77.267 | 0.353 | 9.179 | 0.434 | 0.812 | 54.305 | 76.268 | 0.347 | 9.03 | 0.428 | 0.817 | 56.436 | 77.98 | 0.361 | 9.385 | 0.438 | 0.808 | 55.579 | 76.746 | 0.356 | 9.242 | 0.431 | 0.815 |
| QLD | MC | db1 | 93.960 | 129.578 | 0.333 | 8.495 | 0.400 | 0.840 | 92.132 | 127.522 | 0.327 | 8.330 | 0.394 | 0.845 | 94.880 | 130.782 | 0.337 | 8.578 | 0.404 | 0.837 | 93.860 | 129.462 | 0.333 | 8.486 | 0.400 | 0.840 |
| | MM | | 93.971 | 129.530 | 0.333 | 8.496 | 0.400 | 0.840 | 95.227 | 131.163 | 0.338 | 8.609 | 0.405 | 0.836 | 92.967 | 127.780 | 0.330 | 8.405 | 0.395 | 0.844 | 89.887 | 124.558 | 0.319 | 8.127 | 0.385 | 0.852 |
| | MC | db2 | 93.888 | 129.518 | 0.333 | 8.488 | 0.400 | 0.840 | 94.609 | 130.396 | 0.336 | 8.554 | 0.403 | 0.838 | 94.935 | 130.810 | 0.337 | 8.583 | 0.404 | 0.837 | 93.103 | 128.608 | 0.330 | 8.417 | 0.397 | 0.842 |
| | MM | | 94.661 | 130.336 | 0.336 | 8.558 | 0.402 | 0.838 | 94.185 | 130.279 | 0.334 | 8.515 | 0.402 | 0.838 | 91.357 | 126.043 | 0.324 | 8.260 | 0.389 | 0.849 | 89.245 | 124.380 | 0.317 | 8.069 | 0.384 | 0.852 |
| | MC | db3 | 93.687 | 129.301 | 0.332 | 8.470 | 0.399 | 0.841 | 94.673 | 130.633 | 0.336 | 8.559 | 0.403 | 0.837 | 93.761 | 129.674 | 0.333 | 8.477 | 0.400 | 0.840 | 95.452 | 132.094 | 0.339 | 8.630 | 0.408 | 0.834 |
| | MM | | 94.358 | 130.116 | 0.335 | 8.531 | 0.402 | 0.839 | 93.073 | 128.604 | 0.330 | 8.415 | 0.397 | 0.842 | 88.707 | 123.564 | 0.315 | 8.020 | 0.382 | 0.854 | 88.198 | 123.551 | 0.313 | 7.974 | 0.381 | 0.854 |
| | MC | db4 | 94.418 | 130.176 | 0.335 | 8.536 | 0.402 | 0.838 | 94.407 | 129.978 | 0.335 | 8.535 | 0.401 | 0.839 | 93.015 | 128.539 | 0.330 | 8.409 | 0.397 | 0.842 | 95.468 | 132.104 | 0.339 | 8.631 | 0.408 | 0.834 |
| | MM | | 95.016 | 131.026 | 0.337 | 8.590 | 0.405 | 0.836 | 92.554 | 127.901 | 0.328 | 8.368 | 0.395 | 0.844 | 88.408 | 123.883 | 0.314 | 7.993 | 0.383 | 0.854 | 88.037 | 122.973 | 0.312 | 7.959 | 0.380 | 0.856 |
| | MC | db5 | 94.598 | 130.411 | 0.336 | 8.553 | 0.403 | 0.838 | 93.278 | 128.797 | 0.331 | 8.433 | 0.398 | 0.842 | 95.545 | 132.191 | 0.339 | 8.638 | 0.408 | 0.833 | 95.507 | 132.144 | 0.339 | 8.635 | 0.408 | 0.834 |
| | MM | | 94.119 | 129.986 | 0.334 | 8.509 | 0.401 | 0.839 | 92.617 | 127.594 | 0.328 | 8.373 | 0.394 | 0.845 | 87.794 | 123.528 | 0.311 | 7.937 | 0.381 | 0.855 | 88.552 | 123.492 | 0.314 | 8.006 | 0.381 | 0.855 |
| | MC | db6 | 94.732 | 130.644 | 0.336 | 8.565 | 0.403 | 0.837 | 93.280 | 128.794 | 0.331 | 8.433 | 0.398 | 0.842 | 95.486 | 132.123 | 0.339 | 8.633 | 0.408 | 0.834 | 95.466 | 132.110 | 0.339 | 8.631 | 0.408 | 0.834 |
| | MM | | 94.007 | 129.803 | 0.333 | 8.499 | 0.401 | 0.839 | 92.527 | 127.279 | 0.328 | 8.365 | 0.393 | 0.846 | 87.268 | 122.676 | 0.310 | 7.890 | 0.379 | 0.857 | 89.824 | 123.755 | 0.319 | 8.121 | 0.382 | 0.854 |
| | MC | db7 | 94.951 | 130.931 | 0.337 | 8.585 | 0.404 | 0.837 | 93.137 | 128.320 | 0.330 | 8.421 | 0.396 | 0.843 | 94.961 | 131.487 | 0.337 | 8.585 | 0.406 | 0.835 | 95.432 | 132.071 | 0.338 | 8.628 | 0.408 | 0.834 |
| | MM | | 93.871 | 129.630 | 0.333 | 8.487 | 0.400 | 0.840 | 92.528 | 127.070 | 0.328 | 8.365 | 0.392 | 0.846 | 86.903 | 121.875 | 0.308 | 7.857 | 0.376 | 0.858 | 90.297 | 125.194 | 0.320 | 8.164 | 0.387 | 0.851 |

## B. Performance of Models Utilising SWT for Each Setting

Tables B.1-B.8 summarise the performance of the models with respect to the various WT parameters. These tables also demonstrate the influence of the padding strategies on the prediction accuracy of the models. In most cases, $pad_{REP}$ outperforms or is comparable to $pad_{LR}$ in terms of prediction performance. For example, for both datasets, when using $pad_{LR}$ as the padding method, with LR as the prediction model and $db1$ as the $\psi$, the prediction accuracy drops when the DL goes from 1 to 4. The prediction accuracy, on the other hand, is more stable when $pad_{REP}$ is employed as the padding method. This holds true for all other prediction models as well.

**Table B.2**
Overall performance of all wavelet approaches using paddings generated by $pad_{LR}$ with LR prediction model

| Dataset | Approach | $\psi$ | DL =1 | | | | | | DL =2 | | | | | | DL =3 | | | | | | DL =4 | | | | | |
|---|---|---|---|---|---|---|---|---|---|---|---|---|---|---|---|---|---|---|---|---|---|---|---|---|---|---|
| | | | MAE | RMSE | RAE | MRE | RRSE | $R^2$ | MAE | RMSE | RAE | MRE | RRSE | $R^2$ | MAE | RMSE | RAE | MRE | RRSE | $R^2$ | MAE | RMSE | RAE | MRE | RRSE | $R^2$ |
| NSW | MC | db1 | 56.542 | 79.651 | 0.362 | 9.402 | 0.447 | 0.800 | 57.563 | 81.103 | 0.368 | 9.572 | 0.455 | 0.793 | 57.209 | 80.111 | 0.366 | 9.513 | 0.450 | 0.798 | 57.187 | 79.919 | 0.366 | 9.509 | 0.448 | 0.799 |
| | MM | | 55.605 | 78.072 | 0.356 | 9.246 | 0.438 | 0.808 | 55.204 | 77.380 | 0.353 | 9.180 | 0.434 | 0.811 | 55.385 | 77.336 | 0.354 | 9.210 | 0.434 | 0.812 | 55.151 | 76.961 | 0.353 | 9.171 | 0.432 | 0.813 |
| | MC | db2 | 55.772 | 78.267 | 0.357 | 9.274 | 0.439 | 0.807 | 56.259 | 78.705 | 0.360 | 9.355 | 0.442 | 0.805 | 56.213 | 78.380 | 0.360 | 9.347 | 0.440 | 0.807 | 58.154 | 80.537 | 0.372 | 9.670 | 0.452 | 0.796 |
| | MM | | 55.757 | 78.345 | 0.357 | 9.272 | 0.440 | 0.807 | 55.268 | 77.340 | 0.354 | 9.190 | 0.434 | 0.812 | 56.536 | 78.424 | 0.362 | 9.401 | 0.440 | 0.806 | 54.748 | 76.365 | 0.350 | 9.104 | 0.429 | 0.816 |
| | MC | db3 | 55.903 | 78.428 | 0.358 | 9.296 | 0.440 | 0.806 | 56.138 | 78.416 | 0.359 | 9.335 | 0.440 | 0.806 | 57.996 | 80.460 | 0.371 | 9.644 | 0.452 | 0.796 | 61.210 | 84.540 | 0.392 | 10.178 | 0.474 | 0.775 |
| | MM | | 55.466 | 77.665 | 0.355 | 9.223 | 0.436 | 0.810 | 55.124 | 77.419 | 0.353 | 9.166 | 0.434 | 0.811 | 56.580 | 78.377 | 0.362 | 9.408 | 0.440 | 0.807 | 55.550 | 76.767 | 0.355 | 9.237 | 0.431 | 0.814 |
| | MC | db4 | 56.072 | 78.616 | 0.359 | 9.324 | 0.441 | 0.805 | 56.210 | 78.572 | 0.360 | 9.347 | 0.441 | 0.806 | 59.595 | 82.392 | 0.381 | 9.910 | 0.462 | 0.786 | 61.201 | 84.550 | 0.392 | 10.177 | 0.474 | 0.775 |
| | MM | | 55.372 | 77.479 | 0.354 | 9.208 | 0.435 | 0.811 | 55.143 | 77.592 | 0.353 | 9.170 | 0.435 | 0.810 | 56.282 | 78.184 | 0.360 | 9.359 | 0.439 | 0.807 | 56.302 | 76.922 | 0.360 | 9.362 | 0.432 | 0.814 |
| | MC | db5 | 56.305 | 78.896 | 0.360 | 9.363 | 0.443 | 0.804 | 56.768 | 78.855 | 0.363 | 9.440 | 0.443 | 0.804 | 61.213 | 84.545 | 0.392 | 10.179 | 0.474 | 0.775 | 61.034 | 84.190 | 0.391 | 10.149 | 0.472 | 0.777 |
| | MM | | 55.068 | 77.118 | 0.352 | 9.157 | 0.433 | 0.813 | 55.047 | 77.439 | 0.352 | 9.154 | 0.435 | 0.811 | 55.856 | 78.189 | 0.357 | 9.288 | 0.439 | 0.807 | 56.573 | 77.264 | 0.362 | 9.407 | 0.434 | 0.812 |
| | MC | db6 | 56.290 | 78.793 | 0.360 | 9.360 | 0.442 | 0.804 | 57.076 | 78.575 | 0.365 | 9.491 | 0.441 | 0.806 | 61.182 | 84.504 | 0.391 | 10.174 | 0.474 | 0.775 | 61.305 | 85.790 | 0.392 | 10.194 | 0.481 | 0.768 |
| | MM | | 54.810 | 76.878 | 0.351 | 9.114 | 0.431 | 0.814 | 54.880 | 77.261 | 0.351 | 9.126 | 0.434 | 0.812 | 55.684 | 77.988 | 0.356 | 9.259 | 0.438 | 0.808 | 56.723 | 77.628 | 0.363 | 9.432 | 0.436 | 0.810 |
| | MC | db7 | 56.170 | 78.530 | 0.359 | 9.340 | 0.441 | 0.806 | 57.699 | 80.235 | 0.369 | 9.595 | 0.450 | 0.797 | 61.416 | 85.279 | 0.393 | 10.213 | 0.479 | 0.771 | 61.523 | 84.712 | 0.394 | 10.230 | 0.475 | 0.774 |
| | MM | | 54.805 | 76.951 | 0.351 | 9.113 | 0.432 | 0.814 | 54.741 | 77.103 | 0.350 | 9.103 | 0.433 | 0.813 | 56.263 | 78.356 | 0.360 | 9.356 | 0.440 | 0.807 | 56.898 | 77.748 | 0.364 | 9.461 | 0.436 | 0.810 |
| QLD | MC | db1 | 93.715 | 129.277 | 0.332 | 8.473 | 0.399 | 0.841 | 95.320 | 131.121 | 0.338 | 8.618 | 0.405 | 0.836 | 97.450 | 133.116 | 0.346 | 8.810 | 0.411 | 0.831 | 97.322 | 132.934 | 0.345 | 8.799 | 0.410 | 0.832 |
| | MM | | 93.589 | 129.016 | 0.332 | 8.461 | 0.398 | 0.841 | 96.907 | 133.415 | 0.344 | 8.761 | 0.412 | 0.830 | 90.887 | 124.705 | 0.322 | 8.217 | 0.385 | 0.852 | 90.325 | 123.226 | 0.320 | 8.166 | 0.380 | 0.855 |
| | MC | db2 | 95.892 | 131.799 | 0.340 | 8.670 | 0.407 | 0.834 | 98.460 | 135.033 | 0.349 | 8.902 | 0.417 | 0.826 | 100.750 | 137.222 | 0.357 | 9.109 | 0.424 | 0.820 | 100.413 | 136.950 | 0.356 | 9.078 | 0.423 | 0.821 |
| | MM | | 94.329 | 129.825 | 0.335 | 8.528 | 0.401 | 0.839 | 96.423 | 132.877 | 0.342 | 8.718 | 0.410 | 0.832 | 90.633 | 124.780 | 0.321 | 8.194 | 0.385 | 0.852 | 89.931 | 123.471 | 0.319 | 8.131 | 0.381 | 0.855 |
| | MC | db3 | 95.124 | 130.944 | 0.337 | 8.600 | 0.404 | 0.837 | 98.380 | 134.573 | 0.349 | 8.894 | 0.416 | 0.827 | 98.930 | 135.404 | 0.351 | 8.944 | 0.418 | 0.825 | 101.432 | 138.926 | 0.360 | 9.170 | 0.429 | 0.816 |
| | MM | | 94.620 | 130.005 | 0.336 | 8.555 | 0.401 | 0.839 | 95.248 | 130.685 | 0.338 | 8.611 | 0.404 | 0.837 | 90.571 | 124.758 | 0.321 | 8.188 | 0.385 | 0.852 | 89.825 | 123.107 | 0.319 | 8.121 | 0.380 | 0.856 |
| | MC | db4 | 96.659 | 132.795 | 0.343 | 8.739 | 0.410 | 0.832 | 99.164 | 135.232 | 0.352 | 8.965 | 0.418 | 0.826 | 100.101 | 136.413 | 0.355 | 9.050 | 0.421 | 0.823 | 101.521 | 139.026 | 0.360 | 9.178 | 0.429 | 0.816 |
| | MM | | 94.750 | 130.325 | 0.336 | 8.566 | 0.402 | 0.838 | 94.972 | 129.947 | 0.337 | 8.586 | 0.401 | 0.839 | 90.126 | 124.304 | 0.320 | 8.148 | 0.384 | 0.853 | 90.114 | 123.072 | 0.320 | 8.147 | 0.380 | 0.856 |
| | MC | db5 | 98.413 | 134.969 | 0.349 | 8.897 | 0.417 | 0.826 | 98.545 | 134.598 | 0.350 | 8.909 | 0.416 | 0.827 | 101.565 | 139.079 | 0.360 | 9.182 | 0.429 | 0.816 | 101.594 | 139.115 | 0.360 | 9.185 | 0.430 | 0.815 |
| | MM | | 94.228 | 130.079 | 0.334 | 8.519 | 0.402 | 0.839 | 95.046 | 130.297 | 0.337 | 8.593 | 0.402 | 0.838 | 90.205 | 124.399 | 0.320 | 8.155 | 0.384 | 0.852 | 90.248 | 123.141 | 0.320 | 8.159 | 0.380 | 0.855 |
| | MC | db6 | 98.397 | 134.918 | 0.349 | 8.896 | 0.417 | 0.826 | 98.282 | 134.435 | 0.349 | 8.886 | 0.415 | 0.828 | 101.615 | 139.142 | 0.360 | 9.187 | 0.430 | 0.815 | 100.340 | 137.670 | 0.356 | 9.072 | 0.425 | 0.819 |
| | MM | | 94.085 | 129.914 | 0.334 | 8.506 | 0.401 | 0.839 | 94.889 | 130.381 | 0.337 | 8.579 | 0.403 | 0.838 | 90.287 | 124.106 | 0.320 | 8.163 | 0.383 | 0.853 | 90.173 | 122.952 | 0.320 | 8.152 | 0.380 | 0.856 |
| | MC | db7 | 99.100 | 135.326 | 0.351 | 8.960 | 0.418 | 0.825 | 98.510 | 134.429 | 0.349 | 8.906 | 0.415 | 0.828 | 101.512 | 139.014 | 0.360 | 9.178 | 0.429 | 0.816 | 101.344 | 138.844 | 0.359 | 9.163 | 0.429 | 0.816 |
| | MM | | 94.476 | 130.068 | 0.335 | 8.542 | 0.402 | 0.839 | 94.927 | 130.481 | 0.337 | 8.582 | 0.403 | 0.838 | 90.528 | 123.990 | 0.321 | 8.185 | 0.383 | 0.853 | 90.205 | 122.836 | 0.320 | 8.155 | 0.379 | 0.856 |

**Table B.3**
Overall performance of all wavelet approaches using paddings generated by $pad_{REP}$ with SVR prediction model

| Dataset | Approach | $\psi$ | DL =1 | | | | | | DL =2 | | | | | | DL =3 | | | | | | DL =4 | | | | | |
|---|---|---|---|---|---|---|---|---|---|---|---|---|---|---|---|---|---|---|---|---|---|---|---|---|---|---|
| | | | MAE | RMSE | RAE | MRE | RRSE | $R^2$ | MAE | RMSE | RAE | MRE | RRSE | $R^2$ | MAE | RMSE | RAE | MRE | RRSE | $R^2$ | MAE | RMSE | RAE | MRE | RRSE | $R^2$ |
| NSW | MC | db1 | 54.440 | 78.412 | 0.348 | 9.053 | 0.440 | 0.806 | 54.418 | 78.433 | 0.348 | 9.049 | 0.440 | 0.806 | 54.364 | 78.394 | 0.348 | 9.040 | 0.440 | 0.806 | 54.281 | 78.311 | 0.347 | 9.026 | 0.439 | 0.807 |
| | MM | | 53.566 | 78.093 | 0.343 | 8.907 | 0.438 | 0.808 | 53.681 | 77.740 | 0.343 | 8.926 | 0.436 | 0.810 | 53.248 | 75.994 | 0.341 | 8.855 | 0.426 | 0.818 | 53.391 | 76.036 | 0.342 | 8.878 | 0.427 | 0.818 |
| | MC | db2 | 54.637 | 78.503 | 0.350 | 9.085 | 0.441 | 0.806 | 54.392 | 78.153 | 0.348 | 9.045 | 0.439 | 0.808 | 54.132 | 77.835 | 0.346 | 9.001 | 0.437 | 0.809 | 54.171 | 77.927 | 0.347 | 9.008 | 0.437 | 0.809 |
| | MM | | 53.522 | 78.146 | 0.342 | 8.900 | 0.439 | 0.808 | 53.512 | 77.680 | 0.342 | 8.898 | 0.436 | 0.810 | 52.995 | 75.604 | 0.339 | 8.812 | 0.424 | 0.820 | 53.036 | 75.380 | 0.339 | 8.819 | 0.423 | 0.821 |
| | MC | db3 | 54.526 | 78.283 | 0.349 | 9.067 | 0.439 | 0.807 | 54.145 | 77.675 | 0.346 | 9.004 | 0.436 | 0.810 | 54.530 | 78.188 | 0.349 | 9.068 | 0.439 | 0.807 | 54.407 | 78.178 | 0.348 | 9.047 | 0.439 | 0.808 |
| | MM | | 53.528 | 78.192 | 0.342 | 8.901 | 0.439 | 0.807 | 53.464 | 77.657 | 0.342 | 8.890 | 0.436 | 0.810 | 53.052 | 75.640 | 0.339 | 8.822 | 0.424 | 0.820 | 52.936 | 75.206 | 0.339 | 8.803 | 0.422 | 0.822 |
| | MC | db4 | 54.223 | 77.705 | 0.347 | 9.017 | 0.436 | 0.810 | 54.227 | 77.777 | 0.347 | 9.017 | 0.436 | 0.809 | 54.577 | 78.305 | 0.349 | 9.075 | 0.439 | 0.807 | 54.603 | 77.955 | 0.349 | 9.080 | 0.437 | 0.809 |
| | MM | | 53.523 | 78.233 | 0.342 | 8.900 | 0.439 | 0.807 | 53.412 | 77.622 | 0.342 | 8.882 | 0.436 | 0.810 | 53.159 | 75.690 | 0.340 | 8.840 | 0.425 | 0.820 | 52.980 | 75.219 | 0.339 | 8.810 | 0.422 | 0.822 |
| | MC | db5 | 54.064 | 77.402 | 0.346 | 8.990 | 0.434 | 0.811 | 54.468 | 78.097 | 0.348 | 9.057 | 0.438 | 0.808 | 55.388 | 79.341 | 0.354 | 9.210 | 0.445 | 0.802 | 54.562 | 77.975 | 0.349 | 9.073 | 0.438 | 0.809 |
| | MM | | 53.521 | 78.261 | 0.342 | 8.900 | 0.439 | 0.807 | 53.372 | 77.544 | 0.341 | 8.875 | 0.435 | 0.811 | 53.194 | 75.660 | 0.340 | 8.845 | 0.425 | 0.820 | 52.957 | 75.165 | 0.339 | 8.806 | 0.422 | 0.822 |
| | MC | db6 | 54.000 | 77.331 | 0.345 | 8.980 | 0.434 | 0.812 | 54.878 | 78.696 | 0.351 | 9.125 | 0.442 | 0.805 | 56.065 | 80.109 | 0.359 | 9.323 | 0.450 | 0.798 | 54.382 | 77.717 | 0.348 | 9.043 | 0.436 | 0.810 |
| | MM | | 53.519 | 78.272 | 0.342 | 8.900 | 0.439 | 0.807 | 53.366 | 77.507 | 0.341 | 8.874 | 0.435 | 0.811 | 53.316 | 75.758 | 0.341 | 8.866 | 0.425 | 0.819 | 53.088 | 75.257 | 0.340 | 8.828 | 0.422 | 0.822 |
| | MC | db7 | 54.317 | 77.617 | 0.348 | 9.032 | 0.436 | 0.810 | 55.527 | 79.653 | 0.355 | 9.233 | 0.447 | 0.800 | 55.961 | 80.071 | 0.358 | 9.306 | 0.449 | 0.798 | 54.176 | 77.872 | 0.347 | 9.009 | 0.437 | 0.809 |
| | MM | | 53.518 | 78.282 | 0.342 | 8.899 | 0.439 | 0.807 | 53.367 | 77.482 | 0.341 | 8.874 | 0.435 | 0.811 | 53.404 | 75.929 | 0.342 | 8.880 | 0.426 | 0.818 | 53.084 | 75.295 | 0.340 | 8.827 | 0.423 | 0.821 |
| QLD | MC | db1 | 88.921 | 123.318 | 0.315 | 8.039 | 0.381 | 0.855 | 89.175 | 123.544 | 0.316 | 8.062 | 0.381 | 0.854 | 88.978 | 122.800 | 0.316 | 8.044 | 0.379 | 0.856 | 88.602 | 122.295 | 0.314 | 8.010 | 0.378 | 0.857 |
| | MM | | 88.487 | 122.283 | 0.314 | 8.000 | 0.378 | 0.857 | 87.469 | 120.705 | 0.310 | 7.908 | 0.373 | 0.861 | 84.969 | 118.660 | 0.301 | 7.682 | 0.366 | 0.866 | 84.957 | 118.219 | 0.301 | 7.681 | 0.365 | 0.867 |
| | MC | db2 | 89.044 | 123.371 | 0.316 | 8.050 | 0.381 | 0.855 | 88.609 | 122.761 | 0.314 | 8.011 | 0.379 | 0.856 | 88.348 | 122.525 | 0.313 | 7.987 | 0.378 | 0.857 | 87.903 | 121.456 | 0.312 | 7.947 | 0.375 | 0.859 |
| | MM | | 89.173 | 123.258 | 0.316 | 8.062 | 0.381 | 0.855 | 87.007 | 120.416 | 0.309 | 7.866 | 0.372 | 0.862 | 84.781 | 118.683 | 0.301 | 7.665 | 0.366 | 0.866 | 84.907 | 118.210 | 0.301 | 7.676 | 0.365 | 0.867 |
| | MC | db3 | 89.838 | 124.331 | 0.319 | 8.122 | 0.384 | 0.853 | 88.644 | 123.061 | 0.314 | 8.014 | 0.380 | 0.856 | 88.375 | 122.237 | 0.313 | 7.990 | 0.377 | 0.858 | 87.288 | 120.919 | 0.310 | 7.892 | 0.373 | 0.861 |
| | MM | | 88.651 | 122.716 | 0.314 | 8.015 | 0.379 | 0.856 | 87.301 | 120.870 | 0.310 | 7.893 | 0.373 | 0.861 | 84.895 | 118.833 | 0.301 | 7.675 | 0.367 | 0.865 | 85.049 | 118.249 | 0.302 | 7.689 | 0.365 | 0.867 |
| | MC | db4 | 88.839 | 123.249 | 0.315 | 8.032 | 0.381 | 0.855 | 88.322 | 122.663 | 0.313 | 7.985 | 0.379 | 0.857 | 88.404 | 122.384 | 0.314 | 7.993 | 0.378 | 0.857 | 86.963 | 120.802 | 0.308 | 7.862 | 0.373 | 0.861 |
| | MM | | 88.879 | 123.061 | 0.315 | 8.036 | 0.380 | 0.856 | 87.017 | 120.608 | 0.309 | 7.867 | 0.372 | 0.861 | 84.906 | 118.802 | 0.301 | 7.676 | 0.367 | 0.865 | 85.075 | 118.156 | 0.302 | 7.692 | 0.365 | 0.867 |
| | MC | db5 | 88.543 | 123.068 | 0.314 | 8.005 | 0.380 | 0.856 | 88.393 | 122.083 | 0.313 | 7.992 | 0.377 | 0.858 | 91.020 | 126.576 | 0.323 | 8.229 | 0.391 | 0.847 | 87.280 | 120.887 | 0.310 | 7.891 | 0.373 | 0.861 |
| | MM | | 89.099 | 123.313 | 0.316 | 8.055 | 0.381 | 0.855 | 86.786 | 120.442 | 0.308 | 7.846 | 0.372 | 0.862 | 85.033 | 118.832 | 0.302 | 7.688 | 0.367 | 0.865 | 84.912 | 118.011 | 0.301 | 7.677 | 0.364 | 0.867 |
| | MC | db6 | 87.640 | 121.938 | 0.311 | 7.923 | 0.377 | 0.858 | 88.921 | 122.971 | 0.315 | 8.039 | 0.380 | 0.856 | 92.653 | 129.578 | 0.329 | 8.377 | 0.400 | 0.840 | 86.886 | 120.741 | 0.308 | 7.855 | 0.373 | 0.861 |
| | MM | | 89.219 | 123.470 | 0.316 | 8.066 | 0.381 | 0.855 | 86.774 | 120.371 | 0.308 | 7.845 | 0.372 | 0.862 | 85.587 | 119.117 | 0.304 | 7.738 | 0.368 | 0.865 | 85.394 | 118.238 | 0.303 | 7.720 | 0.365 | 0.867 |
| | MC | db7 | 87.446 | 121.781 | 0.310 | 7.906 | 0.376 | 0.859 | 90.317 | 125.095 | 0.320 | 8.166 | 0.386 | 0.851 | 90.791 | 126.824 | 0.322 | 8.208 | 0.392 | 0.847 | 87.220 | 121.084 | 0.309 | 7.886 | 0.374 | 0.860 |
| | MM | | 89.186 | 123.466 | 0.316 | 8.063 | 0.381 | 0.855 | 86.614 | 120.172 | 0.307 | 7.831 | 0.371 | 0.862 | 85.883 | 119.170 | 0.305 | 7.765 | 0.368 | 0.865 | 85.582 | 118.205 | 0.304 | 7.737 | 0.365 | 0.867 |

**Table B.4**
Overall performance of all wavelet approaches using paddings generated by $pad_{LR}$ with SVR prediction model

| Dataset | Approach | $\psi$ | DL =1 | | | | | | DL =2 | | | | | | DL =3 | | | | | | DL =4 | | | | | |
|---|---|---|---|---|---|---|---|---|---|---|---|---|---|---|---|---|---|---|---|---|---|---|---|---|---|---|
| | | | MAE | RMSE | RAE | MRE | RRSE | $R^2$ | MAE | RMSE | RAE | MRE | RRSE | $R^2$ | MAE | RMSE | RAE | MRE | RRSE | $R^2$ | MAE | RMSE | RAE | MRE | RRSE | $R^2$ |
| NSW | MC | db1 | 54.559 | 78.555 | 0.349 | 9.072 | 0.441 | 0.806 | 54.492 | 78.412 | 0.349 | 9.061 | 0.440 | 0.806 | 54.497 | 78.572 | 0.349 | 9.062 | 0.441 | 0.806 | 54.537 | 78.731 | 0.349 | 9.069 | 0.442 | 0.805 |
| | MM | | 53.564 | 78.086 | 0.343 | 8.907 | 0.438 | 0.808 | 53.660 | 77.695 | 0.343 | 8.923 | 0.436 | 0.810 | 53.691 | 76.961 | 0.344 | 8.928 | 0.432 | 0.813 | 53.976 | 76.863 | 0.345 | 8.976 | 0.431 | 0.814 |
| | MC | db2 | 54.707 | 78.633 | 0.350 | 9.097 | 0.441 | 0.805 | 54.536 | 78.408 | 0.349 | 9.069 | 0.440 | 0.806 | 54.364 | 78.209 | 0.348 | 9.040 | 0.439 | 0.807 | 54.384 | 78.357 | 0.348 | 9.043 | 0.440 | 0.807 |
| | MM | | 53.562 | 78.189 | 0.343 | 8.907 | 0.439 | 0.807 | 53.629 | 77.691 | 0.343 | 8.918 | 0.436 | 0.810 | 53.644 | 76.675 | 0.343 | 8.920 | 0.430 | 0.815 | 54.175 | 76.666 | 0.347 | 9.009 | 0.430 | 0.815 |
| | MC | db3 | 54.665 | 78.476 | 0.350 | 9.090 | 0.440 | 0.806 | 54.445 | 77.996 | 0.348 | 9.054 | 0.438 | 0.808 | 54.813 | 78.658 | 0.351 | 9.115 | 0.441 | 0.805 | 54.750 | 78.623 | 0.350 | 9.104 | 0.441 | 0.805 |
| | MM | | 53.572 | 78.239 | 0.343 | 8.908 | 0.439 | 0.807 | 53.606 | 77.690 | 0.343 | 8.914 | 0.436 | 0.810 | 53.793 | 76.753 | 0.344 | 8.945 | 0.431 | 0.814 | 54.320 | 76.686 | 0.348 | 9.033 | 0.430 | 0.815 |
| | MC | db4 | 54.315 | 77.748 | 0.348 | 9.032 | 0.436 | 0.810 | 54.575 | 78.229 | 0.349 | 9.075 | 0.439 | 0.807 | 54.883 | 78.795 | 0.351 | 9.126 | 0.442 | 0.804 | 55.307 | 79.280 | 0.354 | 9.197 | 0.445 | 0.802 |
| | MM | | 53.530 | 78.263 | 0.342 | 8.901 | 0.439 | 0.807 | 53.558 | 77.678 | 0.343 | 8.906 | 0.436 | 0.810 | 53.850 | 76.757 | 0.345 | 8.955 | 0.431 | 0.814 | 54.371 | 76.707 | 0.348 | 9.041 | 0.430 | 0.815 |
| | MC | db5 | 54.268 | 77.629 | 0.347 | 9.024 | 0.436 | 0.810 | 54.686 | 78.560 | 0.350 | 9.094 | 0.441 | 0.806 | 55.870 | 80.065 | 0.357 | 9.290 | 0.449 | 0.798 | 55.593 | 79.681 | 0.356 | 9.244 | 0.447 | 0.800 |
| | MM | | 53.527 | 78.277 | 0.342 | 8.901 | 0.439 | 0.807 | 53.532 | 77.635 | 0.343 | 8.902 | 0.436 | 0.810 | 54.117 | 77.118 | 0.346 | 8.999 | 0.433 | 0.813 | 54.382 | 76.791 | 0.348 | 9.043 | 0.431 | 0.814 |
| | MC | db6 | 54.173 | 77.403 | 0.347 | 9.008 | 0.434 | 0.811 | 55.380 | 79.636 | 0.354 | 9.209 | 0.447 | 0.800 | 56.577 | 81.175 | 0.362 | 9.408 | 0.456 | 0.792 | 55.536 | 79.672 | 0.355 | 9.235 | 0.447 | 0.800 |
| | MM | | 53.532 | 78.284 | 0.343 | 8.902 | 0.439 | 0.807 | 53.516 | 77.604 | 0.342 | 8.899 | 0.436 | 0.810 | 54.231 | 77.215 | 0.347 | 9.018 | 0.433 | 0.812 | 54.455 | 76.870 | 0.348 | 9.055 | 0.431 | 0.814 |
| | MC | db7 | 54.379 | 77.829 | 0.348 | 9.042 | 0.437 | 0.809 | 55.811 | 80.155 | 0.357 | 9.281 | 0.450 | 0.798 | 56.474 | 81.012 | 0.361 | 9.391 | 0.455 | 0.793 | 55.409 | 79.604 | 0.355 | 9.214 | 0.447 | 0.800 |
| | MM | | 53.537 | 78.296 | 0.343 | 8.902 | 0.439 | 0.807 | 53.511 | 77.595 | 0.342 | 8.898 | 0.435 | 0.810 | 54.302 | 77.274 | 0.347 | 9.030 | 0.434 | 0.812 | 54.511 | 76.924 | 0.349 | 9.065 | 0.432 | 0.814 |
| QLD | MC | db1 | 88.585 | 122.784 | 0.314 | 8.009 | 0.379 | 0.856 | 90.352 | 124.855 | 0.320 | 8.169 | 0.386 | 0.851 | 88.786 | 122.375 | 0.315 | 8.027 | 0.378 | 0.857 | 88.288 | 121.553 | 0.313 | 7.982 | 0.375 | 0.859 |
| | MM | | 88.359 | 122.170 | 0.313 | 7.989 | 0.377 | 0.858 | 87.475 | 120.646 | 0.310 | 7.909 | 0.373 | 0.861 | 85.079 | 118.544 | 0.302 | 7.692 | 0.366 | 0.866 | 85.101 | 118.002 | 0.302 | 7.694 | 0.364 | 0.867 |
| | MC | db2 | 90.157 | 124.731 | 0.320 | 8.151 | 0.385 | 0.852 | 89.528 | 123.672 | 0.318 | 8.094 | 0.382 | 0.854 | 88.914 | 122.532 | 0.315 | 8.039 | 0.378 | 0.857 | 88.742 | 122.462 | 0.315 | 8.023 | 0.378 | 0.857 |
| | MM | | 88.738 | 122.719 | 0.315 | 8.023 | 0.379 | 0.856 | 86.989 | 120.435 | 0.309 | 7.865 | 0.372 | 0.862 | 85.191 | 118.516 | 0.302 | 7.702 | 0.366 | 0.866 | 85.002 | 117.791 | 0.301 | 7.685 | 0.364 | 0.868 |
| | MC | db3 | 90.933 | 125.803 | 0.323 | 8.221 | 0.388 | 0.849 | 89.221 | 123.380 | 0.316 | 8.066 | 0.381 | 0.855 | 89.139 | 122.926 | 0.316 | 8.059 | 0.380 | 0.856 | 88.891 | 122.990 | 0.315 | 8.037 | 0.380 | 0.856 |
| | MM | | 88.859 | 122.880 | 0.315 | 8.034 | 0.379 | 0.856 | 87.007 | 120.565 | 0.309 | 7.866 | 0.372 | 0.861 | 85.042 | 118.449 | 0.302 | 7.689 | 0.366 | 0.866 | 84.875 | 117.649 | 0.301 | 7.674 | 0.363 | 0.868 |
| | MC | db4 | 88.571 | 123.008 | 0.314 | 8.008 | 0.380 | 0.856 | 90.352 | 124.695 | 0.320 | 8.169 | 0.385 | 0.852 | 89.096 | 123.113 | 0.316 | 8.055 | 0.380 | 0.855 | 90.855 | 125.765 | 0.322 | 8.214 | 0.388 | 0.849 |
| | MM | | 88.826 | 122.883 | 0.315 | 8.031 | 0.379 | 0.856 | 86.831 | 120.452 | 0.308 | 7.850 | 0.372 | 0.862 | 85.061 | 118.429 | 0.302 | 7.690 | 0.366 | 0.866 | 84.970 | 117.557 | 0.301 | 7.682 | 0.363 | 0.868 |
| | MC | db5 | 88.240 | 122.443 | 0.313 | 7.978 | 0.378 | 0.857 | 88.723 | 122.193 | 0.315 | 8.021 | 0.377 | 0.858 | 91.858 | 126.640 | 0.326 | 8.305 | 0.391 | 0.847 | 89.895 | 124.415 | 0.319 | 8.127 | 0.384 | 0.852 |
| | MM | | 89.076 | 123.193 | 0.316 | 8.053 | 0.380 | 0.855 | 86.791 | 120.309 | 0.308 | 7.847 | 0.371 | 0.862 | 85.248 | 118.557 | 0.302 | 7.707 | 0.366 | 0.866 | 85.062 | 117.516 | 0.302 | 7.690 | 0.363 | 0.868 |
| | MC | db6 | 87.461 | 121.516 | 0.310 | 7.907 | 0.375 | 0.859 | 89.338 | 122.674 | 0.317 | 8.077 | 0.379 | 0.857 | 94.472 | 131.165 | 0.335 | 8.541 | 0.405 | 0.836 | 88.596 | 122.906 | 0.314 | 8.010 | 0.380 | 0.856 |
| | MM | | 89.265 | 123.425 | 0.317 | 8.070 | 0.381 | 0.855 | 86.617 | 120.069 | 0.307 | 7.831 | 0.371 | 0.863 | 85.385 | 118.695 | 0.303 | 7.720 | 0.367 | 0.866 | 85.283 | 117.661 | 0.302 | 7.710 | 0.363 | 0.868 |
| | MC | db7 | 87.240 | 121.235 | 0.309 | 7.887 | 0.374 | 0.860 | 91.430 | 124.852 | 0.324 | 8.266 | 0.386 | 0.851 | 95.057 | 132.351 | 0.337 | 8.594 | 0.409 | 0.833 | 89.241 | 123.322 | 0.317 | 8.068 | 0.381 | 0.855 |
| | MM | | 89.277 | 123.436 | 0.317 | 8.072 | 0.381 | 0.855 | 86.465 | 119.853 | 0.307 | 7.817 | 0.370 | 0.863 | 85.564 | 118.830 | 0.303 | 7.736 | 0.367 | 0.865 | 85.320 | 117.622 | 0.303 | 7.714 | 0.363 | 0.868 |

**Table B.5**
Overall performance of all wavelet approaches using paddings generated by $pad_{REP}$ with RF prediction model

| Dataset | Approach | $\psi$ | DL =1 | | | | | | DL =2 | | | | | | DL =3 | | | | | | DL =4 | | | | | |
|---|---|---|---|---|---|---|---|---|---|---|---|---|---|---|---|---|---|---|---|---|---|---|---|---|---|---|
| | | | MAE | RMSE | RAE | MRE | RRSE | $R^2$ | MAE | RMSE | RAE | MRE | RRSE | $R^2$ | MAE | RMSE | RAE | MRE | RRSE | $R^2$ | MAE | RMSE | RAE | MRE | RRSE | $R^2$ |
| NSW | MC | db1 | 53.227 | 76.175 | 0.341 | 8.851 | 0.427 | 0.817 | 53.838 | 77.088 | 0.344 | 8.953 | 0.433 | 0.813 | 53.989 | 77.151 | 0.345 | 8.978 | 0.433 | 0.813 | 53.830 | 76.548 | 0.344 | 8.951 | 0.430 | 0.815 |
| | MM | | 53.816 | 77.511 | 0.344 | 8.949 | 0.435 | 0.811 | 55.094 | 79.377 | 0.352 | 9.161 | 0.445 | 0.802 | 55.564 | 78.738 | 0.356 | 9.239 | 0.442 | 0.805 | 55.020 | 74.839 | 0.352 | 9.149 | 0.420 | 0.824 |
| | MC | db2 | 53.965 | 77.499 | 0.345 | 8.974 | 0.435 | 0.811 | 54.558 | 78.657 | 0.349 | 9.072 | 0.441 | 0.805 | 54.679 | 78.053 | 0.350 | 9.092 | 0.438 | 0.808 | 54.677 | 78.261 | 0.350 | 9.092 | 0.439 | 0.807 |
| | MM | | 53.132 | 76.502 | 0.340 | 8.835 | 0.429 | 0.816 | 54.849 | 78.432 | 0.351 | 9.121 | 0.440 | 0.806 | 56.043 | 79.353 | 0.359 | 9.319 | 0.445 | 0.802 | 55.373 | 75.599 | 0.354 | 9.208 | 0.424 | 0.820 |
| | MC | db3 | 53.770 | 77.063 | 0.344 | 8.941 | 0.432 | 0.813 | 54.138 | 77.311 | 0.346 | 9.002 | 0.434 | 0.812 | 55.387 | 78.295 | 0.354 | 9.210 | 0.439 | 0.807 | 55.440 | 78.227 | 0.355 | 9.219 | 0.439 | 0.807 |
| | MM | | 53.386 | 76.856 | 0.342 | 8.877 | 0.431 | 0.814 | 54.470 | 77.814 | 0.349 | 9.058 | 0.437 | 0.809 | 56.976 | 80.227 | 0.365 | 9.474 | 0.450 | 0.797 | 55.343 | 75.735 | 0.354 | 9.203 | 0.425 | 0.819 |
| | MC | db4 | 53.810 | 76.860 | 0.344 | 8.948 | 0.431 | 0.814 | 54.146 | 76.510 | 0.346 | 9.004 | 0.429 | 0.816 | 55.589 | 78.089 | 0.356 | 9.244 | 0.438 | 0.808 | 54.460 | 76.776 | 0.348 | 9.056 | 0.431 | 0.814 |
| | MM | | 53.271 | 76.799 | 0.341 | 8.858 | 0.431 | 0.814 | 54.657 | 78.274 | 0.350 | 9.089 | 0.439 | 0.807 | 56.869 | 79.744 | 0.364 | 9.457 | 0.448 | 0.800 | 55.089 | 75.709 | 0.352 | 9.161 | 0.425 | 0.819 |
| | MC | db5 | 53.697 | 76.700 | 0.344 | 8.929 | 0.430 | 0.815 | 55.089 | 77.859 | 0.352 | 9.161 | 0.437 | 0.809 | 56.381 | 78.599 | 0.361 | 9.375 | 0.441 | 0.805 | 54.898 | 76.845 | 0.351 | 9.129 | 0.431 | 0.814 |
| | MM | | 53.403 | 76.871 | 0.342 | 8.880 | 0.431 | 0.814 | 54.842 | 78.589 | 0.351 | 9.119 | 0.441 | 0.805 | 57.155 | 80.157 | 0.366 | 9.504 | 0.450 | 0.798 | 55.050 | 76.085 | 0.352 | 9.154 | 0.427 | 0.818 |
| | MC | db6 | 54.444 | 77.069 | 0.348 | 9.053 | 0.433 | 0.813 | 56.571 | 79.271 | 0.362 | 9.407 | 0.445 | 0.802 | 56.186 | 78.973 | 0.359 | 9.343 | 0.443 | 0.804 | 55.063 | 77.120 | 0.352 | 9.156 | 0.433 | 0.813 |
| | MM | | 53.486 | 76.910 | 0.342 | 8.894 | 0.432 | 0.814 | 54.903 | 78.728 | 0.351 | 9.130 | 0.442 | 0.805 | 57.435 | 80.310 | 0.367 | 9.551 | 0.451 | 0.797 | 55.621 | 76.541 | 0.356 | 9.249 | 0.430 | 0.815 |
| | MC | db7 | 54.752 | 77.252 | 0.350 | 9.104 | 0.434 | 0.812 | 56.062 | 77.618 | 0.359 | 9.322 | 0.436 | 0.810 | 56.185 | 79.021 | 0.359 | 9.343 | 0.443 | 0.803 | 55.399 | 77.313 | 0.354 | 9.212 | 0.434 | 0.812 |
| | MM | | 53.347 | 76.673 | 0.341 | 8.871 | 0.430 | 0.815 | 54.806 | 78.612 | 0.351 | 9.114 | 0.441 | 0.805 | 57.659 | 80.440 | 0.369 | 9.588 | 0.451 | 0.796 | 55.750 | 76.740 | 0.357 | 9.270 | 0.431 | 0.815 |
| QLD | MC | db1 | 89.288 | 124.338 | 0.317 | 8.072 | 0.384 | 0.853 | 90.654 | 126.591 | 0.322 | 8.196 | 0.391 | 0.847 | 91.749 | 128.222 | 0.325 | 8.295 | 0.396 | 0.843 | 90.058 | 125.770 | 0.319 | 8.142 | 0.388 | 0.849 |
| | MM | | 89.395 | 124.558 | 0.317 | 8.082 | 0.385 | 0.852 | 92.490 | 127.264 | 0.328 | 8.362 | 0.393 | 0.846 | 94.025 | 127.192 | 0.333 | 8.501 | 0.393 | 0.846 | 93.532 | 124.858 | 0.332 | 8.456 | 0.386 | 0.851 |
| | MC | db2 | 90.898 | 126.415 | 0.322 | 8.218 | 0.390 | 0.848 | 93.382 | 128.108 | 0.331 | 8.443 | 0.396 | 0.844 | 95.354 | 130.134 | 0.338 | 8.621 | 0.402 | 0.839 | 94.121 | 128.323 | 0.334 | 8.509 | 0.396 | 0.843 |
| | MM | | 90.392 | 125.310 | 0.321 | 8.172 | 0.387 | 0.850 | 92.981 | 127.747 | 0.330 | 8.406 | 0.394 | 0.844 | 93.280 | 126.817 | 0.331 | 8.433 | 0.392 | 0.847 | 92.652 | 124.025 | 0.329 | 8.377 | 0.383 | 0.853 |
| | MC | db3 | 90.420 | 125.778 | 0.321 | 8.175 | 0.388 | 0.849 | 91.928 | 127.371 | 0.326 | 8.311 | 0.393 | 0.845 | 93.708 | 129.028 | 0.332 | 8.472 | 0.398 | 0.841 | 93.082 | 127.543 | 0.330 | 8.416 | 0.394 | 0.845 |
| | MM | | 90.473 | 125.616 | 0.321 | 8.180 | 0.388 | 0.850 | 93.524 | 127.748 | 0.332 | 8.455 | 0.394 | 0.844 | 94.169 | 127.480 | 0.334 | 8.514 | 0.394 | 0.845 | 91.355 | 122.583 | 0.324 | 8.259 | 0.379 | 0.857 |
| | MC | db4 | 90.890 | 125.878 | 0.322 | 8.217 | 0.389 | 0.849 | 93.077 | 128.377 | 0.330 | 8.415 | 0.396 | 0.843 | 97.617 | 133.030 | 0.346 | 8.825 | 0.411 | 0.831 | 95.641 | 130.242 | 0.339 | 8.647 | 0.402 | 0.838 |
| | MM | | 90.330 | 125.538 | 0.320 | 8.167 | 0.388 | 0.850 | 92.931 | 126.839 | 0.330 | 8.402 | 0.392 | 0.847 | 94.418 | 127.984 | 0.335 | 8.536 | 0.395 | 0.844 | 91.082 | 122.107 | 0.323 | 8.235 | 0.377 | 0.858 |
| | MC | db5 | 91.134 | 125.324 | 0.323 | 8.239 | 0.387 | 0.850 | 94.860 | 130.371 | 0.336 | 8.576 | 0.403 | 0.838 | 105.262 | 142.646 | 0.373 | 9.517 | 0.440 | 0.806 | 94.600 | 127.926 | 0.336 | 8.553 | 0.395 | 0.844 |
| | MM | | 90.177 | 125.578 | 0.320 | 8.153 | 0.388 | 0.850 | 92.786 | 126.581 | 0.329 | 8.389 | 0.391 | 0.847 | 94.733 | 128.645 | 0.336 | 8.565 | 0.397 | 0.842 | 90.981 | 122.194 | 0.323 | 8.226 | 0.377 | 0.858 |
| | MC | db6 | 90.027 | 124.659 | 0.319 | 8.139 | 0.385 | 0.852 | 94.797 | 130.312 | 0.336 | 8.571 | 0.402 | 0.838 | 109.429 | 147.743 | 0.388 | 9.893 | 0.456 | 0.792 | 92.778 | 125.870 | 0.329 | 8.388 | 0.389 | 0.849 |
| | MM | | 90.003 | 125.306 | 0.319 | 8.137 | 0.387 | 0.850 | 92.805 | 126.663 | 0.329 | 8.390 | 0.391 | 0.847 | 95.370 | 129.078 | 0.338 | 8.622 | 0.399 | 0.841 | 92.089 | 123.469 | 0.327 | 8.326 | 0.381 | 0.855 |
| | MC | db7 | 91.620 | 126.335 | 0.325 | 8.283 | 0.390 | 0.848 | 95.968 | 131.591 | 0.340 | 8.676 | 0.406 | 0.835 | 109.722 | 150.108 | 0.389 | 9.920 | 0.463 | 0.785 | 93.265 | 127.799 | 0.331 | 8.432 | 0.395 | 0.844 |
| | MM | | 89.940 | 125.073 | 0.319 | 8.131 | 0.386 | 0.851 | 93.007 | 126.829 | 0.330 | 8.409 | 0.392 | 0.847 | 95.579 | 129.634 | 0.339 | 8.641 | 0.400 | 0.840 | 92.009 | 123.268 | 0.326 | 8.319 | 0.381 | 0.855 |

**Table B.6**
Overall performance of all wavelet approaches using paddings generated by $pad_{LR}$ with RF prediction model

| Dataset | Approach | $\psi$ | DL =1 | | | | | | DL =2 | | | | | | DL =3 | | | | | | DL =4 | | | | | |
|---|---|---|---|---|---|---|---|---|---|---|---|---|---|---|---|---|---|---|---|---|---|---|---|---|---|---|
| | | | MAE | RMSE | RAE | MRE | RRSE | $R^2$ | MAE | RMSE | RAE | MRE | RRSE | $R^2$ | MAE | RMSE | RAE | MRE | RRSE | $R^2$ | MAE | RMSE | RAE | MRE | RRSE | $R^2$ |
| NSW | MC | db1 | 53.249 | 76.628 | 0.341 | 8.855 | 0.430 | 0.815 | 53.779 | 77.618 | 0.344 | 8.943 | 0.436 | 0.810 | 54.344 | 78.156 | 0.348 | 9.037 | 0.439 | 0.808 | 54.048 | 77.672 | 0.346 | 8.987 | 0.436 | 0.810 |
| | MM | | 54.210 | 77.708 | 0.347 | 9.014 | 0.436 | 0.810 | 55.166 | 79.170 | 0.353 | 9.173 | 0.444 | 0.803 | 56.481 | 79.060 | 0.361 | 9.392 | 0.444 | 0.803 | 55.981 | 76.344 | 0.358 | 9.309 | 0.428 | 0.816 |
| | MC | db2 | 53.980 | 77.746 | 0.345 | 8.976 | 0.436 | 0.810 | 54.608 | 78.393 | 0.349 | 9.081 | 0.440 | 0.806 | 54.949 | 77.871 | 0.352 | 9.137 | 0.437 | 0.809 | 54.760 | 77.538 | 0.350 | 9.106 | 0.435 | 0.811 |
| | MM | | 53.602 | 77.044 | 0.343 | 8.913 | 0.432 | 0.813 | 54.939 | 78.547 | 0.352 | 9.136 | 0.441 | 0.806 | 56.029 | 79.125 | 0.358 | 9.317 | 0.444 | 0.803 | 55.946 | 76.988 | 0.358 | 9.303 | 0.432 | 0.813 |
| | MC | db3 | 54.249 | 77.700 | 0.347 | 9.021 | 0.436 | 0.810 | 54.587 | 77.532 | 0.349 | 9.077 | 0.435 | 0.811 | 56.233 | 79.107 | 0.360 | 9.351 | 0.444 | 0.803 | 56.083 | 79.425 | 0.359 | 9.326 | 0.446 | 0.801 |
| | MM | | 53.730 | 77.212 | 0.344 | 8.935 | 0.433 | 0.812 | 54.846 | 78.338 | 0.351 | 9.120 | 0.440 | 0.807 | 56.497 | 79.457 | 0.361 | 9.395 | 0.446 | 0.801 | 56.226 | 77.038 | 0.360 | 9.350 | 0.432 | 0.813 |
| | MC | db4 | 54.145 | 77.288 | 0.346 | 9.004 | 0.434 | 0.812 | 54.809 | 77.299 | 0.351 | 9.114 | 0.434 | 0.812 | 56.655 | 79.302 | 0.362 | 9.421 | 0.445 | 0.802 | 55.427 | 78.565 | 0.355 | 9.217 | 0.441 | 0.806 |
| | MM | | 53.716 | 77.206 | 0.344 | 8.932 | 0.433 | 0.812 | 54.886 | 78.607 | 0.351 | 9.127 | 0.441 | 0.805 | 56.510 | 79.379 | 0.362 | 9.397 | 0.445 | 0.802 | 55.400 | 76.168 | 0.354 | 9.212 | 0.427 | 0.817 |
| | MC | db5 | 53.905 | 77.088 | 0.345 | 8.964 | 0.433 | 0.813 | 55.654 | 78.640 | 0.356 | 9.255 | 0.441 | 0.805 | 56.078 | 78.622 | 0.359 | 9.325 | 0.441 | 0.805 | 55.542 | 78.425 | 0.355 | 9.236 | 0.440 | 0.806 |
| | MM | | 53.847 | 77.259 | 0.345 | 8.954 | 0.434 | 0.812 | 54.974 | 78.672 | 0.352 | 9.141 | 0.441 | 0.805 | 56.807 | 79.434 | 0.363 | 9.446 | 0.446 | 0.801 | 55.417 | 75.936 | 0.355 | 9.215 | 0.426 | 0.818 |
| | MC | db6 | 54.849 | 77.746 | 0.351 | 9.121 | 0.436 | 0.810 | 56.249 | 79.077 | 0.360 | 9.353 | 0.444 | 0.803 | 56.312 | 78.942 | 0.360 | 9.364 | 0.443 | 0.804 | 55.777 | 78.321 | 0.357 | 9.275 | 0.440 | 0.807 |
| | MM | | 53.817 | 77.191 | 0.344 | 8.949 | 0.433 | 0.812 | 54.987 | 78.742 | 0.352 | 9.144 | 0.442 | 0.805 | 57.200 | 79.689 | 0.366 | 9.512 | 0.447 | 0.800 | 55.560 | 75.839 | 0.355 | 9.239 | 0.426 | 0.819 |
| | MC | db7 | 55.172 | 77.806 | 0.353 | 9.174 | 0.437 | 0.809 | 56.519 | 78.244 | 0.362 | 9.398 | 0.439 | 0.807 | 56.539 | 78.886 | 0.362 | 9.402 | 0.443 | 0.804 | 55.494 | 77.823 | 0.355 | 9.228 | 0.437 | 0.809 |
| | MM | | 53.663 | 76.857 | 0.343 | 8.923 | 0.431 | 0.814 | 55.158 | 78.898 | 0.353 | 9.172 | 0.443 | 0.804 | 57.249 | 79.725 | 0.366 | 9.520 | 0.447 | 0.800 | 55.602 | 75.773 | 0.356 | 9.246 | 0.425 | 0.819 |
| QLD | MC | db1 | 89.717 | 124.722 | 0.318 | 8.111 | 0.385 | 0.852 | 90.769 | 126.614 | 0.322 | 8.206 | 0.391 | 0.847 | 91.730 | 128.037 | 0.325 | 8.293 | 0.395 | 0.844 | 90.120 | 125.689 | 0.320 | 8.148 | 0.388 | 0.849 |
| | MM | | 89.579 | 124.767 | 0.318 | 8.099 | 0.385 | 0.852 | 92.428 | 127.404 | 0.328 | 8.356 | 0.393 | 0.845 | 93.894 | 127.232 | 0.333 | 8.489 | 0.393 | 0.846 | 93.872 | 125.216 | 0.333 | 8.487 | 0.387 | 0.851 |
| | MC | db2 | 90.778 | 126.136 | 0.322 | 8.207 | 0.389 | 0.848 | 93.186 | 127.810 | 0.330 | 8.425 | 0.395 | 0.844 | 96.128 | 130.982 | 0.341 | 8.691 | 0.404 | 0.836 | 95.414 | 129.278 | 0.338 | 8.626 | 0.399 | 0.841 |
| | MM | | 90.416 | 125.446 | 0.321 | 8.174 | 0.387 | 0.850 | 93.440 | 128.405 | 0.331 | 8.448 | 0.396 | 0.843 | 93.957 | 127.306 | 0.333 | 8.495 | 0.393 | 0.845 | 93.611 | 125.090 | 0.332 | 8.463 | 0.386 | 0.851 |
| | MC | db3 | 90.163 | 125.442 | 0.320 | 8.152 | 0.387 | 0.850 | 91.716 | 126.958 | 0.325 | 8.292 | 0.392 | 0.846 | 93.895 | 129.179 | 0.333 | 8.489 | 0.399 | 0.841 | 92.598 | 126.578 | 0.328 | 8.372 | 0.391 | 0.847 |
| | MM | | 90.250 | 125.231 | 0.320 | 8.159 | 0.387 | 0.850 | 93.662 | 128.072 | 0.332 | 8.468 | 0.395 | 0.844 | 94.361 | 127.471 | 0.335 | 8.531 | 0.394 | 0.845 | 93.657 | 125.007 | 0.332 | 8.467 | 0.386 | 0.851 |
| | MC | db4 | 90.668 | 125.644 | 0.322 | 8.197 | 0.388 | 0.849 | 92.697 | 128.027 | 0.329 | 8.381 | 0.395 | 0.844 | 95.723 | 130.501 | 0.339 | 8.654 | 0.403 | 0.838 | 97.661 | 132.894 | 0.346 | 8.830 | 0.410 | 0.832 |
| | MM | | 90.179 | 125.261 | 0.320 | 8.153 | 0.387 | 0.850 | 93.257 | 127.431 | 0.331 | 8.431 | 0.393 | 0.845 | 94.908 | 128.274 | 0.337 | 8.581 | 0.396 | 0.843 | 94.024 | 125.237 | 0.333 | 8.501 | 0.387 | 0.850 |
| | MC | db5 | 90.856 | 124.894 | 0.322 | 8.214 | 0.386 | 0.851 | 94.663 | 129.687 | 0.336 | 8.558 | 0.400 | 0.840 | 102.731 | 139.787 | 0.364 | 9.288 | 0.432 | 0.814 | 98.944 | 134.480 | 0.351 | 8.946 | 0.415 | 0.828 |
| | MM | | 90.097 | 125.328 | 0.320 | 8.146 | 0.387 | 0.850 | 92.777 | 126.889 | 0.329 | 8.388 | 0.392 | 0.846 | 94.782 | 128.415 | 0.336 | 8.569 | 0.397 | 0.843 | 93.588 | 124.858 | 0.332 | 8.461 | 0.386 | 0.851 |
| | MC | db6 | 90.024 | 124.663 | 0.319 | 8.139 | 0.385 | 0.852 | 94.450 | 129.547 | 0.335 | 8.539 | 0.400 | 0.840 | 109.276 | 147.914 | 0.388 | 9.880 | 0.457 | 0.791 | 100.439 | 135.863 | 0.356 | 9.081 | 0.420 | 0.824 |
| | MM | | 90.198 | 125.445 | 0.320 | 8.155 | 0.387 | 0.850 | 92.798 | 126.936 | 0.329 | 8.390 | 0.392 | 0.846 | 94.858 | 128.477 | 0.336 | 8.576 | 0.397 | 0.843 | 92.942 | 124.046 | 0.330 | 8.403 | 0.383 | 0.853 |
| | MC | db7 | 91.202 | 125.690 | 0.323 | 8.246 | 0.388 | 0.849 | 96.418 | 131.694 | 0.342 | 8.717 | 0.407 | 0.835 | 107.962 | 147.554 | 0.383 | 9.761 | 0.456 | 0.792 | 100.963 | 136.400 | 0.358 | 9.128 | 0.421 | 0.823 |
| | MM | | 90.033 | 125.230 | 0.319 | 8.140 | 0.387 | 0.850 | 93.002 | 126.926 | 0.330 | 8.408 | 0.392 | 0.846 | 95.194 | 128.873 | 0.338 | 8.606 | 0.398 | 0.842 | 92.692 | 123.097 | 0.329 | 8.380 | 0.380 | 0.856 |

**Table B.7**
Overall performance of all wavelet approaches using paddings generated by $pad_{REP}$ with CNN prediction model

| Dataset | Approach | $\psi$ | DL =1 | | | | | | DL =2 | | | | | | DL =3 | | | | | | DL =4 | | | | | |
|---|---|---|---|---|---|---|---|---|---|---|---|---|---|---|---|---|---|---|---|---|---|---|---|---|---|---|
| | | | MAE | RMSE | RAE | MRE | RRSE | $R^2$ | MAE | RMSE | RAE | MRE | RRSE | $R^2$ | MAE | RMSE | RAE | MRE | RRSE | $R^2$ | MAE | RMSE | RAE | MRE | RRSE | $R^2$ |
| | MC | db1 | 53.582 | 76.434 | 0.343 | 8.910 | 0.429 | 0.816 | 55.045 | 76.678 | 0.352 | 9.153 | 0.430 | 0.815 | 53.472 | 76.355 | 0.342 | 8.892 | 0.429 | 0.816 | 54.320 | 75.765 | 0.348 | 9.033 | 0.425 | 0.819 |
| | MI | | 53.208 | 76.621 | 0.340 | 8.848 | 0.430 | 0.815 | 53.836 | 76.544 | 0.344 | 8.952 | 0.430 | 0.815 | 54.973 | 76.242 | 0.352 | 9.141 | 0.428 | 0.817 | 55.125 | 76.985 | 0.353 | 9.167 | 0.432 | 0.813 |
| | MM | | 53.245 | 75.499 | 0.341 | 8.854 | 0.424 | 0.820 | 52.629 | 74.699 | 0.337 | 8.751 | 0.419 | 0.824 | 53.823 | 75.125 | 0.344 | 8.950 | 0.422 | 0.822 | 56.133 | 76.821 | 0.359 | 9.334 | 0.431 | 0.814 |
| | MC | db2 | 54.074 | 76.729 | 0.346 | 8.992 | 0.431 | 0.815 | 54.882 | 78.099 | 0.351 | 9.126 | 0.438 | 0.808 | 54.854 | 77.377 | 0.351 | 9.122 | 0.434 | 0.811 | 54.704 | 77.787 | 0.350 | 9.097 | 0.437 | 0.809 |
| | MI | | 53.741 | 76.083 | 0.344 | 8.936 | 0.427 | 0.818 | 53.972 | 76.146 | 0.345 | 8.975 | 0.427 | 0.817 | 55.580 | 77.475 | 0.356 | 9.242 | 0.435 | 0.811 | 55.450 | 77.588 | 0.355 | 9.221 | 0.435 | 0.810 |
| | MM | | 52.771 | 75.388 | 0.338 | 8.775 | 0.423 | 0.821 | 52.995 | 75.081 | 0.339 | 8.812 | 0.421 | 0.822 | 53.603 | 75.141 | 0.343 | 8.914 | 0.422 | 0.822 | 56.413 | 77.341 | 0.361 | 9.381 | 0.434 | 0.812 |
| | MC | db3 | 53.507 | 75.775 | 0.342 | 8.898 | 0.425 | 0.819 | 54.812 | 77.475 | 0.351 | 9.115 | 0.435 | 0.811 | 54.943 | 77.976 | 0.352 | 9.136 | 0.438 | 0.809 | 56.143 | 79.551 | 0.359 | 9.336 | 0.446 | 0.801 |
| | MI | | 53.722 | 75.936 | 0.344 | 8.933 | 0.426 | 0.818 | 54.881 | 78.130 | 0.351 | 9.126 | 0.438 | 0.808 | 56.934 | 79.684 | 0.364 | 9.467 | 0.447 | 0.800 | 55.719 | 78.157 | 0.356 | 9.265 | 0.439 | 0.808 |
| | MM | | 52.825 | 75.742 | 0.338 | 8.784 | 0.425 | 0.819 | 53.066 | 74.784 | 0.340 | 8.824 | 0.420 | 0.824 | 53.832 | 75.466 | 0.344 | 8.951 | 0.424 | 0.821 | 56.946 | 77.994 | 0.364 | 9.469 | 0.438 | 0.808 |
| | MC | db4 | 55.121 | 76.658 | 0.353 | 9.166 | 0.430 | 0.815 | 56.847 | 78.873 | 0.364 | 9.453 | 0.443 | 0.804 | 56.674 | 79.603 | 0.363 | 9.424 | 0.447 | 0.800 | 55.616 | 78.562 | 0.356 | 9.248 | 0.441 | 0.806 |
| NSW | MI | | 53.442 | 76.002 | 0.342 | 8.887 | 0.427 | 0.818 | 56.229 | 78.582 | 0.360 | 9.350 | 0.441 | 0.806 | 57.255 | 79.047 | 0.366 | 9.521 | 0.444 | 0.803 | 55.466 | 79.233 | 0.355 | 9.223 | 0.445 | 0.802 |
| | MM | | 53.019 | 75.645 | 0.339 | 8.816 | 0.425 | 0.820 | 53.063 | 74.497 | 0.339 | 8.824 | 0.418 | 0.825 | 53.985 | 75.624 | 0.345 | 8.977 | 0.424 | 0.820 | 56.825 | 77.647 | 0.364 | 9.449 | 0.436 | 0.810 |
| | MC | db5 | 54.772 | 76.209 | 0.350 | 9.108 | 0.428 | 0.817 | 57.308 | 80.661 | 0.367 | 9.530 | 0.453 | 0.795 | 56.722 | 79.465 | 0.363 | 9.432 | 0.446 | 0.801 | 55.401 | 77.520 | 0.354 | 9.212 | 0.435 | 0.811 |
| | MI | | 54.186 | 75.690 | 0.347 | 9.010 | 0.425 | 0.820 | 57.066 | 79.897 | 0.365 | 9.489 | 0.448 | 0.799 | 56.811 | 79.295 | 0.363 | 9.447 | 0.445 | 0.802 | 54.470 | 76.715 | 0.349 | 9.058 | 0.431 | 0.815 |
| | MM | | 52.958 | 75.381 | 0.339 | 8.806 | 0.423 | 0.821 | 52.595 | 74.931 | 0.337 | 8.746 | 0.421 | 0.823 | 54.289 | 75.522 | 0.347 | 9.028 | 0.424 | 0.820 | 56.965 | 77.913 | 0.364 | 9.473 | 0.437 | 0.809 |
| | MC | db6 | 54.844 | 77.270 | 0.351 | 9.120 | 0.434 | 0.812 | 56.691 | 80.520 | 0.363 | 9.427 | 0.452 | 0.796 | 56.816 | 80.514 | 0.364 | 9.448 | 0.452 | 0.796 | 55.338 | 77.478 | 0.354 | 9.202 | 0.435 | 0.811 |
| | MI | | 54.722 | 77.288 | 0.350 | 9.100 | 0.434 | 0.812 | 58.476 | 81.886 | 0.374 | 9.724 | 0.460 | 0.789 | 56.811 | 79.926 | 0.363 | 9.447 | 0.449 | 0.799 | 55.682 | 76.949 | 0.356 | 9.259 | 0.432 | 0.814 |
| | MM | | 52.931 | 75.441 | 0.339 | 8.802 | 0.423 | 0.821 | 53.251 | 74.799 | 0.341 | 8.855 | 0.420 | 0.824 | 54.296 | 75.508 | 0.347 | 9.029 | 0.424 | 0.820 | 57.067 | 77.754 | 0.365 | 9.489 | 0.436 | 0.810 |
| | MC | db7 | 55.673 | 78.189 | 0.356 | 9.258 | 0.439 | 0.807 | 57.628 | 81.320 | 0.369 | 9.583 | 0.456 | 0.792 | 56.475 | 79.593 | 0.361 | 9.391 | 0.447 | 0.800 | 54.831 | 77.118 | 0.351 | 9.118 | 0.433 | 0.813 |
| | MI | | 55.206 | 77.213 | 0.353 | 9.180 | 0.433 | 0.812 | 57.653 | 80.741 | 0.369 | 9.587 | 0.453 | 0.795 | 56.850 | 80.135 | 0.364 | 9.453 | 0.450 | 0.798 | 55.134 | 77.539 | 0.353 | 9.168 | 0.435 | 0.811 |
| | MM | | 52.958 | 75.466 | 0.339 | 8.806 | 0.424 | 0.821 | 53.782 | 77.883 | 0.344 | 8.943 | 0.437 | 0.809 | 54.270 | 75.794 | 0.347 | 9.024 | 0.425 | 0.819 | 57.200 | 77.993 | 0.366 | 9.512 | 0.438 | 0.808 |
| | MC | db1 | 92.768 | 126.905 | 0.329 | 8.387 | 0.392 | 0.846 | 93.535 | 126.694 | 0.332 | 8.456 | 0.391 | 0.847 | 94.192 | 127.281 | 0.334 | 8.516 | 0.393 | 0.846 | 95.126 | 128.491 | 0.337 | 8.600 | 0.397 | 0.843 |
| | MI | | 90.749 | 124.058 | 0.322 | 8.205 | 0.383 | 0.853 | 91.653 | 124.734 | 0.325 | 8.286 | 0.385 | 0.852 | 94.125 | 126.735 | 0.334 | 8.510 | 0.391 | 0.847 | 96.034 | 129.154 | 0.341 | 8.682 | 0.399 | 0.841 |
| | MM | | 89.799 | 123.045 | 0.318 | 8.119 | 0.380 | 0.856 | 88.336 | 121.315 | 0.313 | 7.986 | 0.375 | 0.860 | 89.894 | 121.999 | 0.319 | 8.127 | 0.377 | 0.858 | 90.217 | 120.946 | 0.320 | 8.156 | 0.373 | 0.861 |
| | MC | db2 | 90.660 | 124.274 | 0.322 | 8.197 | 0.384 | 0.853 | 92.387 | 125.540 | 0.328 | 8.353 | 0.388 | 0.850 | 95.162 | 129.149 | 0.338 | 8.604 | 0.399 | 0.841 | 96.511 | 130.332 | 0.342 | 8.726 | 0.402 | 0.838 |
| | MI | | 90.810 | 123.592 | 0.322 | 8.210 | 0.382 | 0.854 | 90.450 | 124.222 | 0.321 | 8.178 | 0.384 | 0.853 | 93.294 | 126.105 | 0.331 | 8.435 | 0.389 | 0.848 | 94.503 | 126.925 | 0.335 | 8.544 | 0.392 | 0.846 |
| | MM | | 88.897 | 121.655 | 0.315 | 8.037 | 0.376 | 0.859 | 87.947 | 120.870 | 0.312 | 7.951 | 0.373 | 0.861 | 89.785 | 123.162 | 0.318 | 8.117 | 0.380 | 0.855 | 95.015 | 124.803 | 0.337 | 8.590 | 0.385 | 0.851 |
| | MC | db3 | 90.619 | 124.068 | 0.321 | 8.193 | 0.383 | 0.853 | 95.527 | 128.466 | 0.339 | 8.637 | 0.397 | 0.843 | 95.053 | 128.020 | 0.337 | 8.594 | 0.395 | 0.844 | 100.705 | 137.155 | 0.357 | 9.105 | 0.424 | 0.821 |
| | MI | | 91.026 | 123.941 | 0.323 | 8.230 | 0.383 | 0.854 | 89.691 | 122.702 | 0.318 | 8.109 | 0.379 | 0.856 | 96.635 | 129.374 | 0.343 | 8.737 | 0.399 | 0.840 | 94.892 | 128.096 | 0.337 | 8.579 | 0.396 | 0.844 |
| | MM | | 89.419 | 122.673 | 0.317 | 8.084 | 0.379 | 0.857 | 88.055 | 120.643 | 0.312 | 7.961 | 0.373 | 0.861 | 89.488 | 122.807 | 0.317 | 8.091 | 0.379 | 0.856 | 92.962 | 123.474 | 0.330 | 8.405 | 0.381 | 0.855 |
| | MC | db4 | 94.104 | 128.736 | 0.334 | 8.508 | 0.398 | 0.842 | 95.450 | 128.744 | 0.339 | 8.630 | 0.398 | 0.842 | 100.558 | 135.255 | 0.357 | 9.091 | 0.418 | 0.826 | 102.889 | 137.694 | 0.365 | 9.302 | 0.425 | 0.819 |
| QLD | MI | | 91.803 | 124.758 | 0.326 | 8.300 | 0.385 | 0.852 | 96.693 | 131.930 | 0.343 | 8.742 | 0.407 | 0.834 | 97.621 | 132.751 | 0.346 | 8.826 | 0.410 | 0.832 | 93.881 | 127.715 | 0.333 | 8.488 | 0.394 | 0.844 |
| | MM | | 89.429 | 122.679 | 0.317 | 8.085 | 0.379 | 0.857 | 88.094 | 120.552 | 0.312 | 7.965 | 0.372 | 0.861 | 89.182 | 122.714 | 0.316 | 8.063 | 0.379 | 0.856 | 93.275 | 123.677 | 0.331 | 8.433 | 0.382 | 0.854 |
| | MC | db5 | 91.443 | 125.334 | 0.324 | 8.267 | 0.387 | 0.850 | 98.537 | 131.028 | 0.349 | 8.909 | 0.405 | 0.836 | 110.013 | 147.683 | 0.390 | 9.946 | 0.456 | 0.792 | 100.812 | 136.290 | 0.358 | 9.114 | 0.421 | 0.823 |
| | MI | | 91.938 | 124.884 | 0.326 | 8.312 | 0.386 | 0.851 | 93.616 | 128.677 | 0.332 | 8.464 | 0.397 | 0.842 | 105.604 | 142.929 | 0.375 | 9.548 | 0.441 | 0.805 | 97.553 | 131.643 | 0.346 | 8.820 | 0.406 | 0.835 |
| | MM | | 89.654 | 122.326 | 0.318 | 8.106 | 0.378 | 0.857 | 87.913 | 120.325 | 0.312 | 7.948 | 0.372 | 0.862 | 89.751 | 123.207 | 0.318 | 8.114 | 0.380 | 0.855 | 92.428 | 123.301 | 0.328 | 8.356 | 0.381 | 0.855 |
| | MC | db6 | 92.662 | 126.851 | 0.329 | 8.378 | 0.392 | 0.847 | 101.243 | 136.511 | 0.359 | 9.153 | 0.422 | 0.822 | 108.245 | 147.081 | 0.384 | 9.786 | 0.454 | 0.794 | 93.427 | 127.146 | 0.331 | 8.447 | 0.393 | 0.846 |
| | MI | | 93.848 | 126.074 | 0.333 | 8.485 | 0.389 | 0.848 | 100.234 | 137.077 | 0.355 | 9.062 | 0.423 | 0.821 | 104.969 | 143.235 | 0.372 | 9.490 | 0.442 | 0.804 | 95.936 | 129.335 | 0.340 | 8.673 | 0.399 | 0.841 |
| | MM | | 89.820 | 122.614 | 0.319 | 8.121 | 0.379 | 0.857 | 87.844 | 120.286 | 0.312 | 7.942 | 0.371 | 0.862 | 89.677 | 123.145 | 0.318 | 8.108 | 0.380 | 0.855 | 88.435 | 120.131 | 0.314 | 7.995 | 0.371 | 0.862 |
| | MC | db7 | 93.501 | 126.860 | 0.332 | 8.453 | 0.392 | 0.847 | 99.201 | 134.337 | 0.352 | 8.969 | 0.415 | 0.828 | 106.104 | 143.516 | 0.376 | 9.593 | 0.443 | 0.804 | 95.772 | 130.069 | 0.340 | 8.659 | 0.402 | 0.839 |
| | MI | | 93.172 | 126.743 | 0.330 | 8.424 | 0.391 | 0.847 | 95.460 | 130.888 | 0.339 | 8.630 | 0.404 | 0.837 | 103.483 | 140.688 | 0.367 | 9.356 | 0.434 | 0.811 | 93.585 | 127.561 | 0.332 | 8.461 | 0.394 | 0.845 |
| | MM | | 89.829 | 122.664 | 0.319 | 8.121 | 0.379 | 0.857 | 87.507 | 119.951 | 0.310 | 7.911 | 0.370 | 0.863 | 88.248 | 121.239 | 0.313 | 7.978 | 0.374 | 0.860 | 92.500 | 123.241 | 0.328 | 8.363 | 0.381 | 0.855 |

**Table B.8**
Overall performance of all wavelet approaches using paddings generated by $pad_{LR}$ with CNN prediction model

| Dataset | Approach | $\psi$ | DL =1 | | | | | | DL =2 | | | | | | DL =3 | | | | | | DL =4 | | | | | |
|---|---|---|---|---|---|---|---|---|---|---|---|---|---|---|---|---|---|---|---|---|---|---|---|---|---|---|
| | | | MAE | RMSE | RAE | MRE | RRSE | $R^2$ | MAE | RMSE | RAE | MRE | RRSE | $R^2$ | MAE | RMSE | RAE | MRE | RRSE | $R^2$ | MAE | RMSE | RAE | MRE | RRSE | $R^2$ |
| | MC | db1 | 53.582 | 77.292 | 0.343 | 8.910 | 0.434 | 0.812 | 54.752 | 77.201 | 0.350 | 9.104 | 0.433 | 0.812 | 55.676 | 76.738 | 0.356 | 9.258 | 0.431 | 0.815 | 55.984 | 77.438 | 0.358 | 9.309 | 0.435 | 0.811 |
| | MI | | 53.350 | 76.237 | 0.341 | 8.871 | 0.428 | 0.817 | 54.144 | 77.363 | 0.346 | 9.003 | 0.434 | 0.812 | 56.122 | 80.449 | 0.359 | 9.332 | 0.451 | 0.796 | 56.630 | 79.304 | 0.362 | 9.417 | 0.445 | 0.802 |
| | MM | | 52.842 | 75.945 | 0.338 | 8.787 | 0.426 | 0.818 | 53.052 | 75.219 | 0.339 | 8.822 | 0.422 | 0.822 | 53.680 | 75.210 | 0.343 | 8.926 | 0.422 | 0.822 | 55.744 | 76.581 | 0.357 | 9.269 | 0.430 | 0.815 |
| | MC | db2 | 53.951 | 77.901 | 0.345 | 8.971 | 0.437 | 0.809 | 54.362 | 76.799 | 0.348 | 9.040 | 0.431 | 0.814 | 54.101 | 77.383 | 0.346 | 8.996 | 0.434 | 0.811 | 55.010 | 77.830 | 0.352 | 9.147 | 0.437 | 0.809 |
| | MI | | 53.688 | 76.656 | 0.344 | 8.928 | 0.430 | 0.815 | 54.021 | 76.264 | 0.346 | 8.983 | 0.428 | 0.817 | 55.381 | 77.475 | 0.354 | 9.209 | 0.435 | 0.811 | 56.410 | 77.650 | 0.361 | 9.380 | 0.436 | 0.810 |
| | MM | | 52.969 | 76.291 | 0.339 | 8.808 | 0.428 | 0.817 | 53.047 | 75.406 | 0.339 | 8.821 | 0.423 | 0.821 | 54.114 | 75.997 | 0.346 | 8.999 | 0.426 | 0.818 | 55.946 | 76.537 | 0.358 | 9.303 | 0.430 | 0.816 |
| | MC | db3 | 53.708 | 77.474 | 0.344 | 8.931 | 0.435 | 0.811 | 54.197 | 77.632 | 0.347 | 9.012 | 0.436 | 0.810 | 55.839 | 78.210 | 0.357 | 9.285 | 0.439 | 0.807 | 57.031 | 79.967 | 0.365 | 9.483 | 0.449 | 0.799 |
| | MI | | 53.646 | 76.568 | 0.343 | 8.921 | 0.430 | 0.815 | 54.910 | 77.251 | 0.351 | 9.131 | 0.434 | 0.812 | 56.272 | 78.234 | 0.360 | 9.357 | 0.439 | 0.807 | 56.808 | 79.429 | 0.363 | 9.446 | 0.446 | 0.801 |
| | MM | | 52.927 | 76.475 | 0.339 | 8.801 | 0.429 | 0.816 | 53.314 | 75.513 | 0.341 | 8.865 | 0.424 | 0.820 | 54.122 | 76.115 | 0.346 | 9.000 | 0.427 | 0.818 | 55.954 | 76.848 | 0.358 | 9.304 | 0.431 | 0.814 |
| | MC | db4 | 54.876 | 76.392 | 0.351 | 9.125 | 0.429 | 0.816 | 56.730 | 78.699 | 0.363 | 9.433 | 0.442 | 0.805 | 55.996 | 79.767 | 0.358 | 9.311 | 0.448 | 0.800 | 56.008 | 79.832 | 0.358 | 9.313 | 0.448 | 0.799 |
| NSW | MI | | 53.470 | 76.512 | 0.342 | 8.891 | 0.429 | 0.816 | 56.141 | 78.183 | 0.359 | 9.335 | 0.439 | 0.807 | 57.643 | 79.926 | 0.369 | 9.585 | 0.449 | 0.799 | 59.227 | 82.792 | 0.379 | 9.849 | 0.465 | 0.784 |
| | MM | | 52.807 | 76.131 | 0.338 | 8.781 | 0.427 | 0.817 | 52.812 | 74.596 | 0.338 | 8.782 | 0.419 | 0.825 | 54.398 | 76.205 | 0.348 | 9.046 | 0.428 | 0.817 | 56.149 | 77.189 | 0.359 | 9.337 | 0.433 | 0.812 |
| | MC | db5 | 54.205 | 76.924 | 0.347 | 9.014 | 0.432 | 0.814 | 57.844 | 79.931 | 0.370 | 9.619 | 0.449 | 0.799 | 57.979 | 81.362 | 0.371 | 9.641 | 0.457 | 0.792 | 57.491 | 81.576 | 0.368 | 9.560 | 0.458 | 0.790 |
| | MI | | 54.269 | 76.727 | 0.347 | 9.024 | 0.431 | 0.815 | 58.296 | 80.199 | 0.373 | 9.694 | 0.450 | 0.797 | 56.882 | 80.120 | 0.364 | 9.459 | 0.450 | 0.798 | 59.260 | 82.400 | 0.379 | 9.854 | 0.462 | 0.786 |
| | MM | | 52.718 | 76.259 | 0.337 | 8.766 | 0.428 | 0.817 | 53.475 | 75.708 | 0.342 | 8.892 | 0.425 | 0.819 | 54.087 | 75.772 | 0.346 | 8.994 | 0.425 | 0.819 | 56.082 | 77.639 | 0.359 | 9.326 | 0.436 | 0.810 |
| | MC | db6 | 55.279 | 77.863 | 0.354 | 9.192 | 0.437 | 0.809 | 56.342 | 79.745 | 0.360 | 9.369 | 0.448 | 0.800 | 57.071 | 80.523 | 0.365 | 9.490 | 0.452 | 0.796 | 57.816 | 81.227 | 0.370 | 9.614 | 0.456 | 0.792 |
| | MI | | 54.095 | 76.608 | 0.346 | 8.995 | 0.430 | 0.815 | 57.323 | 80.532 | 0.367 | 9.532 | 0.452 | 0.796 | 56.832 | 80.839 | 0.364 | 9.450 | 0.454 | 0.794 | 58.077 | 81.069 | 0.372 | 9.657 | 0.455 | 0.793 |
| | MM | | 52.738 | 76.313 | 0.337 | 8.770 | 0.428 | 0.817 | 53.578 | 75.668 | 0.343 | 8.909 | 0.425 | 0.820 | 54.174 | 75.873 | 0.347 | 9.008 | 0.426 | 0.819 | 56.026 | 77.196 | 0.358 | 9.316 | 0.433 | 0.812 |
| | MC | db7 | 55.464 | 77.544 | 0.355 | 9.223 | 0.435 | 0.811 | 57.817 | 81.564 | 0.370 | 9.614 | 0.458 | 0.790 | 56.790 | 80.645 | 0.363 | 9.443 | 0.453 | 0.795 | 56.405 | 79.701 | 0.361 | 9.379 | 0.447 | 0.800 |
| | MI | | 55.111 | 77.659 | 0.353 | 9.164 | 0.436 | 0.810 | 57.687 | 79.716 | 0.369 | 9.593 | 0.447 | 0.800 | 58.376 | 81.389 | 0.373 | 9.707 | 0.457 | 0.791 | 57.140 | 82.709 | 0.366 | 9.502 | 0.464 | 0.785 |
| | MM | | 52.766 | 76.360 | 0.338 | 8.774 | 0.429 | 0.816 | 52.973 | 75.376 | 0.339 | 8.809 | 0.423 | 0.821 | 54.438 | 76.388 | 0.348 | 9.052 | 0.429 | 0.816 | 56.153 | 77.232 | 0.359 | 9.338 | 0.433 | 0.812 |
| | MC | db1 | 92.621 | 126.003 | 0.328 | 8.374 | 0.389 | 0.849 | 94.312 | 127.352 | 0.334 | 8.527 | 0.393 | 0.845 | 94.999 | 128.249 | 0.337 | 8.589 | 0.396 | 0.843 | 96.824 | 130.795 | 0.343 | 8.754 | 0.404 | 0.837 |
| | MI | | 91.771 | 124.392 | 0.325 | 8.297 | 0.384 | 0.852 | 92.619 | 125.038 | 0.328 | 8.374 | 0.386 | 0.851 | 91.386 | 123.532 | 0.324 | 8.262 | 0.381 | 0.855 | 96.055 | 129.355 | 0.341 | 8.684 | 0.399 | 0.840 |
| | MM | | 89.332 | 122.559 | 0.317 | 8.076 | 0.378 | 0.857 | 87.928 | 120.863 | 0.312 | 7.950 | 0.373 | 0.861 | 88.847 | 121.470 | 0.315 | 8.033 | 0.375 | 0.859 | 94.188 | 123.630 | 0.334 | 8.515 | 0.382 | 0.854 |
| | MC | db2 | 90.124 | 123.944 | 0.320 | 8.148 | 0.383 | 0.854 | 94.192 | 127.212 | 0.334 | 8.516 | 0.393 | 0.846 | 95.057 | 129.074 | 0.337 | 8.594 | 0.399 | 0.841 | 98.112 | 132.122 | 0.348 | 8.870 | 0.408 | 0.834 |
| | MI | | 92.485 | 125.118 | 0.328 | 8.362 | 0.386 | 0.851 | 93.612 | 128.063 | 0.332 | 8.463 | 0.395 | 0.844 | 93.156 | 126.058 | 0.330 | 8.422 | 0.389 | 0.848 | 95.171 | 128.234 | 0.338 | 8.604 | 0.396 | 0.843 |
| | MM | | 88.292 | 121.172 | 0.313 | 7.982 | 0.374 | 0.860 | 87.965 | 120.950 | 0.312 | 7.953 | 0.373 | 0.861 | 88.843 | 122.306 | 0.315 | 8.032 | 0.378 | 0.857 | 94.377 | 123.963 | 0.335 | 8.533 | 0.383 | 0.853 |
| | MC | db3 | 92.284 | 127.385 | 0.327 | 8.343 | 0.393 | 0.845 | 95.340 | 128.721 | 0.338 | 8.620 | 0.397 | 0.842 | 94.848 | 128.088 | 0.336 | 8.575 | 0.396 | 0.844 | 99.454 | 135.291 | 0.353 | 8.992 | 0.418 | 0.825 |
| | MI | | 92.636 | 125.695 | 0.329 | 8.375 | 0.388 | 0.849 | 90.222 | 124.417 | 0.320 | 8.157 | 0.384 | 0.852 | 95.699 | 128.629 | 0.339 | 8.652 | 0.397 | 0.842 | 99.903 | 134.142 | 0.354 | 9.032 | 0.414 | 0.828 |
| | MM | | 88.916 | 122.180 | 0.315 | 8.039 | 0.377 | 0.858 | 87.365 | 120.456 | 0.310 | 7.899 | 0.372 | 0.862 | 88.286 | 121.543 | 0.313 | 7.982 | 0.375 | 0.859 | 93.733 | 123.405 | 0.332 | 8.474 | 0.381 | 0.855 |
| | MC | db4 | 93.678 | 129.245 | 0.332 | 8.469 | 0.399 | 0.841 | 97.126 | 131.053 | 0.344 | 8.781 | 0.405 | 0.836 | 100.646 | 135.367 | 0.357 | 9.099 | 0.418 | 0.825 | 103.187 | 138.529 | 0.366 | 9.329 | 0.428 | 0.817 |
| QLD | MI | | 94.367 | 127.078 | 0.335 | 8.532 | 0.392 | 0.846 | 97.713 | 132.731 | 0.347 | 8.834 | 0.410 | 0.832 | 95.588 | 132.152 | 0.339 | 8.642 | 0.408 | 0.833 | 99.970 | 133.664 | 0.355 | 9.038 | 0.413 | 0.830 |
| | MM | | 89.142 | 122.320 | 0.316 | 8.059 | 0.378 | 0.857 | 86.795 | 119.803 | 0.308 | 7.847 | 0.370 | 0.863 | 87.957 | 121.561 | 0.312 | 7.952 | 0.375 | 0.859 | 92.357 | 122.508 | 0.328 | 8.350 | 0.378 | 0.857 |
| | MC | db5 | 92.255 | 126.748 | 0.327 | 8.341 | 0.391 | 0.847 | 96.790 | 129.738 | 0.343 | 8.751 | 0.401 | 0.840 | 106.508 | 144.381 | 0.378 | 9.629 | 0.446 | 0.801 | 110.636 | 148.729 | 0.392 | 10.003 | 0.459 | 0.789 |
| | MI | | 94.559 | 127.430 | 0.335 | 8.549 | 0.393 | 0.845 | 96.744 | 131.599 | 0.343 | 8.747 | 0.406 | 0.835 | 105.526 | 142.211 | 0.374 | 9.541 | 0.439 | 0.807 | 124.997 | 165.932 | 0.443 | 11.301 | 0.512 | 0.737 |
| | MM | | 88.964 | 121.773 | 0.316 | 8.043 | 0.376 | 0.859 | 87.479 | 119.906 | 0.310 | 7.909 | 0.370 | 0.863 | 86.863 | 119.924 | 0.308 | 7.853 | 0.370 | 0.863 | 92.120 | 122.513 | 0.327 | 8.328 | 0.378 | 0.857 |
| | MC | db6 | 92.940 | 126.538 | 0.330 | 8.403 | 0.391 | 0.847 | 99.922 | 134.295 | 0.354 | 9.034 | 0.415 | 0.828 | 109.305 | 147.429 | 0.388 | 9.882 | 0.455 | 0.793 | 97.456 | 133.004 | 0.346 | 8.811 | 0.411 | 0.831 |
| | MI | | 93.674 | 126.038 | 0.332 | 8.469 | 0.389 | 0.849 | 99.098 | 133.338 | 0.351 | 8.959 | 0.412 | 0.830 | 103.740 | 140.450 | 0.368 | 9.379 | 0.434 | 0.812 | 132.345 | 175.673 | 0.469 | 11.965 | 0.542 | 0.706 |
| | MM | | 89.383 | 122.345 | 0.317 | 8.081 | 0.378 | 0.857 | 87.593 | 119.946 | 0.311 | 7.919 | 0.370 | 0.863 | 88.083 | 121.228 | 0.312 | 7.964 | 0.374 | 0.860 | 92.391 | 122.814 | 0.328 | 8.353 | 0.379 | 0.856 |
| | MC | db7 | 92.697 | 125.701 | 0.329 | 8.381 | 0.388 | 0.849 | 102.242 | 136.774 | 0.363 | 9.244 | 0.422 | 0.822 | 106.415 | 142.988 | 0.377 | 9.621 | 0.442 | 0.805 | 101.109 | 137.162 | 0.359 | 9.141 | 0.424 | 0.821 |
| | MI | | 92.568 | 125.245 | 0.328 | 8.369 | 0.387 | 0.850 | 98.734 | 133.050 | 0.350 | 8.927 | 0.411 | 0.831 | 104.164 | 141.129 | 0.369 | 9.417 | 0.436 | 0.810 | 95.572 | 130.926 | 0.339 | 8.641 | 0.404 | 0.837 |
| | MM | | 89.838 | 123.002 | 0.319 | 8.122 | 0.380 | 0.856 | 87.770 | 120.005 | 0.311 | 7.935 | 0.371 | 0.863 | 88.304 | 121.489 | 0.313 | 7.984 | 0.375 | 0.859 | 92.572 | 122.865 | 0.328 | 8.369 | 0.379 | 0.856 |